\documentclass[conference]{IEEEtran}
\usepackage{times}

\usepackage[numbers]{natbib}
\usepackage{multicol}
\usepackage[bookmarks=true]{hyperref}
\usepackage[utf8]{inputenc}
\usepackage{booktabs}
\usepackage{multirow}
\usepackage{array}
\usepackage{amsmath}
\usepackage{tabularx}
\usepackage[table]{xcolor}
\usepackage{xspace}
\usepackage{cleveref}
\usepackage{graphicx}
\let\labelindent\relax
\usepackage{enumitem}
\usepackage{subcaption}
\usepackage{subfiles}
\usepackage{algorithm}
\usepackage{algpseudocode}
\usepackage{listings}

\usepackage{float} 
\usepackage{placeins}
\newcolumntype{Y}{>{\raggedright\arraybackslash}X}
\newcommand{\I}{\hspace*{0.9em}}
\newcommand{\II}{\hspace*{1.8em}}
\newcommand{\III}{\hspace*{2.7em}}
\newcommand{\IV}{\hspace*{3.6em}}

\usepackage[most]{tcolorbox}
\usepackage{enumitem}

\newtcolorbox{tasklistbox}{
    breakable,
    enhanced,
    colback=white,
    colframe=gray!30,
    boxrule=0.5pt,
    arc=2pt,
    left=5pt,
    right=5pt,
    top=5pt,
    bottom=5pt,
    fontupper=\small\ttfamily,
}

\newcommand{\taskseparator}[2]{%
    \par\addvspace{10pt}
    {\centering\hfill\small\bfseries\sffamily\color{#1}--- #2 ---\hfill}\par
    \addvspace{10pt}
}

\newcommand{\task}[2]{%
    \noindent\hangindent=3.5em\hangafter=1
    \makebox[3.5em][l]{\color{gray}\scriptsize #1}#2\par\addvspace{4pt}
}

\usepackage[most]{tcolorbox}

\newtcolorbox{indexbox}{
    breakable,
    enhanced,
    colback=white,
    colframe=gray!30,
    boxrule=0.5pt,
    arc=2pt,
    left=5pt,
    right=5pt,
    top=5pt,
    bottom=5pt,
    fontupper=\small\ttfamily,
}

\newlength{\indexlabelw}
\newcommand{\indexitem}[2]{%
  \noindent
  \makebox[\indexlabelw][l]{\bfseries\color{blue!70!black}#1:}%
  \parbox[t]{\dimexpr\linewidth-\indexlabelw\relax}{#2}%
  \par\addvspace{12pt}%
}

\newtcolorbox{pddlbox}{
    breakable,
    enhanced,
    colback=white,
    colframe=gray!30,
    boxrule=0.5pt,
    arc=2pt,
    fontupper=\small\ttfamily,
    left=15pt,
    right=15pt,
    top=10pt,
    bottom=10pt,
    before skip=10pt,
    after skip=10pt,
}

\newcommand{\pddlline}[1]{%
    \noindent\hangindent=2em\hangafter=1 #1\par\addvspace{1pt}
}

\newtcolorbox{jsonbox}{
    breakable,
    enhanced,
    colback=white,
    colframe=gray!30,
    boxrule=0.5pt,
    arc=2pt,
    fontupper=\small\ttfamily,
    left=10pt,
    right=10pt,
    top=5pt,
    bottom=5pt,
}

\newcommand{\jsonline}[1]{%
    \noindent\hangindent=2em\hangafter=1 #1\par
}

\newtcolorbox{sayplanbox}{
    breakable,
    enhanced,
    colback=white,
    colframe=gray!30,
    boxrule=0.5pt,
    arc=2pt,
    fontupper=\small\ttfamily,
    left=10pt,
    right=10pt,
    top=10pt,
    bottom=10pt,
}

\usepackage{pifont}
\usepackage{multicol}

\newcommand{\redcross}{\textcolor{red}{\ding{55}}}

\newtcolorbox{examplebox}[1]{
    breakable,
    enhanced,
    colback=white,
    colframe=gray!40,
    boxrule=0.8pt,
    arc=3pt,
    fontupper=\small,
    title=\textbf{Task #1},
    coltitle=black,
    colbacktitle=gray!10,
    attach boxed title to top left={yshift=-2mm, xshift=2mm},
    boxed title style={boxrule=0.5pt, colframe=gray!40}
}

\newcommand{\methodlabel}[1]{\vspace{5pt}\noindent\textbf{#1:}}

\newcommand{\algname}{UniPlan\xspace}

\newcommand{\SayPlan}{SayPlan\xspace}
\newcommand{\LLMAsPlanner}{LLM as Planner\xspace}
\newcommand{\DELTA}{DELTA\xspace}

\newcommand{\objects}{\texttt{(:objects)}\xspace}
\newcommand{\init}{\texttt{(:init)}\xspace}
\newcommand{\goal}{\texttt{(:goal)}\xspace}
\newcommand{\rob}{\texttt{(rob\_at\_node ?r ?n)}\xspace}
\newcommand{\obj}{\texttt{(obj\_at\_node ?o ?n)}\xspace}

\newcommand{\AblationVisionSG}{Vision\xspace}
\newcommand{\AblationDomain}{Expansion\xspace}
\newcommand{\AblationProblem}{Injection\xspace}
\newcommand{\AblationCompression}{Compression\xspace}

\begin{document}

\title{\algname: Vision-Language Task Planning for Mobile Manipulation with Unified PDDL Formulation}

\author{\authorblockN{Haoming Ye\authorrefmark{1}\authorrefmark{2},
Yunxiao Xiao\authorrefmark{2}\authorrefmark{3},
Cewu Lu\authorrefmark{1}\authorrefmark{2} and
Panpan Cai\authorrefmark{1}\authorrefmark{2}\authorrefmark{4}}
\authorblockA{\authorrefmark{1}Shanghai Jiao Tong University}
\authorblockA{\authorrefmark{2}Shanghai Innovation Institute}
\authorblockA{\authorrefmark{3}Beijing University of Posts and Telecommunications}
\authorblockA{\authorrefmark{4}Corresponding author: cai\_panpan@sjtu.edu.cn}}



%

\maketitle

\begin{abstract}

Integration of VLM reasoning with symbolic planning has proven to be a promising approach to real-world robot task planning. 
Existing work like UniDomain effectively learns symbolic manipulation domains from real-world demonstrations, described in Planning Domain Definition Language (PDDL), and has successfully applied them to real-world tasks. These domains, however, are restricted to tabletop manipulation.
We propose \algname, a vision-language task planning system for long-horizon mobile-manipulation in large-scale indoor environments, that unifies scene topology, visuals, and robot capabilities into a holistic PDDL representation.
\algname programmatically extends learned tabletop domains from UniDomain to support navigation, door traversal, and bimanual coordination. 
It operates on a visual-topological map, comprising navigation landmarks anchored with scene images.
Given a language instruction, \algname retrieves task-relevant nodes from the map and uses a VLM to ground the anchored image into task-relevant objects and their PDDL states; next, it reconnects these nodes to a compressed, densely-connected topological map, also represented in PDDL, with connectivity and costs derived from the original map; Finally, a mobile-manipulation plan is generated using off-the-shelf PDDL solvers.
Evaluated on human-raised tasks in a large-scale map with real-world imagery, \algname significantly outperforms VLM and LLM+PDDL planning in success rate, plan quality, and computational efficiency.
\end{abstract}

\begin{figure*}[!t]
\centering
\includegraphics[width=\textwidth]{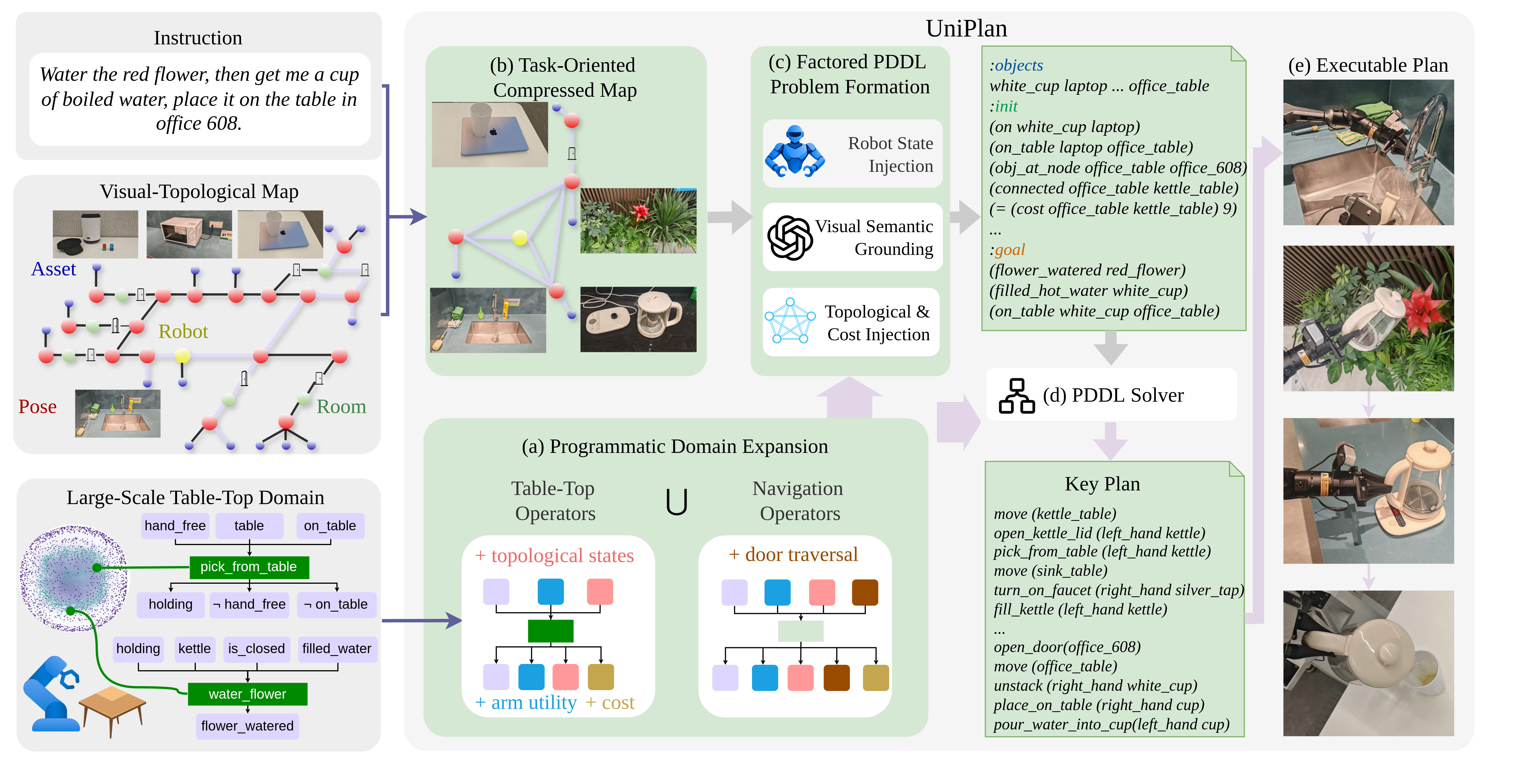}
\caption{Overview of \algname. See detailed descriptions in Section \ref{sec:overview}.}
\label{fig:overview}
\end{figure*}

\section{Introduction}

Vision-language task planning for long-horizon mobile manipulation in large-scale indoor environments remains a fundamental challenge in robotics.
Such tasks require a robot to reason jointly over visual observations, spatial structure, and symbolic constraints, while operating across many rooms, objects, and navigation states.
The core difficulty lies in scaling task reasoning beyond local, tabletop settings, without sacrificing perceptual fidelity or planning reliability under complex, long-horizon constraints.

A dominant line of work \cite{rana2023sayplan, liu2024delta} formulates planning purely in language, where visual observations are first converted into textual scene graphs \cite{3d-scene-graph}, and an LLM performs reasoning over these symbolic descriptions.
While effective in constrained settings, this paradigm inevitably incurs irreversible information loss when mapping vision to language.
Crucially, which visual attributes matter depends on the task itself: retrieving an apple for eating depends on freshness, while retrieving one for decoration depends on appearance.
No fixed, task-agnostic scene graph can anticipate all such requirements in advance.
This observation suggests that visual information must be preserved and queried on demand, motivating planning frameworks that operate directly over images rather than static textual abstractions.

Recent vision-language planners \cite{embodied-reasoner} attempt to reason directly from images, using VLMs to predict high-level mobile-manipulation actions from current and historical observations.
However, existing approaches are fundamentally mapless and operate over a small number of online images, typically limited to a single room.
As a result, they struggle to form a holistic understanding of large environments containing hundreds of visual contexts.
Moreover, relying solely on VLM reasoning for long-horizon planning has been shown to be brittle under complex operational constraints, while most methods depend heavily on simulation-based training \cite{ai2thor}, limiting robustness and generalization to real-world settings.

In this work, we propose a vision-language task planning system, \algname, designed explicitly for large-scale, real-world indoor environments, that unifies all reasoning in a Planning Domain Definition Language \cite{pddl, pddl2.1} formulation.
Our system is built on four principles:
\begin{enumerate}
    \item maintaining a visual-topological map that preserves comprehensive image-level information of the environment;
    \item combining VLM-based visual grounding with symbolic PDDL planning to handle long-horizon tasks and complex constraints robustly;
    \item avoiding task- or environment-specific training by leveraging strong pre-trained VLMs and large-scale pre-trained symbolic domains;
    \item ensuring scalability by reasoning only over task-relevant visual contexts, rather than the entire observation history.
\end{enumerate}

Concretely, we build \algname upon UniDomain \cite{unidomain}, which learned a unified, large-scale PDDL manipulation domain from real-world demonstrations. We programmatically expand these domains to support navigation, door traversal, and bimanual coordination, enabling mobile manipulation in large indoor environments.
Given a visual-topological map and a natural language instruction, our system retrieves task-relevant nodes and constructs a compressed, task-oriented topological map with recalibrated connectivity and traversal costs.
Visual grounding is performed only on the retrieved images to infer object states, while robot states and navigation dynamics are injected programmatically, yielding a factored PDDL problem that decouples perception, embodiment, and environment structure.
An off-the-shelf PDDL solver (e.g. Fast Downward \cite{fastdownward}) then generates the final mobile-manipulation plan.

We evaluate \algname on a large-scale indoor environment under four settings: single-arm vs. dual-arm, and environments with vs. without doors.
Across all settings, our method significantly outperforms strong baselines, including direct VLM planning, language-only planners such as SayPlan \cite{rana2023sayplan}, and PDDL-augmented approaches like DELTA \cite{liu2024delta}, achieving substantial gains in success rate, plan quality, and computational efficiency.
Ablation studies further demonstrate that factored problem construction is critical for success of planning, domain expansion is essential for plan optimality, and task-oriented map compression is the key to scalability.

\section{Related Work}

\subsection{Scene Graph}

3D Scene Graphs (SGs) \cite{3d-scene-graph} have been widely adopted to abstract environments into hierarchical concepts. Early systems construct these graphs through geometric clustering \cite{chang2023hydra-multi, hughes2022hydra}, while recent open-vocabulary methods enrich graph nodes with semantic labels or feature embeddings from foundation models \cite{gu2024conceptgraphs, werby2024hov-sg}. However, this abstraction into rigid hierarchies can create a semantic gap, discarding nuanced visual details (e.g., object states, precise spatial relations) critical for task planning.

While multi-modal SGs attempt to mitigate this by anchoring nodes with visual data \cite{werby2025keysg, li2024llm-enhanced-sg, ali2025graphpad}, their primary focus remains on open-ended querying rather than generating symbolic states for task planning. In contrast, we adopt a vision-anchored topological map as our environmental representation. Instead of maintaining explicit object nodes, our approach retains rich perceptual details implicitly within the anchored images, using VLMs to ground dynamic objects into PDDL states only on-demand, directly supporting robust task planning.

\subsection{LLM-based Planning}

Recent works integrate LLMs with SGs for robot task planning. SayPlan \cite{rana2023sayplan} performs semantic search on a collapsed SG to enable high-level planning and replanning. Other approaches focus on managing information overload by employing retrieval mechanisms, such as dynamically selecting task-relevant subgraphs \cite{booker2024embodiedrag} or using structured queries \cite{ray2025structured} to overcome context window limitations.

Another line of work uses the SG as a verifier. VeriGraph \cite{ekpo2024verigraph} checks action preconditions to correct LLM-generated plans, while LookPlanGraph \cite{onishchenko2025lookplangraph} updates the SG with VLMs to handle dynamic environments. However, by using SGs primarily for retrieval or verification, these methods still rely on the LLM's implicit, language-based reasoning for the core planning process. Consequently, they often struggle to enforce physical and operational constraints in long-horizon mobile manipulation, leading to ungrounded or infeasible plans.

\subsection{Integration of PDDL and LLM Planning}
To address the lack of formal structure in pure LLM-based planners, many approaches integrate the symbolic reasoning like PDDL. These methods often ground PDDL states within Scene Graphs; for example, by parsing states directly from visual nodes \cite{herzog2025domain-conditioned-sg}, using LLMs to relax constraints for infeasible goals \cite{musumeci2025context-matters}, or attempting to generate the entire PDDL model from scene context \cite{liu2024delta}. However, these approaches face a critical challenge: they rely on either hand-crafted domains that are rigid and environment-specific, or on language-generated domains that are often fragile, as LLMs struggle to capture the precise physical dynamics required for robust execution.

To overcome the fragility of language-generated rules, a recent line of work focuses on learning or refining PDDL domains from external signals, such as human or environment feedback \cite{interpret, ada, lasp} or visual demonstrations \cite{blade, pix2pred, 2025icra, 2021iros}. Notably, UniDomain \cite{unidomain} scales this by pre-training a general-purpose PDDL domain from large-scale manipulation datasets, enabling strong compositional generalization. However, the success of these methods has been predominantly confined to local, tabletop manipulation. They lack the inherent spatial logic to model robot navigation (e.g., room traversal) and cannot directly interface with large-scale environment representations.

Instead of collecting expensive mobile demonstrations to re-learn an entire domain, our work introduces a general programmatic extension algorithm. This approach effectively augments powerful, learned tabletop domains (e.g. from UniDomain) with crucial capabilities like navigation, door traversal, and bimanual actions, enabling the seamless integration of learned manipulation skills with visual-topological maps, unlocking robust planning for long-horizon mobile manipulation tasks.

\section{System Overview}
\label{sec:overview}
\textit{\algname} (Figure~\ref{fig:overview}) is a vision-language task planning system that generates high-level mobile-manipulation plans from natural language instructions in large-scale indoor environments.

\algname operates on a visual-topological map that explicitly encodes spatial structure while preserving visual information.
The map consists of two layers: a topological layer that connects navigation landmarks such as room entrances and operational standing points, and a visual layer that anchors scene images to nodes in the topological graph.

\algname is also built upon UniDomain, which learned a large-scale PDDL domain from manipulation demonstrations.
We programmatically extend these manipulation-only domains with navigation operators, arm-specific pre- and post-conditions, and door-related predicates and dynamics (Figure~\ref{fig:overview}a).
These extensions are performed once and are reusable across environments and tasks, enabling mobile manipulation without additional training. With an expanded PDDL domain from learned tabletop domains~\cite{unidomain, interpret, blade} and a visual-topological map, \algname performs task planning in the following stages:

\textbf{Task-Oriented Map Compression (Figure~\ref{fig:overview}b).}
Given a natural language instruction, \algname retrieves task-relevant nodes from the visual-topological map and collects the images anchored at these nodes.
Connectivity and traversal costs among the retrieved nodes are recalibrated by querying the original map, and the nodes are recombined into a compressed topological map .
This compressed map is small yet information-complete for the task, allowing planning to scale efficiently to large environments while reasoning over only a few images.

\textbf{Factored PDDL Problem Formation (Figure~\ref{fig:overview}c).}
The PDDL planning problem is constructed by decoupling robot, object, and environment factors:
\begin{itemize}
    \item Robot states (e.g., hand occupancy and arm availability) are injected programmatically as PDDL predicates.
    \item Object entities and task-relevant object states are grounded by the VLM from the retrieved images.
    \item Environment structure, including the robot’s global location, node connectivity, and traversal costs, is extracted from the compressed map and encoded as PDDL predicates and action costs.
\end{itemize}
This factorization substantially reduces the burden on the VLM.
Rather than generating the entire symbolic problem, the VLM only performs local visual grounding, while global spatial relationships are calibrated deterministically from the map.
As a result, \algname avoids cross-image reasoning by the VLM, which is a key limitation of existing vision-language PDDL planners.

\textbf{Unified PDDL Planning.}
Finally, an off-the-shelf PDDL solver (Figure~\ref{fig:overview}d) generates a mobile-manipulation plan over the unified domain (Figure~\ref{fig:overview}e).
Because navigation and manipulation actions are modeled within a single PDDL formulation, the planner can reason jointly over movement and manipulation, producing globally consistent and cost-aware plans.
This unified planning is significantly more efficient and reliable than decoupled navigation–manipulation pipelines.

\section{Programmatic Domain Expansion}
\label{sec:domain_extension}
To scale a learned tabletop PDDL domain to long-horizon mobile manipulation, we employ a programmatic expansion strategy that modifies the domain's structure directly. We leverage the fact that PDDL is defined by a context-free grammar (CFG), which allows us to rigorously parse the domain into an Abstract Syntax Tree (AST), enabling systematic traversal and modification of its logical structure without risking syntax errors.
A PDDL operator is first decomposed into its syntactic constituents (parameters, preconditions, and effects).
We then apply a set of deterministic, general-purpose rewrite rules over this tree structure.
This transformation process ensures that the expansion is (i) \textit{syntactically valid} by construction, (ii) \textit{consistent} across all operators in the domain, and (iii) \textit{reusable} across different environments.
We implement this rewriting logic using the Unified Planning framework~\cite{unified_planning_softwarex2025} to facilitate AST manipulation. Visualizations of the AST structure and tree traversal examples are in the supplementary materials.

\subsection{Semantic Anchors}
Learned domains often employ heterogeneous predicate signatures to represent gripper availability and object attachment (e.g., \texttt{(free)} vs. \texttt{(hand\_empty)}, or \texttt{(inhand ?o)} vs. \texttt{(in\_gripper ?r ?o)}).
Acknowledging their functional equivalence despite syntactic variations, we adopt two canonical predicates as our semantic anchors: \texttt{(hand\_free ?r)} and \texttt{(holding ?r ?o)}. This formulation, consistent with UniDomain \cite{unidomain}, explicitly models the robot parameter \texttt{?r}.
These anchors provide a stable interface for subsequent AST rewrites: by reliably identifying gripper-state subtrees, we can systematically inject navigation logic and multi-arm parameters without depending on the specific naming conventions of the original source domain.
\label{sec:anchors}

\subsection{Navigation and Door Expansion}
Tabletop domains implicitly assume all actions happen in a fixed workspace. 
To make every manipulation operator location-aware, we introduce two topological predicates: \rob and \obj, to denote that the robot \texttt{?r} and the object \texttt{?o} are located at a node \texttt{?n} in the visual--topological map.
For each operator, we first augment its parameter list with a location node \texttt{?n} and then perform two coordinated AST rewrites:

\textbf{Precondition.}
We inject \rob into all operators to restrict actions to a specific map node.
Furthermore, for each \textbf{non-robot} parameter \texttt{?o}, we add the precondition \obj unless the object is already held by the robot, which is indicated by an existing \texttt{(holding ?r ?o)} precondition.

\textbf{Effect.}
The operator's effects are updated to reflect changes in object location driven by the \texttt{holding} state.
Asserting \texttt{(holding ?r ?o)} (e.g., a pick action) signifies the object is no longer at the node; we therefore inject \texttt{(not \obj)} into the effects.
Conversely, negating \texttt{(holding ?r ?o)} (e.g., a place action) means the object is now at the node, so we add the effect \obj.
Operators that do not alter the holding state, such as \texttt{turn\_on\_faucet}, represent in-place interactions that only update an object’s functional state without moving it, and thus leave its node-level location unchanged.

Since manipulation operators are now location-dependent, a corresponding action is required for the robot to travel between nodes. To this end, we introduce a connectivity predicate \texttt{(connected ?n1 ?n2)} to denote that node ?n1 and node ?n2 are connected in the visual–topological map, and a navigation operator \texttt{move\_robot}, to update \rob.

In the topological map, \texttt{(connected ?n1 ?n2)} captures only geometric adjacency, while door traversal depends on the door state. We therefore model doors as stateful edges by adding predicate \texttt{(has\_door ?n1 ?n2)}. We then introduce the operator \texttt{open\_door} to enforce ``open before traverse''. The full PDDL definitions for the \texttt{move\_robot} and \texttt{open\_door} in the supplementary materials.

\subsection{Bimanual Expansion}
To support bimanual operation, we lift the anchors to be hand-specific:
\texttt{(hand\_free ?r ?h)} and \texttt{(holding ?r ?h ?o)}.
We add a static configuration predicate \texttt{(rob\_has\_hand ?r ?h)} and rewrite each operator by:
(1) adding a hand parameter \texttt{?h}, and
(2) propagating \texttt{?h} to all anchor occurrences in preconditions and effects.
This makes hand occupancy explicit, enabling the solver to reason about which hand is available for actions.

\subsection{Cost Modeling}
The above steps ensure \emph{feasibility}. To support cost-aware planning over the compressed map introduced in Sec.~\ref{sec:map_compression}, we further make plans \emph{efficient} by enabling cost-optimal planning.
We introduce a numeric function \texttt{(travel\_cost ?n1 ?n2)} (instantiated from the compressed map) to represent the cost of traversing between node \texttt{?n1} and node \texttt{?n2} in the visual--topological map, and modify the navigation operator \texttt{move\_robot} to accumulate action cost.
All manipulation operators are assigned a constant cost.
Finally, we specify the optimization objective in the \emph{problem} file by setting the metric so that the solver jointly optimizes navigation costs and manipulation effort in a single unified plan.

\par\smallskip
\noindent\textbf{Generality.}
While our implementation builds on the UniDomain, the proposed AST-based rewrite rules are domain-agnostic: they rely only on identifying a small set of semantic anchors (Sec.~\ref{sec:anchors}) and applying deterministic, syntax-preserving transformations. Consequently, the same systematic expansion can be applied to any tabletop PDDL domain, including learned domains from \cite{interpret, blade, 2021iros, 2025icra}, as well as manually engineered tabletop domains.

\section{Visual-Topological Map}
\label{sec:map}

\subsection{Map Representation}
We represent the environment as a \textbf{vision-anchored topological map}.
The node set comprises three distinct categories of \textit{static} environmental elements:
\textbf{(1) Pose Nodes}, representing discrete navigation waypoints and the connectivity between them;
\textbf{(2) Room Nodes}, marking transitions between spatial zones (e.g., doorways or hallway entrances) to serve as topological bottlenecks for pathfinding; and
\textbf{(3) Asset Nodes}, denoting large, static furniture (e.g., tables, counters) where manipulation tasks take place.
Crucially, each \textit{Asset} node is anchored with a set of high-resolution observation images, capturing the visual state of the surface and its contents from strategic viewpoints.

In this work, we assume the visual-topological map is constructed offline prior to deployment.
The topological structure can be derived from previous work \cite{hughes2022hydra, chang2023hydra-multi}.
Similarly, the selection of anchored images for Asset nodes can be similarly inspired by methods for selecting visual coverage of objects or rooms, as discussed in \cite{li2024llm-enhanced-sg, werby2025keysg}, but adapted to the visual coverage of manipulation surfaces.
Once constructed, the map provides a static backbone for the environment, over which \algname performs dynamic, task-oriented reasoning.

\subsection{Task-Oriented Map Compression}
\label{sec:map_compression}
Given a language instruction for a task, we generate a compact, task-specific map through a two-stage process: semantic retrieval and topology compression.

\textbf{Semantic Retrieval.}
We first identify the specific locations required for the task.
During an Offline Indexing phase, we prompt a VLM to generate unstructured captions for images anchored to Asset nodes. These captions simply list visible objects and furniture, avoiding complex attribute descriptions, to serve as a lightweight textual index.
During Online Selection, given a user instruction, an LLM identifies the subset of relevant asset nodes by matching the instruction against these pre-computed captions. This avoids processing the entire collection of anchored images in the map, allowing the system to focus exclusively on task-relevant contexts.

\textbf{Topology Compression.}
Next, we construct a dense, abstract graph connecting the selected nodes, alongside the robot's current location, to encode navigation dynamics. We compute pairwise connectivity between nodes using the Dijkstra algorithm on the original raw map. For paths within the same accessible zone, we create logical ``shortcut'' edges weighted by the actual travel cost, abstracting away intermediate waypoints.
Crucially, closed doors are treated as physical boundaries. We partition the compression so that direct shortcuts are never created through a closed door; instead, the specific door traversal edges are preserved.
The resulting map effectively represents the environment as a set of fully connected components separated by doors, providing the downstream planner with a minimal yet physically consistent state space.

\section{Factored PDDL Problem Formation}
\label{sec:problem_generation}
With the extended PDDL domain and the compressed topological map, we construct the PDDL problem for a task by synthesizing three decoupled factors: the robot's intrinsic state, VLM-grounded semantic states, and the environment's topological constraints.

\paragraph{Robot State Injection}
The robot's state is obtained directly from its onboard perception system. We inject the robot’s metadata and arm configurations (e.g., \texttt{left\_hand}, \texttt{right\_hand}) into the PDDL \objects block. In the \init block, we define its initial location, such as \texttt{(rob\_at\_node rob pose\_1)}, and static kinematic predicates, such as \texttt{(rob\_has\_hand rob left\_hand)}.

\paragraph{Visual Semantic Grounding}
For each task-relevant node identified in Section \ref{sec:map_compression}, we prompt a VLM with its anchored images, the task instruction, and the domain definition. The VLM is tasked with generating the semantic components of the PDDL problem, including task-relevant object entities for the \objects block, their initial symbolic states (e.g., \texttt{(on\_table apple table)}) for the \init block, and the final \goal condition.
We programmatically post-process the VLM's output by injecting spatial anchors, such as \texttt{(obj\_at\_node apple asset\_node\_5)}, ensuring every detected object is grounded to its respective location. To maintain state consistency, the VLM is explicitly prohibited from generating robot-specific predicates, which are managed exclusively by the robot configuration module.

\paragraph{Topological and Cost Injection}
Finally, we encode the compressed map structure into the PDDL \init block. 
Graph connectivity and edge weights are directly mapped to predicates, e.g., \texttt{(connected pose\_1 pose\_2)}, and numeric functions, e.g., \texttt{(= (cost pose\_1 pose\_2) 15)}. 
Door states are similarly injected, such as \texttt{(has\_door pose\_3 pose\_4)} for prohibiting traversal. 
This provides the planner with a cost-aware topological representation, enabling it to jointly optimize navigation and manipulation in its plan.

\noindent \textbf{Plan Generation and Refinement.}
We solve the unified PDDL problem using an off-the-shelf planner. As the resulting plan contains abstract \texttt{move\_robot} actions between compressed nodes, we perform a final refinement step: each abstract edge is expanded back into the sequence of low-level waypoints cached during map compression. 
This yields a seamless executable plan bridging high-level symbolic reasoning and low-level geometric navigation.

\section{Experiments}
\label{sec:experiments}

\subsection{Experimental Setup}

\noindent \textbf{Evaluation Tasks} 
We pre-build a visual-topological map from a real-world large-scale indoor space, comprising 43 pose nodes, 18 room nodes, 18 doors, 31 asset nodes and more than 100 objects.

We collected 50 real-world tasks from humans, spanning 17 primitive action types: \texttt{pick}, \texttt{place\_in}, \texttt{place\_on}, \texttt{place\_under}, \texttt{open}, \texttt{close}, \texttt{pour}, \texttt{cut}, \texttt{stir}, \texttt{scoop}, \texttt{fold}, \texttt{wipe}, \texttt{turn\_on}, \texttt{turn\_off}, \texttt{hang\_on}, \texttt{open\_door}, and \texttt{move}, categorized into three complexity levels:
\begin{itemize}
    \item \textbf{Simple (Tasks 1-10):} Involve 1-2 asset nodes, short horizons (2-4 high-level actions\footnote{We define ``high-level actions'' as the resulting sequence obtained by compressing consecutive ``move'' actions between nodes into a single action, while retaining the remaining actions from the original sequence.}), and limited action types (e.g., pick-and-place). An example task is ``Place the mango into the fridge."
    \item \textbf{Moderate (Tasks 11-40):} Involve 2-4 asset nodes, moderate horizons (5-10 high-level actions), spanning all 17 action types. An example task is ``Wash the cloth on the couch and hang it on the drying rack."
    \item \textbf{Compositional (Tasks 41-50):} Involve 3-6 nodes, long-horizons (10-20 high-level actions), composed from \textbf{Moderate Tasks}. An example task is ``Wash the apple in the fridge, put it on the table in office 612, fold the cloth from the couch, and place it on the same table."
\end{itemize}

All tasks are evaluated under four settings: single-arm vs. dual-arm, and environments with vs. without closed doors (requiring navigation-manipulation coordination). The complete task list is provided in the supplementary materials.

\noindent \textbf{Base Domain.} 

Our base domain uses the UniDomain \cite{unidomain} pipeline to construct PDDL domains from real-world videos tailored for household tasks. These are integrated with reused tabletop manipulation domains into a comprehensive tabletop domain containing 81 operators and 79 predicates. It has been expanded to a mobile manipulation domain (Section \ref{sec:domain_extension}) to support PDDL formulation of tasks and planning.

\begin{table*}[t!]
    \centering
    \caption{Performance comparison of \algname and baseline methods across different robotic task configurations.}
    \label{tab:main_results}
    \small
    \setlength{\tabcolsep}{4pt}
    \begin{tabular}{llcccc}
        \toprule
        \textbf{Setting} & \textbf{Method} & \textbf{SR (\%)} $\uparrow$ & \textbf{LLM Calls} $\downarrow$ & \textbf{$T_{\text{think}}$ / $T_{\text{plan}}$ [s]} $\downarrow$ & \textbf{RPQG (\%)} $\uparrow$ \\
        \midrule
        \multirow{4}{*}{\shortstack[l]{Single-Arm \\ No Door}} 
        & \LLMAsPlanner & 40.00 $\pm$ 4.62 & \textbf{2.00 $\pm$ 0.00} & \makebox[5.5em][r]{\textbf{19.07 $\pm$ 0.47}} / \makebox[5.5em][c]{---} & 2.90 $\pm$ 0.71 \\
         
        & \SayPlan & 41.00 $\pm$ 5.29 & 9.13 $\pm$ 0.14 & \makebox[5.5em][r]{64.18 $\pm$ 2.27} / \makebox[5.5em][c]{---} & 5.42 $\pm$ 1.66 \\
         
        & \DELTA & 26.00 $\pm$ 5.89 & 4.04 $\pm$ 0.01 & \makebox[5.5em][r]{71.50 $\pm$ 2.92} / \makebox[5.5em][l]{0.31 $\pm$ 0.01} & 7.77 $\pm$ 1.21 \\
         
        & \textbf{\algname (Ours)} & \textbf{83.50 $\pm$ 4.43} & \textbf{2.00 $\pm$ 0.00} & \makebox[5.5em][r]{21.23 $\pm$ 0.34} / \makebox[5.5em][l]{0.43 $\pm$ 0.33} & \textbf{---} \\
        \midrule
        \multirow{4}{*}{\shortstack[l]{Single-Arm \\ With Door}} 
        & \LLMAsPlanner & 17.50 $\pm$ 4.73 & \textbf{2.00 $\pm$ 0.00} & \makebox[5.5em][r]{22.63 $\pm$ 0.89} / \makebox[5.5em][c]{---} & 5.49 $\pm$ 2.72 \\
         
        & \SayPlan & 33.00 $\pm$ 5.03 & 9.75 $\pm$ 0.08 & \makebox[5.5em][r]{71.30 $\pm$ 1.08} / \makebox[5.5em][c]{---} & 10.26 $\pm$ 2.99 \\
         
        & \DELTA & 18.50 $\pm$ 5.97 & 4.09 $\pm$ 0.02 & \makebox[5.5em][r]{75.12 $\pm$ 2.72} / \makebox[5.5em][l]{0.35 $\pm$ 0.03} & 16.17 $\pm$ 7.12 \\
         
        & \textbf{\algname (Ours)} & \textbf{84.00 $\pm$ 2.83} & \textbf{2.00 $\pm$ 0.00} & \makebox[5.5em][r]{\textbf{21.55 $\pm$ 0.78}} / \makebox[5.5em][l]{0.42 $\pm$ 0.31} & \textbf{---} \\
        \midrule
        \multirow{4}{*}{\shortstack[l]{Dual-Arm \\ No Door}} 
        & \LLMAsPlanner & 40.50 $\pm$ 1.91 & \textbf{2.00 $\pm$ 0.00} & \makebox[5.5em][r]{\textbf{18.38 $\pm$ 0.66}} / \makebox[5.5em][c]{---} & 2.49 $\pm$ 0.43 \\
         
        & \SayPlan & 49.50 $\pm$ 5.26 & 8.77 $\pm$ 0.18 & \makebox[5.5em][r]{60.55 $\pm$ 1.82} / \makebox[5.5em][c]{---} & 6.12 $\pm$ 0.90 \\
         
        & \DELTA & 26.50 $\pm$ 4.73 & 4.05 $\pm$ 0.03 & \makebox[5.5em][r]{71.46 $\pm$ 7.45} / \makebox[5.5em][l]{0.35 $\pm$ 0.02} & 12.19 $\pm$ 3.33 \\
         
        & \textbf{\algname (Ours)} & \textbf{82.00 $\pm$ 2.31} & \textbf{2.00 $\pm$ 0.00} & \makebox[5.5em][r]{21.73 $\pm$ 1.83} / \makebox[5.5em][l]{0.66 $\pm$ 0.61} & \textbf{---} \\
        \midrule
        \multirow{4}{*}{\shortstack[l]{Dual-Arm \\ With Door}} 
        & \LLMAsPlanner & 39.50 $\pm$ 1.91 & \textbf{2.00 $\pm$ 0.00} & \makebox[5.5em][r]{22.50 $\pm$ 0.49} / \makebox[5.5em][c]{---} & 2.74 $\pm$ 1.10 \\
         
        & \SayPlan & 42.50 $\pm$ 4.43 & 9.09 $\pm$ 0.33 & \makebox[5.5em][r]{59.64 $\pm$ 2.15} / \makebox[5.5em][c]{---} & 4.65 $\pm$ 2.44 \\
         
        & \DELTA & 21.50 $\pm$ 4.43 & 4.02 $\pm$ 0.00 & \makebox[5.5em][r]{72.60 $\pm$ 3.37} / \makebox[5.5em][l]{0.33 $\pm$ 0.01} & 7.25 $\pm$ 3.22 \\
         
        & \textbf{\algname (Ours)} & \textbf{85.50 $\pm$ 2.52} & \textbf{2.00 $\pm$ 0.00} & \makebox[5.5em][r]{\textbf{21.89 $\pm$ 0.91}} / \makebox[5.5em][l]{0.11 $\pm$ 0.01} & \textbf{---} \\
        \bottomrule
        \addlinespace

    \end{tabular}
\end{table*}

\subsection{Baselines}

We compare \algname against three state-of-the-art approaches. All methods use the same base prompts where applicable for fairness.

\noindent \textbf{LLM as Planner.} 
A naive baseline using an LLM directly for planning. To ensure a fair comparison and reduce reasoning overhead, we equip it with the same retrieval and compressed map as \algname (Section~\ref{sec:map_compression}). The LLM receives the compressed map with images and a prompt stating all relevant constraints encoded in the tabletop PDDL domain. The output is post-processed to expand the action sequence.

\noindent \textbf{SayPlan.} 
SayPlan plans on a collapsed scene graph, iteratively expanding nodes and refining plans based on simulator feedback. Since it requires a text-based scene graph, we prompt a VLM to convert node images into textual descriptions. Initially, all rooms are contracted, and asset nodes outside rooms are visible but unexpanded. To ensure a competitive implementation, we use our task-environment emulator as the ``scene graph simulator'' to verify execution feasibility. We set the maximum search depth to 16 and re-planning attempts to 5. Constraints are hinted in the prompt similarly to the LLM baseline.

\noindent \textbf{DELTA.} 
DELTA follows a five-stage pipeline: (1) Domain Generation, (2) Scene Graph Pruning, (3) Problem Generation, (4) Goal Decomposition, and (5) Autoregressive Sub-Task Planning.
Since end-to-end domain generation from natural language descriptions proved intractable for our complex tasks, we adapt DELTA by replacing the ``Domain Generation'' with ``Domain Selection'', where the LLM selects necessary operators from our expanded domain based on the task instruction. The text scene graph is constructed similarly to SayPlan.

\subsection{Metrics}

\begin{itemize}
    \item \textbf{Success Rate (SR):} The ratio of successfully completed tasks. A task is failed if: (1) action constraints are violated (e.g., opening a door with an occupied hand), (2) the final state does not match the goal, or (3) the agent operates on non-existent objects.
    \item \textbf{LLM Calls:} The total number of API calls per task.
    \item \textbf{Thinking Time ($T_{\text{think}}$):} The total duration of LLM API inference.
    \item \textbf{Planner Time ($T_{\text{plan}}$):} The time spent by the symbolic planner (e.g., Fast Downward).

    \item \textbf{Relative Plan Quality Gain (RPQG):} 
    Since obtaining absolute optimal ground-truth plans is computationally intractable in complex, open-ended domains, and success rates vary significantly across methods, standard efficiency metrics are often misleading. 
    Directly averaging step counts introduces bias, as baselines often fail on longer, more complex tasks, artificially lowering their average step count. Conversely, metrics like SPL (Success weighted by Path Length) \cite{anderson2018SPL} tend to be dominated by the success rate itself when the performance gap is large, diluting the assessment of plan quality.
    
    To address this, RPQG measures the relative efficiency of \algname against a specific baseline \textit{only} on the intersection of tasks where \textbf{both methods succeed}. For a set of \textbf{both-succeeded successful} tasks, RPQG is defined as the average relative reduction in action steps:
    \begin{equation}
        \text{RPQG} = \frac{1}{N} \sum_{i=1}^{N} \frac{S_{\text{base}}^{(i)} - S_{\text{ours}}^{(i)}}{S_{\text{base}}^{(i)}} \times 100\%
    \end{equation}
    where $S^{(i)}$ denotes the total action steps for task $i$. A positive RPQG indicates that \algname generates shorter, more efficient plans than the baseline within the same feasible task references.
\end{itemize}

\noindent \textbf{Simulation and Environment.} 
Many existing works \cite{rana2023sayplan, embodied-reasoner, unidomain} in real-world robot task planning often rely on human or LLM judges for evaluation, which can be inefficient or unstable. To ensure reproducible, objective, and automatic evaluation, we implemented a task-environment emulator as a python program. This environment rigorously encodes all action preconditions, effects, state transitions and goal tests. It allows for the automated verification of plan feasibility and goal satisfaction, bypassing the inconsistencies of subjective evaluation methods.

To ground actions in the simulated environment, we implemented action parsers for each method. For \LLMAsPlanner and \SayPlan, we prompt the LLM to follow a fixed action format (e.g., pick(robot, left\_hand, apple)). For \DELTA and \algname, we parse each PDDL operator using a pre-built, parameterized mapping table. After grounding the heterogeneous action formats, we map each generated object name to a unique object entity in the environment, leveraging the LLM to ensure maximum accuracy.

To isolate high-level planning performance from low-level execution errors, we assume perfect low-level policy for action execution, following the standard evaluation protocol established in prior work~\cite{rana2023sayplan, liu2024delta}.

To ensure fast response times, all methods utilize \textbf{GPT-5.2} with \texttt{temperature=0.0} and ``none'' reasoning effort level. Despite this, outputs remain stochastic; thus, we run each setting $N=4$ times and report the mean and standard deviation.
For all PDDL-based methods, we utilize the Fast Downward \cite{fastdownward} with the \texttt{seq-opt-lmcut} engine to ensure cost-optimal planning.

\subsection{Results and Analysis}
\label{sec:results}

Quantitative results are summarized in Table~\ref{tab:main_results}. \algname demonstrates strong performance across all metrics, achieving the highest success rates while generating the most efficient plans.

\noindent \textbf{Success Rate and Robustness.}
\begin{itemize}
    \item LLM-centric baselines exhibit distinct grounding limitations. \textbf{\LLMAsPlanner} struggles with \textit{Constraint Satisfaction}, lacking the physical grounding to respect \textit{Physical Limits} in constrained settings (e.g., Single-Arm tasks). While \textbf{\SayPlan} leverages \textit{Semantic Priors} for efficient retrieval, it falters in \textit{Semantically Atypical} scenarios, where its bias toward high-probability locations causes it to overlook targets placed in unconventional spots.
    \item \textbf{\DELTA}, relying on formal logic via \textit{Domain Selection}, faces a severe \textit{Modeling Challenge}. The \textit{Complexity} of translating intricate, coupled physics into a consistent PDDL subset often yields unsolvable problem files, limiting performance in long-horizon tasks.
    \item In contrast, \textbf{\algname} achieves a robust 83.75\% average Success Rate (SR). By effectively balancing reasoning with operational constraints, our method maintains stability even in the demanding \textit{Single-Arm with Door} setup. unlike baselines constrained by hallucinations or rigid formalisms, \textbf{\algname} demonstrates strong \textit{Generalization} across both physical obstacles and irregular layouts.
\end{itemize}

\noindent \textbf{Plan Quality (RPQG).}
Beyond success rates, \algname generates significantly higher-quality plans, as evidenced by positive Relative Plan Quality Gains (RPQG) across all baselines (ranging from $\sim2\%$ to $\sim16\%$).
\begin{itemize}
    \item \textbf{Local vs. Global Optimality:} 
    \textbf{\DELTA} decomposes tasks into subgoals, which simplifies solving on a full map but does not guarantee global optimality, leading to redundant traversals (reflected in the high RPQG of 16.17\% in constrained settings). In contrast, \algname leverages the symbolic planner on a compressed map, guaranteeing a globally optimal path for the constructed problem.
    \item \textbf{Implicit vs. Explicit Reasoning:} \textbf{\LLMAsPlanner} and \textbf{\SayPlan} rely on implicit LLM reasoning. Without an underlying cost model, they tend to generate ``meandering'' plans that are logically valid but inefficient. \algname effectively performs explicit symbolic reasoning to ensure plan optimality.
\end{itemize}

\noindent \textbf{Computational Overhead.}
\SayPlan and \DELTA require iterative re-planning or autoregressive generation, resulting in high latency ($60\sim70$s). Conversely, \algname is more efficient in terms of both latency and the number of LLM calls. By retrieving task-oriented nodes via a lightweight textual index and decoupling reasoning (VLM grounding) from planning (symbolic), we limit expensive VLM calls to only two passes. Notably, the subsequent symbolic planning time is negligible ($<0.7$s) due to the compressed map.

\begin{table*}[t!]
    \centering
    \caption{Ablation study of \algname components. \textbf{Full \algname} serves as the reference. \textbf{RPQG} indicates the relative quality gain of the Full method compared to each ablated variant (higher indicates the removed component was crucial for efficiency).}
    \label{tab:ablation_results}
    \small
    \setlength{\tabcolsep}{4pt}
    \begin{tabular}{llcccc}
        \toprule
        \textbf{Setting} & \textbf{Variant} & \textbf{SR (\%)} $\uparrow$ & \textbf{$T_{\text{think}}$ (s)} $\downarrow$ & \textbf{$T_{\text{plan}}$ (s)} $\downarrow$ & \textbf{RPQG (\%)} $\uparrow$ \\
        \midrule
        \multirow{5}{*}{\shortstack[l]{Single-Arm \\ No Door}} 
        & \textbf{Full \algname} & 83.50 $\pm$ 4.43 & 21.23 $\pm$ 0.34 & 0.43 $\pm$ 0.33 & \textbf{---} \\
         
        & w/o \AblationVisionSG & 70.00 $\pm$ 4.32 & 13.55 $\pm$ 0.60 & 0.13 $\pm$ 0.01 & 0.14 $\pm$ 0.38 \\
         
        & w/o \AblationDomain & 86.50 $\pm$ 2.52 & 21.53 $\pm$ 0.47 & 0.09 $\pm$ 0.00 & 8.76 $\pm$ 1.58 \\
         
        & w/o \AblationProblem & 66.00 $\pm$ 2.83 & 38.79 $\pm$ 0.86 & 0.12 $\pm$ 0.00 & 0.88 $\pm$ 0.61 \\
         
        & w/o \AblationCompression & 82.00 $\pm$ 6.32 & 21.64 $\pm$ 0.80 & 4.92 $\pm$ 2.99 & 0.43 $\pm$ 0.57 \\
        \midrule
        \multirow{5}{*}{\shortstack[l]{Single-Arm \\ With Door}} 
        & \textbf{Full \algname} & 84.00 $\pm$ 2.83 & 21.55 $\pm$ 0.78 & 0.42 $\pm$ 0.31 & --- \\
         
        & w/o \AblationVisionSG & 75.00 $\pm$ 3.46 & 13.41 $\pm$ 0.31 & 0.15 $\pm$ 0.02 & -0.01 $\pm$ 0.41 \\
         
        & w/o \AblationDomain & 84.50 $\pm$ 1.91 & 22.07 $\pm$ 0.65 & 0.09 $\pm$ 0.00 & 13.37 $\pm$ 1.54 \\
         
        & w/o \AblationProblem & 31.00 $\pm$ 6.22 & 44.05 $\pm$ 2.17 & 0.10 $\pm$ 0.01 & 0.64 $\pm$ 1.33 \\
         
        & w/o \AblationCompression & 72.00 $\pm$ 4.90 & 22.36 $\pm$ 1.30 & 48.17 $\pm$ 3.01 & 0.34 $\pm$ 0.24 \\
        \midrule
        \multirow{5}{*}{\shortstack[l]{Dual-Arm \\ No Door}} 
        & \textbf{Full \algname} & 82.00 $\pm$ 2.31 & 21.73 $\pm$ 1.83 & 0.66 $\pm$ 0.61 & --- \\
         
        & w/o \AblationVisionSG & 71.50 $\pm$ 4.12 & 13.00 $\pm$ 0.61 & 0.09 $\pm$ 0.00 & -0.32 $\pm$ 0.22 \\
         
        & w/o \AblationDomain & 83.50 $\pm$ 5.00 & 24.78 $\pm$ 1.74 & 0.08 $\pm$ 0.00 & 10.55 $\pm$ 2.43 \\
         
        & w/o \AblationProblem & 64.00 $\pm$ 4.90 & 39.52 $\pm$ 1.16 & 0.09 $\pm$ 0.00 & -0.44 $\pm$ 0.35 \\
         
        & w/o \AblationCompression & 84.00 $\pm$ 3.65 & 23.04 $\pm$ 1.35 & 0.32 $\pm$ 0.09 & -0.03 $\pm$ 0.52 \\
        \midrule
        \multirow{5}{*}{\shortstack[l]{Dual-Arm \\ With Door}} 
        & \textbf{Full \algname} & 85.50 $\pm$ 2.52 & 21.89 $\pm$ 0.91 & 0.11 $\pm$ 0.01 & --- \\
         
        & w/o \AblationVisionSG & 73.50 $\pm$ 4.12 & 13.24 $\pm$ 0.36 & 0.11 $\pm$ 0.00 & -0.30 $\pm$ 0.32 \\
         
        & w/o \AblationDomain & 80.50 $\pm$ 2.52 & 26.98 $\pm$ 1.58 & 0.08 $\pm$ 0.00 & 23.60 $\pm$ 1.67 \\
         
        & w/o \AblationProblem & 31.00 $\pm$ 6.63 & 54.89 $\pm$ 4.18 & 0.09 $\pm$ 0.00 & -0.20 $\pm$ 0.65 \\
         
        & w/o \AblationCompression & 76.50 $\pm$ 3.00 & 26.34 $\pm$ 1.33 & 22.45 $\pm$ 2.29 & -0.13 $\pm$ 0.39 \\
        \bottomrule
    \end{tabular}
\end{table*}

\subsection{Ablation Study}
\label{sec:ablation}

We conduct a comprehensive ablation study to validate the contribution of each module in \algname. Results are reported in Table~\ref{tab:ablation_results}.

\noindent \textbf{Impact of Visual Context (w/o \AblationVisionSG).} 
We replace the visual-topological map with a VLM-generated text-only scene graph. As shown in Table~\ref{tab:ablation_results}, removing raw visual input causes a significant performance drop ($83.75\% \rightarrow 72.5\%$ on average). This confirms that pre-converting images to text creates an \textit{information bottleneck}, losing task-specific cues (e.g., ``emptiness" or spatial relations) that \algname extracts on-demand. Furthermore, the modality gap between text descriptions and the physical environment introduces unavoidable grounding errors.

\noindent \textbf{Impact of Expanded Domain (w/o \AblationDomain).} 
We remove all navigation-related predicates and operators from the holistic PDDL domain, solving for manipulation only and inserting navigation post-hoc. While Success Rate remains comparable, plan quality degrades significantly, evidenced by high positive RPQG values (e.g., $\approx$ 23.6\% in \textit{Dual-Arm with Door}). Lacking a global, cost-aware view, the ablated planner produces redundant traversals. In contrast, our unified mobile-manipulation domain enables the planner to optimize the holistic sequence, such as coordinating door opening with navigation, yielding an $\sim8\%$ gain even in simpler tasks.

\noindent \textbf{Impact of Programmatic Topology Injection (w/o \AblationProblem).} 
Instead of programmatic injection, we prompt the VLM to generate the entire PDDL problem file, including map topology. This causes catastrophic failure in complex settings (SR drops to 31.0\% in Single-Arm/Door) and doubles thinking time ($21.5s \rightarrow 44.0s$). The results indicate that while VLMs excel at semantic reasoning, they struggle to encode large-scale graph structures and syntax. Our decoupled approach mitigates this by offloading structural definitions to reliable algorithms.

\noindent \textbf{Impact of Topology Compression (w/o \AblationCompression).} 
Feeding the full, uncompressed topological map to the planner causes symbolic planning time ($T_{plan}$) to explode (e.g., 0.42s$\rightarrow$48.17s), leading to timeouts (300s) and reduced success rates. This confirms that raw topology exponentially increases the search space for the PDDL solver. Topology compression is therefore essential to maintain computational tractability and ensure valid solutions within real-time limits.

\subsection{Failure Analysis}

Despite the strong performance, \algname fails in approximately 16\% of cases. We categorize the primary failure modes into four types:
\begin{enumerate}
    \item \textbf{PDDL Grounding Errors (7.49\%):} The most common failure arises when the VLM correctly identifies objects but misinterprets their states (e.g., missing necessary predicates in the PDDL \texttt{init} state), leading to unsolvable planning problems.
    \item \textbf{Perception Errors (3.69\%):} The VLM fails to detect necessary objects within the anchored images or misclassifies relationships between objects.
    \item \textbf{Retrieval Errors (3.06\%):} The retrieval module fails to locate all task-relevant nodes, preventing the planner from accessing necessary parts of the map.
    \item \textbf{Instruction Misunderstanding (2.03\%):} The system misinterprets the user intent in the natural language instruction.
\end{enumerate}
These results suggest that while the neuro-symbolic architecture solves the planning consistency problem, future improvements should focus on robust visual state estimation.

\section{CONCLUSION AND LIMITATIONS} 
\textbf{Conclusion.}
We present UniPlan, a vision-language task planning system that tackles the challenge of long-horizon mobile manipulation in large-scale indoor environments. UniPlan unifies scene topology, visuals, and robot capabilities into a holistic PDDL formulation, bridging the gap between semantic VLM reasoning and rigorous symbolic planning. By leveraging a vision-anchored topological map, programmatic domain expansion, and factored problem construction, UniPlan enables scalable and physically consistent planning without requiring expensive mobile demonstrations. Evaluations on diverse human-raised tasks demonstrate that UniPlan substantially outperforms prior VLM-based and modular LLM-PDDL baselines, achieving stronger success rates, higher plan quality, and greater computational efficiency.

\textbf{Limitations.} Despite its strong performance, UniPlan has several limitations. First, our programmatic domain expansion currently models dual arms as logically independent entities. While effective for sequential tasks, this formulation limits simultaneous bimanual coordination, such as lifting heavy trays with both hands. Future work will extend the framework to support complex, synchronized bimanual manipulation. Second, our system assumes a static, fully observable environment during the planning phase, which lacks robustness against object uncertainty, occlusion, or dynamic changes occurring after map construction. Addressing these limitations by enabling real-time updates to the visual-topological map for dynamic adaptation is a critical direction for future research.

\bibliographystyle{plainnat}
\bibliography{references}

@inproceedings{li2024llm-enhanced-sg,
  title={LLM-enhanced Scene Graph Learning for Household Rearrangement},
  author={Li, Wenhao and Yu, Zhiyuan and She, Qijin and Yu, Zhinan and Lan, Yuqing and Zhu, Chenyang and Hu, Ruizhen and Xu, Kai},
  booktitle={SIGGRAPH Asia 2024 Conference Papers},
  pages={1--11},
  year={2024}
}

@article{werby2025keysg,
  title={KeySG: Hierarchical Keyframe-Based 3D Scene Graphs},
  author={Werby, Abdelrhman and Rotondi, Dennis and Scaparro, Fabio and Arras, Kai O},
  journal={arXiv preprint arXiv:2510.01049},
  year={2025}
}

@inproceedings{gu2024conceptgraphs,
  title={Conceptgraphs: Open-vocabulary 3d scene graphs for perception and planning},
  author={Gu, Qiao and Kuwajerwala, Ali and Morin, Sacha and Jatavallabhula, Krishna Murthy and Sen, Bipasha and Agarwal, Aditya and Rivera, Corban and Paul, William and Ellis, Kirsty and Chellappa, Rama and others},
  booktitle={2024 IEEE International Conference on Robotics and Automation (ICRA)},
  pages={5021--5028},
  year={2024},
  organization={IEEE}
}

@article{musumeci2025context-matters,
  title={Context Matters! Relaxing Goals with LLMs for Feasible 3D Scene Planning},
  author={Musumeci, Emanuele and Brienza, Michele and Argenziano, Francesco and Drid, Abdel Hakim and Suriani, Vincenzo and Nardi, Daniele and Bloisi, Domenico D},
  journal={arXiv preprint arXiv:2506.15828},
  year={2025}
}

@article{ekpo2024verigraph,
  title={Verigraph: Scene graphs for execution verifiable robot planning},
  author={Ekpo, Daniel and Levy, Mara and Suri, Saksham and Huynh, Chuong and Shrivastava, Abhinav},
  journal={arXiv preprint arXiv:2411.10446},
  year={2024}
}

@article{herzog2025domain-conditioned-sg,
  title={Domain-Conditioned Scene Graphs for State-Grounded Task Planning},
  author={Herzog, Jonas and Liu, Jiangpin and Wang, Yue},
  journal={arXiv preprint arXiv:2504.06661},
  year={2025}
}

@inproceedings{werby2024hov-sg,
  title={Hierarchical open-vocabulary 3d scene graphs for language-grounded robot navigation},
  author={Werby, Abdelrhman and Huang, Chenguang and B{\"u}chner, Martin and Valada, Abhinav and Burgard, Wolfram},
  booktitle={First Workshop on Vision-Language Models for Navigation and Manipulation at ICRA 2024},
  year={2024}
}

@article{ali2025graphpad,
  title={GraphPad: Inference-Time 3D Scene Graph Updates for Embodied Question Answering},
  author={Ali, Muhammad Qasim and Nair, Saeejith and Wong, Alexander and Cui, Yuchen and Chen, Yuhao},
  journal={arXiv preprint arXiv:2506.01174},
  year={2025}
}

@article{ray2025structured,
  title={Structured interfaces for automated reasoning with 3d scene graphs},
  author={Ray, Aaron and Arkin, Jacob and Biggie, Harel and Fan, Chuchu and Carlone, Luca and Roy, Nicholas},
  journal={arXiv preprint arXiv:2510.16643},
  year={2025}
}

@article{onishchenko2025lookplangraph,
  title={Lookplangraph: Embodied instruction following method with VLM graph augmentation},
  author={Onishchenko, Anatoly O and Kovalev, Alexey K and Panov, Aleksandr I},
  journal={arXiv preprint arXiv:2512.21243},
  year={2025}
}

@article{booker2024embodiedrag,
  title={Embodiedrag: Dynamic 3d scene graph retrieval for efficient and scalable robot task planning},
  author={Booker, Meghan and Byrd, Grayson and Kemp, Bethany and Schmidt, Aurora and Rivera, Corban},
  journal={arXiv preprint arXiv:2410.23968},
  year={2024}
}

@article{hughes2022hydra,
    title={Hydra: A Real-time Spatial Perception System for {3D} Scene Graph Construction and Optimization},
    fullauthor={Nathan Hughes, Yun Chang, and Luca Carlone},
    author={N. Hughes and Y. Chang and L. Carlone},
    booktitle={Robotics: Science and Systems (RSS)},
    year={2022},
}

@inproceedings{chang2023hydra-multi,
  title={Hydra-multi: Collaborative online construction of 3d scene graphs with multi-robot teams},
  author={Chang, Yun and Hughes, Nathan and Ray, Aaron and Carlone, Luca},
  booktitle={2023 IEEE/RSJ International Conference on Intelligent Robots and Systems (IROS)},
  pages={10995--11002},
  year={2023},
  organization={IEEE}
}

@article{pddl,
  title={Pddl| the planning domain definition language},
  author={Aeronautiques, Constructions and Howe, Adele and Knoblock, Craig and McDermott, ISI Drew and Ram, Ashwin and Veloso, Manuela and Weld, Daniel and Sri, David Wilkins and Barrett, Anthony and Christianson, Dave and others},
  journal={Technical Report, Tech. Rep.},
  year={1998}
}

@article{interpret,
  title={Interpret: Interactive predicate learning from language feedback for generalizable task planning},
  author={Han, Muzhi and Zhu, Yifeng and Zhu, Song-Chun and Wu, Ying Nian and Zhu, Yuke},
  journal={arXiv preprint arXiv:2405.19758},
  year={2024}
}

@inproceedings{ada,
  title = {Learning Adaptive Planning Representations with Natural Language Guidance},
  author = {Lio Wong and Jiayuan Mao and Pratyusha Sharma and Zachary S. Siegel and Jiahai Feng and Noa Korneev and Joshua B. Tenenbaum and Jacob Andreas},
  booktitle = {International Conference on Learning Representations (ICLR)},
  year = {2024}
}

@article{lasp,
  title={Language-augmented symbolic planner for open-world task planning},
  author={Chen, Guanqi and Yang, Lei and Jia, Ruixing and Hu, Zhe and Chen, Yizhou and Zhang, Wei and Wang, Wenping and Pan, Jia},
  journal={arXiv preprint arXiv:2407.09792},
  year={2024}
}

@inproceedings{
blade,
title={Learning Compositional Behaviors from Demonstration and Language},
author={Weiyu Liu and Neil Nie and Ruohan Zhang and Jiayuan Mao and Jiajun Wu},
booktitle={8th Annual Conference on Robot Learning},
year={2024},
url={https://openreview.net/forum?id=fR1rCXjCQX}
}

@article{pix2pred,
  title={Predicate Invention from Pixels via Pretrained Vision-Language Models},
  author={Athalye, Ashay and Kumar, Nishanth and Silver, Tom and Liang, Yichao and Lozano-P{\'e}rez, Tom{\'a}s and Kaelbling, Leslie Pack},
  journal={arXiv preprint arXiv:2501.00296},
  year={2024}
}

@article{fastdownward,
  title={The fast downward planning system},
  author={Helmert, Malte},
  journal={Journal of Artificial Intelligence Research},
  volume={26},
  pages={191--246},
  year={2006}
}

@article{pddl2.1,
  title={PDDL2. 1: An extension to PDDL for expressing temporal planning domains},
  author={Fox, Maria and Long, Derek},
  journal={Journal of artificial intelligence research},
  volume={20},
  pages={61--124},
  year={2003}
}

@article{embodied-reasoner,
  title={Embodied-reasoner: Synergizing visual search, reasoning, and action for embodied interactive tasks},
  author={Zhang, Wenqi and Wang, Mengna and Liu, Gangao and Huixin, Xu and Jiang, Yiwei and Shen, Yongliang and Hou, Guiyang and Zheng, Zhe and Zhang, Hang and Li, Xin and others},
  journal={arXiv preprint arXiv:2503.21696},
  year={2025}
}

@inproceedings{2021iros,
  title={Automated generation of robotic planning domains from observations},
  author={Diehl, Maximilian and Paxton, Chris and Ramirez-Amaro, Karinne},
  booktitle={2021 IEEE/RSJ International Conference on Intelligent Robots and Systems (IROS)},
  pages={6732--6738},
  year={2021},
  organization={IEEE}
}

@inproceedings{2025icra,
  title={Automated Planning Domain Inference for Task and Motion Planning},
  author={Huang, Jinbang and Tao, Allen and Marco, Rozilyn and Bogdanovic, Miroslav and Kelly, Jonathan and Shkurti, Florian},
  booktitle={2025 IEEE International Conference on Robotics and Automation (ICRA)},
  pages={12534--12540},
  year={2025},
  organization={IEEE}
}

@inproceedings{unidomain,
    title={UniDomain: Pretraining a Unified {PDDL} Domain from Real-World Demonstrations for Generalizable Robot Task Planning},
    author={Haoming Ye and Yunxiao Xiao and Cewu Lu and Panpan Cai},
    booktitle={The Thirty-ninth Annual Conference on Neural Information Processing Systems},
    year={2025},
}

@inproceedings{
        rana2023sayplan,
        title={SayPlan: Grounding Large Language Models using 3D Scene Graphs for Scalable Task Planning},
        author={Krishan Rana and Jesse Haviland and Sourav Garg and Jad Abou-Chakra and Ian Reid and Niko Suenderhauf},
        booktitle={7th Annual Conference on Robot Learning},
        year={2023},
        url={https://openreview.net/forum?id=wMpOMO0Ss7a}
      }

@article{liu2024delta,
  title={Delta: Decomposed efficient long-term robot task planning using large language models},
  author={Liu, Yuchen and Palmieri, Luigi and Koch, Sebastian and Georgievski, Ilche and Aiello, Marco},
  journal={arXiv preprint arXiv:2404.03275},
  year={2024}
}

@InProceedings{3d-scene-graph,
	title ={3D Scene Graph: A Structure for Unified Semantics, 3D Space, and Camera},
	author = {Iro Armeni and Zhi-Yang He and JunYoung Gwak and Amir R. Zamir and Martin Fischer and Jitendra Malik and Silvio Savarese},
	booktitle = {Proceedings of the IEEE International Conference on Computer Vision},
	year = {2019}
}

@article{ai2thor,
  author={Eric Kolve and Roozbeh Mottaghi and Winson Han and
          Eli VanderBilt and Luca Weihs and Alvaro Herrasti and
          Daniel Gordon and Yuke Zhu and Abhinav Gupta and
          Ali Farhadi},
  title={{AI2-THOR: An Interactive 3D Environment for Visual AI}},
  journal={arXiv},
  year={2017}
}

@article{unified_planning_softwarex2025,
  title = {Unified Planning: Modeling, manipulating and solving AI planning problems in Python},
  author = {Andrea Micheli and Arthur Bit-Monnot and Gabriele R{\"o}ger and Enrico Scala and Alessandro Valentini and Luca Framba and Alberto Rovetta and Alessandro Trapasso and Luigi Bonassi and Alfonso Emilio Gerevini and Luca Iocchi and Felix Ingrand and Uwe Köckemann and Fabio Patrizi and Alessandro Saetti and Ivan Serina and Sebastian Stock},
  journal = {SoftwareX},
  volume = {29},
  pages = {102012},
  year = {2025},
  issn = {2352-7110},
  doi = {https://doi.org/10.1016/j.softx.2024.102012},
  url = {https://www.sciencedirect.com/science/article/pii/S2352711024003820}
}

@misc{anderson2018SPL,
      title={On Evaluation of Embodied Navigation Agents}, 
      author={Peter Anderson and Angel Chang and Devendra Singh Chaplot and Alexey Dosovitskiy and Saurabh Gupta and Vladlen Koltun and Jana Kosecka and Jitendra Malik and Roozbeh Mottaghi and Manolis Savva and Amir R. Zamir},
      year={2018},
      eprint={1807.06757},
      archivePrefix={arXiv},
      primaryClass={cs.AI},
      url={https://arxiv.org/abs/1807.06757}, 
}

\onecolumn
\appendix

\section{Supplementary Material}

\subsection{Algorithm Pseudocode}
\begin{figure}[h]
    \centering
    \includegraphics[width=1\linewidth]{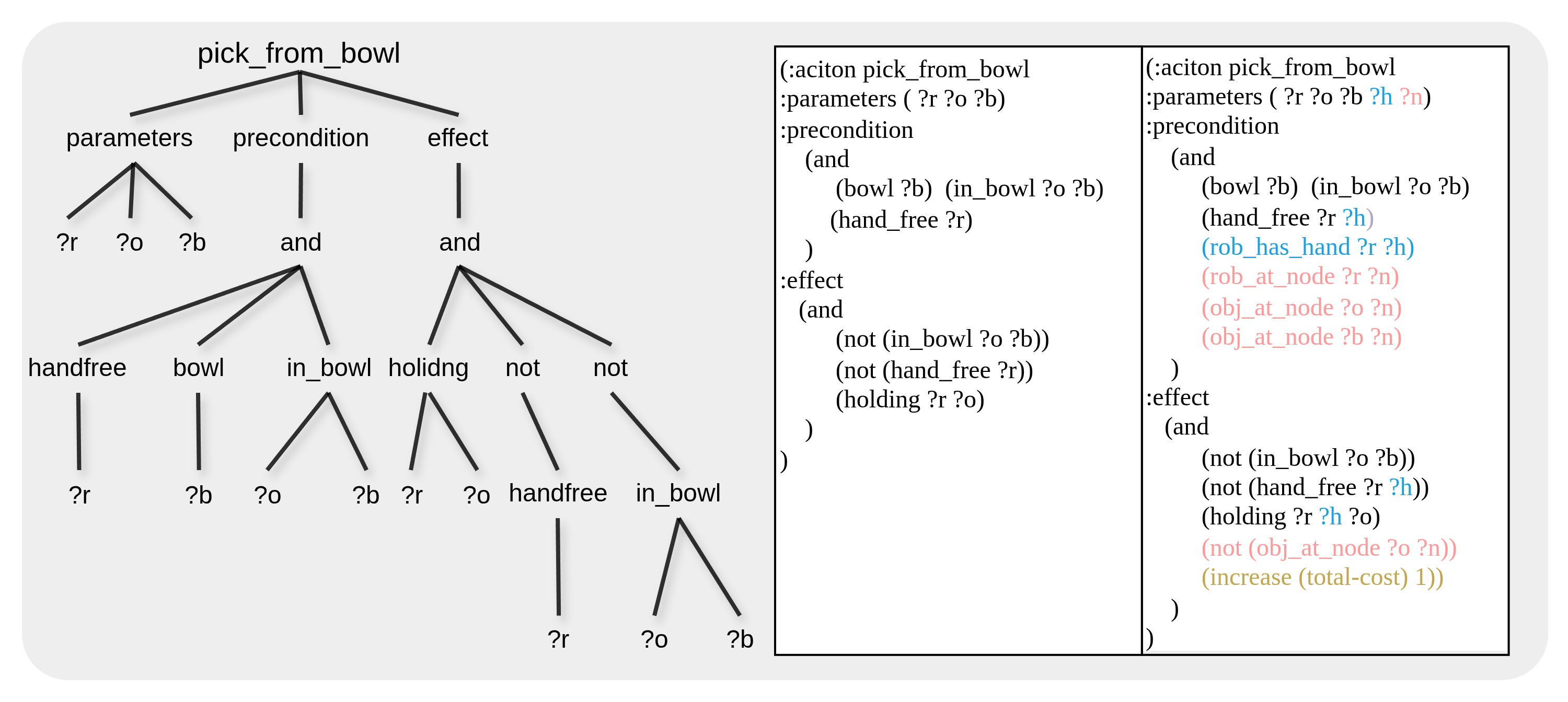}
    \caption{AST-based operator rewriting for domain expansion on \texttt{pick\_from\_bowl}. Left: the parsed AST exposing the syntactic structure of parameters, precondition, and effect. Middle: the original action schema. Right: the expanded schema after rewriting, where blue denotes arm-utility predicates, red denotes topological-state predicates, and orange denotes cost predicates.}
    \label{fig:ast_rewrite}
\end{figure}

We summarize the domain expansion pseudocode in \ref{tab:pseudocode}, covering (1) move and door modeling, (2) bimanual coordination expansion, (3) action cost modeling via travel costs and per-action costs. To make the operator rewriting step concrete, Fig.~\ref{fig:ast_rewrite} visualizes the AST transformation on a representative action: (\texttt{pick\_from\_bowl}).

\ref{tab:problem_construction} describes the problem construction pipeline, which first retrieves task-relevant key nodes from a textual index, compresses the map accordingly, extracts object/state/goal predicates from the associated images, and finally injects robot state, topological and cost constraints before calling a standard PDDL planner.

\newpage
\begin{table}[h]
\small
\centering
\caption{Domain Expansion Pseudocode}
\label{tab:pseudocode}
\setlength{\tabcolsep}{3pt}
\renewcommand{\arraystretch}{1.10}

\begin{tabularx}{\textwidth}{@{}
  >{\hsize=1.20\hsize}Y |
  >{\hsize=0.92\hsize}Y |
  >{\hsize=0.85\hsize}Y @{}}
\toprule
\textbf{(1) Move and Door} & \textbf{(2) Bimanual Coordination} & \textbf{(3) Action Costs} \\
\midrule

\textit{Input:} Base Domain $D$, Anchors: \textit{hand\_free}, \textit{holding}\par
\smallskip
\[
D \leftarrow D \cup
\left\{
\begin{aligned}
&(rob\_at\_node\ ?r\ ?n),\\
&(obj\_at\_node\ ?o\ ?n)
\end{aligned}
\right\}
\]
\smallskip
\textbf{For} operator in $D$:\par
\smallskip
\I operator.parameters.add(?n)\par
\smallskip
\I operator.precondition.add((rob\_at\_node ?r ?n))\par
\smallskip
\I \textbf{For} parameter ?o in operator.params $\setminus \{?r, ?n\}$:\par
\smallskip
\II \textbf{if} ?o $\notin$ precond.(holding ?r ?o):\par
\smallskip
\III operator.precondition.add((obj\_at\_node ?o ?n))\par
\smallskip
\III \textbf{if} ?o $\in$ effect.(holding ?r ?o):\par
\smallskip
\IV operator.effect.add($\neg$(obj\_at\_node ?o ?n))\par
\smallskip
\III \textbf{if} ?o $\in$ effect.($\neg$(holding ?r ?o)):\par
\smallskip
\IV operator.effect.add((obj\_at\_node ?o ?n))\par
\smallskip
\[
D \leftarrow D \cup
\left\{
\begin{aligned}
&(has\_door\ ?n1\ ?n2),\\
&(connected\ ?n1\ ?n2),\\
&\texttt{move\_robot},\\
&\texttt{open\_door}
\end{aligned}
\right\}
\]
&
\textit{Input:} Base Domain $D$,\par
\smallskip
Anchors: \textit{hand\_free}, \textit{holding}\par
\medskip
(hand\_free ?r) $\rightarrow$ (hand\_free ?r ?h)\par
\smallskip
(holding ?r ?o) $\rightarrow$ (holding ?r ?h ?o)\par
\smallskip
$D \leftarrow D \cup \{(rob\_has\_hand\ ?r\ ?h)\}$\par
\medskip
\textbf{For} operator in $D$:\par
\smallskip
\I \textbf{if} (hand\_free ?r) $\in$ operator:\par
\smallskip
\II operator.parameters.add(?h)\par
\smallskip
\II (hand\_free ?r) $\rightarrow$ (hand\_free ?r ?h)\par
\smallskip
\I \textbf{if} (holding ?r ?o) $\in$ operator:\par
\smallskip
\II operator.parameters.add(?h)\par
\smallskip
\II (holding ?r ?o) $\rightarrow$ (holding ?r ?h ?o)\par
&
\textit{Input:} Base Domain $D$\par
\medskip
$D \leftarrow D \cup \{(travel\_cost\ ?n1\ ?n2)\}$\par
\medskip
\texttt{move\_robot}.effect.add(\par
\smallskip
\I (increase (total-cost) (travel\_cost \par
\smallskip
?n1 ?n2))\par
\smallskip
)\par
\medskip
\textbf{For} operator in $D \setminus \{\texttt{move\_robot}\}$:\par
\smallskip
\I operator.effect.add(\par
\smallskip
\II (increase (total-cost) 1)\par
\smallskip
)\par
\\
\bottomrule
\end{tabularx}
\end{table}

\begin{table}[h]
\small
\centering
\caption{Problem Construction Pseudocode}
\label{tab:problem_construction}
\begin{tabularx}{\textwidth}{@{} X @{}}
\toprule
\textbf{Algorithm: Problem Construction} \\
\midrule
\begin{flushleft}
\textbf{Input:} Full Map $M$, Expanded Domain $D$, Task Instruction $T$, Textual Index $I$ \\
\addvspace{8pt}
$key\_nodes = \text{LLM}(T, I)$ \\
$compressed\_map = \text{Compress}(M, key\_nodes \cup \{robot\_node\})$ \\
$Objects, Init, Goal = \emptyset$ \\
\addvspace{5pt}
\textit{// Robot Initialization} \\
$Objects = Objects \cup \{robot, arms\}$ \\
$Init = Init \cup \{(robot\_has\_arm, robot, arms)\}$ \\
\addvspace{5pt}
\textit{// Object State Extraction via VLM} \\
$Objects, Init, Goal = \text{VLM}(T, D, key\_nodes)$ \\
\textbf{For} $key\_node$ in $key\_nodes$: \\
\quad \textbf{For} $object$ in $key\_node$: \\
\quad\quad $Init = Init \cup \{(object\_at\_node, key\_node)\}$ \\
\addvspace{5pt}
\textit{// Map Knowledge Injection} \\
$Objects, Init = \text{Injection}(compressed\_map)$ \\
\addvspace{8pt}
$P = Objects \cup Init \cup Goal$ \\
$Plan = \text{PDDL\_Planner}(D, P)$ \\
\end{flushleft} \\
\bottomrule
\end{tabularx}
\end{table}

\newpage
\subsection{Full Task List}
We collected 50 real-world tasks from humans and group them by difficulty (Simple/Moderate/Compositional) based on asset nodes and horizon length. The complete task list is as follows.
\begin{tasklistbox}
    \taskseparator{blue}{Simple Tasks (1--10)}
    \task{1}{task 1: Place a block on the table in office 613.}
    \task{2}{task 2: Place a bottle of water from the storage room on the table in office 612.}
    \task{3}{task 3: Place the mango into the fridge.}
    \task{4}{task 4: There is an apple in the fridge. Place it on the table in office 604.}
    \task{5}{task 5: There is a pink block in the blue drawer in the storage room. Place it on the table in office 606.}
    \task{6}{task 6: Place the black marker in office 610 on table 2 in office 608.}
    \task{7}{task 7: Hang the hat on the drying rack.}
    \task{8}{task 8: Open the curtain.}
    \task{9}{task 9: Turn on the lamp.}
    \task{10}{task 10: Fill a kettle with water and place it on kitchen table 1.}

    \taskseparator{orange}{Moderate Tasks (11--40)}
    \task{11}{task 11: There is an apple in the fridge. Wash it and put it on the middle plate in the kitchen.}
    \task{12}{task 12: Give me a cup of coffee. Place it on the table in office 612.}
    \task{13}{task 13: Give me a cup of boiled water. Place it on the table in office 608.}
    \task{14}{task 14: There is a bottle of milk in the fridge. Please give me a cup of milk. Place it on the table in office 601.}
    \task{15}{task 15: There is a bottle of milk in the fridge. Please give me a cup of heated milk. Place it on the table in office 605.}
    \task{16}{task 16: Water the flower using a kettle.}
    \task{17}{task 17: Place the empty bottle on the table in office 613 into the trash bin.}
    \task{18}{task 18: Put all fruits in the kitchen into the fridge.}
    \task{19}{task 19: There is a tomato in the fridge. Cut it on the cutting board leaning against the wall.}
    \task{20}{task 20: Scoop rice from the pot and pour it into the yellow bowl on kitchen table 1.}
    \task{21}{task 21: Stir the bowls using some tools on kitchen table 2.}
    \task{22}{task 22: Wash the cloth on the couch and hang it on the drying rack.}
    \task{23}{task 23: Fold the cloth on the couch and put it on the table in office 612.}
    \task{24}{task 24: Pour water from the leftmost bottle in office 613 into the rightmost cup in office 602.}
    \task{25}{task 25: The tables in office 608 are dirty. Wipe them clean.}
    \task{26}{task 26: Put my hat in office 604 into the orange drawer in the storage room.}
    \task{27}{task 27: I am going to work. Give me a mouse and put it on the table in office 612.}
    \task{28}{task 28: A kid is going to play with blocks in office 604. He prefers purple, red, blue, green from top to bottom on the table. Please help him to set up.}
    \task{29}{task 29: A kid played with blocks in office 604. Please help him to put all blocks into the black drawer in the storage room.}
    \task{30}{task 30: I need to work on my laptop. Please help me to turn it on in office 608.}
    \task{31}{task 31: My work is done in office 601. Please help me close the laptop, and put the mouse on the table in office 608.}
    \task{32}{task 32: I will go to France next week. Please give me a related book and place it on the table in office 614. Also, help me to turn the lamp on.}
    \task{33}{task 33: It's a nice day today. Open the windows to let some fresh air in.}
    \task{34}{task 34: Place the rightmost marker in office 610 into the black holder on the meeting table.}
    \task{35}{task 35: There is a piece of bread in the fridge. Toast it and put it on the rightmost plate in the kitchen.}
    \task{36}{task 36: Wipe the blackboard clean.}
    \task{37}{task 37: Fill a kettle with water, pour it into the ice maker, and make some ice cubes.}
    \task{38}{task 38: There is a new battery in the yellow drawer in the storage room. Replace the old one in the remote in office 614, and place the old one into the trash bin.}
    \task{39}{task 39: I want to learn about toasting bread. Give me a related book and place it beside the toaster.}
    \task{40}{task 40: I will meet some friends. Prepare two bottles of water from storage table 2 and place them on the meeting table.}

    \taskseparator{red}{Compositional Tasks (41--50)}
    \task{41}{task 41: Prepare two cups of coffee and place them on the meeting table.}
    \task{42}{task 42: Place the rightmost marker in office 610 into the white holder on the meeting table. Place the blue marker in office 610 into the white holder in office 611.}
    \task{43}{task 43: There are a tomato and a piece of bread in the fridge. Toast the bread and put it on the rightmost plate in the kitchen. Put the tomato on the leftmost plate.}
    \task{44}{task 44: Place the bottles with the white lid and black lid on the table in office 613 into the trash bin. And put my hat into the blue drawer in the storage room.}
    \task{45}{task 45: Help me turn the laptop on in office 608 and put a mouse beside it. And open the curtain.}
    \task{46}{task 46: Wash the cloth on the couch and hang it on the drying rack. And hang the hat on the drying rack.}
    \task{47}{task 47: Scoop rice from the pot and pour it into the bowl on kitchen table 1. And stir the bowls on kitchen table 2.}
    \task{48}{task 48: There are a tomato and a milk bottle in the fridge. Put all fruits in the kitchen into the fridge. Place the tomato and milk bottle from the fridge onto kitchen table 2.}
    \task{49}{task 49: Wash the apple in the fridge, put it on the table in office 612, fold the cloth from the couch, and place it on the same table.}
    \task{50}{task 50: Make me a cup of boiled water. Place it on the table in office 608. And then water the flower.}
\end{tasklistbox}

\newpage
\subsection{Visual-Topological Map Example}
\begin{figure*}[h]
    \centering
    \includegraphics[width=\textwidth]{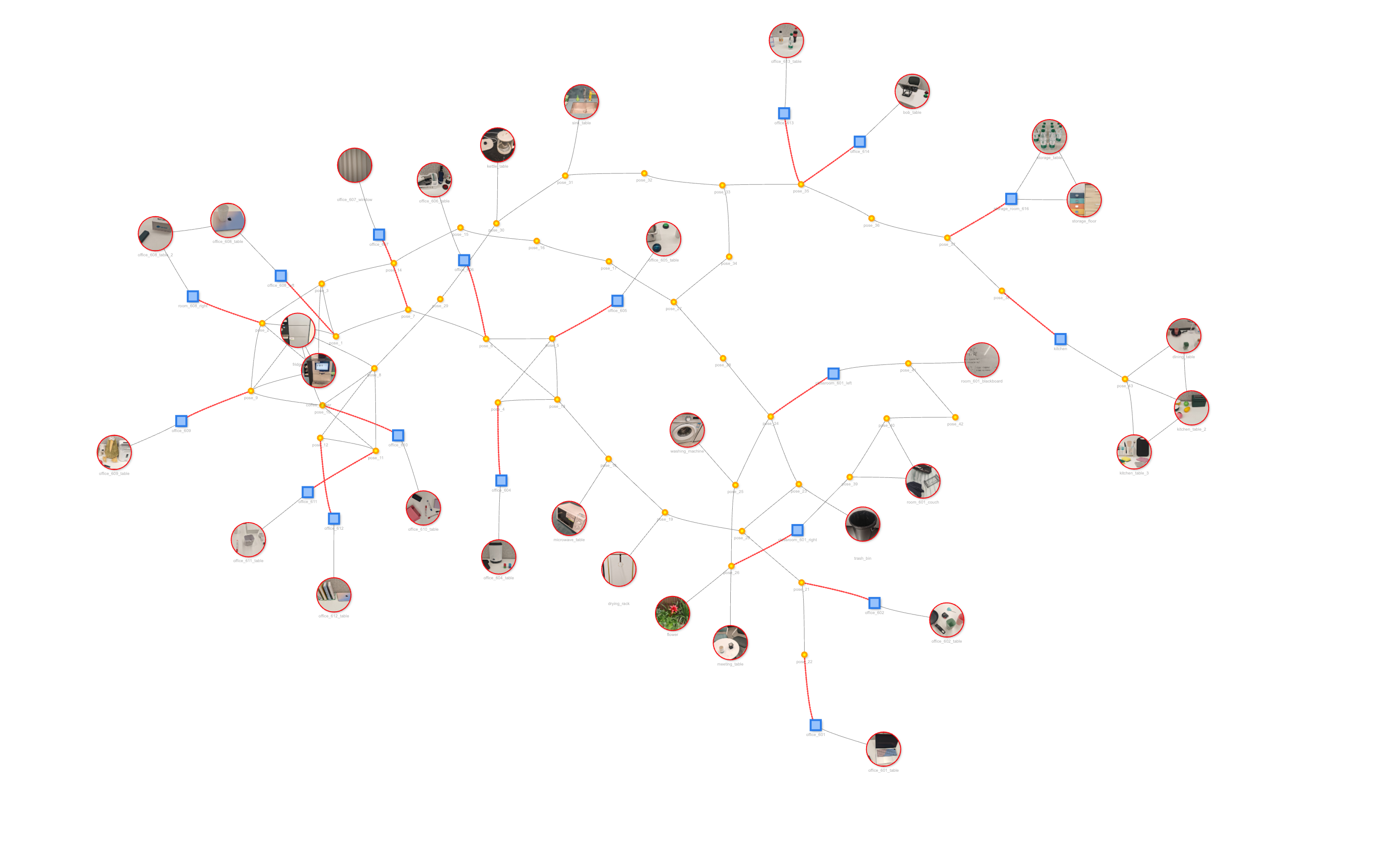} 
    \caption{Visual-topological map example, where yellow nodes represent pose nodes, blue nodes represent room nodes, and images denote asset nodes.}
    \label{fig:topomap}
\end{figure*}

Fugure \ref{fig:topomap} (an illustrative visual-topological map instance) used in our experiments is provided to clarify the underlying state abstraction. The example highlights the node taxonomy (pose/room/asset), the connectivity structure that supports navigation under door constraints, and the image-anchored assets that preserve local visual evidence for grounding.

\newpage
\subsection{Textual Index Example}
The following list provides the textual descriptions of the environment objects and their initial states, as indexed in the visual-topological map.

\begin{indexbox}
    \indexitem{fridge}{one silver three-door refrigerator with stickers, next to one black water dispenser.}
    \indexitem{drying\_rack}{drying rack with one hanging bar and one metal ring.}
    \indexitem{coffee\_maker}{one automatic coffee machine with touchscreen, one bean hopper of coffee beans, one drip tray, and one sign on a countertop.}
    \indexitem{office\_614\_table}{office 614 table with one desk lamp, one remote control, and one wall power outlet with a plugged-in power adapter and cable.}
    \indexitem{kettle\_table}{kettle table with one glass electric kettle, one kettle base, and power cords.}
    \indexitem{washing\_machine}{one front-loading washing machine with control knob and digital display at the washing machine location.}
    \indexitem{room\_601\_couch}{room 601 couch with one racket bag, one cloth, and one metal rack.}
    \indexitem{room\_601\_blackboard}{room 601 blackboard with math equations, vectors, and Jacobian matrix written in black marker.}
    \indexitem{microwave\_table}{microwave table with one microwave oven, one wall power outlet, one power cord, and one sign.}
    \indexitem{sink\_table}{sink table with one sink basin, one faucet, one soap dispenser bottle, one cleaning spray bottle, stacked towels, cloth rags, a sponge brush, and a wall sign.}
    \indexitem{office\_601\_table}{office 601 table: one open laptop and one wireless mouse.}
    \indexitem{office\_602\_table}{office 602 table with one frying pan, one tape measure, one pair of scissors, and two cups/containers.}
    \indexitem{office\_605\_table}{office 605 table: one stainless steel container, one green push button, one blue push button, and one white cup.}
    \indexitem{office\_608\_table}{office 608 table has one closed MacBook laptop with a pink-blue cover and one white plastic cup.}
    \indexitem{office\_608\_table\_2}{office 608 table 2 with a black case and a white tissue box.}
    \indexitem{kitchen\_table\_2}{kitchen table 2 with one two-slot toaster, two bowls, one green apple, one yellow mango, and one pink peach.}
    \indexitem{flower}{one red flower and several green foliage plants in the flower planter bed.}
    \indexitem{storage\_table}{storage table with seven water bottles and two wall power outlets.}
    \indexitem{office\_610\_table}{office 610 table with one eyeglasses case, two sticky note stacks (pink and purple), and four whiteboard markers (1 blue, 1 black, 2 red).}
    \indexitem{meeting\_table}{meeting table with one paper cup, one mesh pen holder, and two office chairs.}
    \indexitem{storage\_floor}{storage floor with a four-drawer plastic organizer, a three-drawer wooden cabinet, and a cardboard box.}
    \indexitem{kitchen\_table\_1}{a dish rack with one yellow bowl and two plates, and one white slow cooker with glass lid.}
    \indexitem{office\_611\_table}{office 611 table with one paper cup, one cardboard box, one plastic bin, two red markers, one clear plastic handle device, and one black clip tool.}
    \indexitem{office\_613\_table}{office 613 table with four bottles: two water bottles, one clear drink bottle, and one soda bottle.}
    \indexitem{office\_612\_table}{office 612 table with a mesh pen holder (glue stick, pen, tape dispenser), one green book, a mint bookend holding four books, and one closed laptop.}
    \indexitem{trash\_bin}{one black trash bin with a trash bag liner and metal handle (trash bin).}
    \indexitem{office\_604\_table}{office 604 table with one black cap, one white 8L bin, and two stacked block towers (red on green, blue on purple).}
    \indexitem{office\_607\_window}{office 607 window with closed vertical blinds and top window frame.}
    \indexitem{office\_606\_table}{office 606 table with one plastic basket holding a tissue box, stapler, organizers, cables and a cleaning brush, plus one folded umbrella, one pen cup with markers, and one red round button light.}
    \indexitem{kitchen\_table\_3}{kitchen table 3 with a utensil holder, two scissors, a knife, a spatula, a spoon, a black cutting board, two sponges, and two cleaning cloths.}
    \indexitem{office\_609\_table}{office 609 table with three stuffed animals (sloth plush, white dog plush, green bird plush), one ice maker machine, and one red frying pan.}
\end{indexbox}

\newpage
\subsection{Base Domain and Expanded Domain}
A selection of representative PDDL operators from the base domain is presented below

\begin{pddlbox}
    \pddlline{(:action fold\_on\_table}
    \pddlline{\quad :parameters (?r ?o ?t)}
    \pddlline{\quad :precondition (and (table ?t) (hand\_free ?r) (unfolded ?o) (on\_table ?o ?t))}
    \pddlline{\quad :effect (and (folded ?o) (not (unfolded ?o)))}
    \pddlline{)}
    \addvspace{8pt}

    \pddlline{(:action put\_in\_bin}
    \pddlline{\quad :parameters (?r ?o ?b)}
    \pddlline{\quad :precondition (and (holding ?r ?o) (bin ?b))}
    \pddlline{\quad :effect (and (in\_bin ?o ?b) (hand\_free ?r) (not (holding ?r ?o)))}
    \pddlline{)}
    \addvspace{8pt}

    \pddlline{(:action wipe\_table}
    \pddlline{\quad :parameters (?r ?o ?t)}
    \pddlline{\quad :precondition (and (holding ?r ?o) (can\_wipe\_table ?o) (table ?t) (not (wiped ?t)))}
    \pddlline{\quad :effect (wiped ?t)}
    \pddlline{)}
    \addvspace{8pt}

    \pddlline{(:action turn\_on\_faucet}
    \pddlline{\quad :parameters (?r ?t)}
    \pddlline{\quad :precondition (and (hand\_free ?r) (tap ?t) (not (is\_on ?t)))}
    \pddlline{\quad :effect (and (is\_on ?t)))}
    \pddlline{)}
    \addvspace{8pt}

    \pddlline{(:action wash\_under\_faucet}
    \pddlline{\quad :parameters (?r ?o ?t)}
    \pddlline{\quad :precondition (and (holding ?r ?o) (tap ?t) (is\_on ?t))}
    \pddlline{\quad :effect (and (washed ?o)))}
    \pddlline{)}
    \addvspace{8pt}

    \pddlline{(:action turn\_off\_faucet}
    \pddlline{\quad :parameters (?r ?t)}
    \pddlline{\quad :precondition (and (hand\_free ?r) (tap ?t) (is\_on ?t))}
    \pddlline{\quad :effect (and (not (is\_on ?t))))}
    \pddlline{)}
    \addvspace{8pt}

    \pddlline{(:action place\_on\_coffee\_maker}
    \pddlline{\quad :parameters (?r ?c ?m)}
    \pddlline{\quad :precondition (and (holding ?r ?c) (cup ?c) (coffee\_maker ?m))}
    \pddlline{\quad :effect (and (on\_coffee\_maker ?c ?m) (hand\_free ?r) (not (holding ?r ?c))))}
    \pddlline{)}
    \addvspace{8pt}

    \pddlline{(:action pick\_from\_coffee\_maker}
    \pddlline{\quad :parameters (?r ?c ?m)}
    \pddlline{\quad :precondition (and (hand\_free ?r) (cup ?c) (coffee\_maker ?m) (on\_coffee\_maker ?c ?m))}
    \pddlline{\quad :effect (and (not (on\_coffee\_maker ?c ?m)) (not (hand\_free ?r)) (holding ?r ?c)))}
    \pddlline{)}
    \addvspace{8pt}

    \pddlline{(:action fill\_coffee\_into\_cup}
    \pddlline{\quad :parameters (?r ?c ?m)}
    \pddlline{\quad :precondition (and (hand\_free ?r) (cup ?c) (coffee\_maker ?m) (on\_coffee\_maker ?c ?m))}
    \pddlline{\quad :effect (and (filled\_coffee ?c)))}
    \pddlline{)}
    \addvspace{8pt}

    \pddlline{(:action open\_laptop}
    \pddlline{\quad :parameters (?r ?l)}
    \pddlline{\quad :precondition (and (hand\_free ?r) (laptop ?l) (not (is\_open ?l)) (not (covered ?l)))}
    \pddlline{\quad :effect (and (is\_open ?l)))}
    \pddlline{)}
    \addvspace{8pt}

    \pddlline{(:action close\_laptop}
    \pddlline{\quad :parameters (?r ?l)}
    \pddlline{\quad :precondition (and (hand\_free ?r) (laptop ?l) (is\_open ?l))}
    \pddlline{\quad :effect (and (not (is\_open ?l))))}
    \pddlline{)}
    \addvspace{8pt}

    \pddlline{(:action turn\_on\_laptop}
    \pddlline{\quad :parameters (?r ?l)}
    \pddlline{\quad :precondition (and (hand\_free ?r) (laptop ?l) (is\_open ?l))}
    \pddlline{\quad :effect (and (is\_on ?l)))}
    \pddlline{)}
    \addvspace{8pt}

    \pddlline{(:action turn\_on\_lamp}
    \pddlline{\quad :parameters (?r ?l)}
    \pddlline{\quad :precondition (and (hand\_free ?r) (lamp ?l))}
    \pddlline{\quad :effect (and (is\_on ?l)))}
    \pddlline{)}
    \addvspace{8pt}

    \pddlline{(:action close\_window}
    \pddlline{\quad :parameters (?r ?w ?c)}
    \pddlline{\quad :precondition (and (hand\_free ?r) (window ?w) (curtain ?c) (link ?w ?c) (is\_open ?w) (is\_open ?c))}
    \pddlline{\quad :effect (and (not (is\_open ?w))))}
    \pddlline{)}
    \addvspace{8pt}

    \pddlline{(:action open\_window}
    \pddlline{\quad :parameters (?r ?w ?c)}
    \pddlline{\quad :precondition (and (hand\_free ?r) (window ?w) (curtain ?c) (link ?w ?c) (not (is\_open ?w)) (is\_open ?c))}
    \pddlline{\quad :effect (and (is\_open ?w)))}
    \pddlline{)}
    \addvspace{8pt}

    \pddlline{(:action open\_curtain}
    \pddlline{\quad :parameters (?r ?c)}
    \pddlline{\quad :precondition (and (hand\_free ?r) (curtain ?c) (not (is\_open ?c)))}
    \pddlline{\quad :effect (and (is\_open ?c)))}
    \pddlline{)}
    \addvspace{8pt}

    \pddlline{(:action close\_curtain}
    \pddlline{\quad :parameters (?r ?c)}
    \pddlline{\quad :precondition (and (hand\_free ?r) (curtain ?c) (is\_open ?c))}
    \pddlline{\quad :effect (and (not (is\_open ?c))))}
    \pddlline{)}
    \addvspace{8pt}

    \pddlline{(:action wipe\_blackboard}
    \pddlline{\quad :parameters (?r ?o ?b)}
    \pddlline{\quad :precondition (and (holding ?r ?o) (can\_wipe\_blackboard ?o) (blackboard ?b))}
    \pddlline{\quad :effect (and (wiped ?b)))}
    \pddlline{)}
    \addvspace{8pt}

    \pddlline{(:action open\_remote}
    \pddlline{\quad :parameters (?r ?rm)}
    \pddlline{\quad :precondition (and (hand\_free ?r) (remote ?rm) (not (is\_open ?rm)))}
    \pddlline{\quad :effect (and (is\_open ?rm)))}
    \pddlline{)}
    \addvspace{8pt}

    \pddlline{(:action close\_remote}
    \pddlline{\quad :parameters (?r ?rm)}
    \pddlline{\quad :precondition (and (hand\_free ?r) (remote ?rm) (is\_open ?rm))}
    \pddlline{\quad :effect (and (not (is\_open ?rm))))}
    \pddlline{)}
    \addvspace{8pt}

    \pddlline{(:action place\_in\_remote}
    \pddlline{\quad :parameters (?r ?b ?rm)}
    \pddlline{\quad :precondition (and (holding ?r ?b) (remote ?rm) (battery ?b) (is\_open ?rm) (not (has\_battery ?rm)))}
    \pddlline{\quad :effect (and (in\_remote ?b ?rm) (hand\_free ?r) (not (holding ?r ?b)) (has\_battery ?rm)))}
    \pddlline{)}
    \addvspace{8pt}

    \pddlline{(:action pick\_from\_remote}
    \pddlline{\quad :parameters (?r ?b ?rm)}
    \pddlline{\quad :precondition (and (hand\_free ?r) (remote ?rm) (battery ?b) (in\_remote ?b ?rm) (has\_battery ?rm) (is\_open ?rm))}
    \pddlline{\quad :effect (and (not (in\_remote ?b ?rm)) (not (hand\_free ?r)) (holding ?r ?b) (not (has\_battery ?rm)))}
    \pddlline{)}
\end{pddlbox}

The expanded domain incorporates mobile manipulation with spatial node constraints, door traversal, bimanual manipulation, and action cost modeling. A selection of the resulting refined operators, which define the enhanced robot capabilities, is presented below.

\begin{pddlbox}
    \pddlline{(:functions}
    \pddlline{\quad (travel\_cost ?n1 ?n2)}
    \pddlline{\quad (total-cost)}
    \pddlline{)}
    \addvspace{8pt}

    \pddlline{(:action fold\_on\_table}
    \pddlline{\quad :parameters (?r ?hand ?o ?t ?node)}
    \pddlline{\quad :precondition (and (robot\_has\_hand ?r ?hand) (table ?t) (hand\_free ?r ?hand) (unfolded ?o) (on\_table ?o ?t) (robot\_at\_node ?r ?node) (object\_at\_node ?o ?node) (object\_at\_node ?t ?node))}
    \pddlline{\quad :effect (and (folded ?o) (not (unfolded ?o)) (increase (total-cost) 1))}
    \pddlline{)}
    \addvspace{8pt}

    \pddlline{(:action put\_in\_bin}
    \pddlline{\quad :parameters (?r ?hand ?o ?b ?node)}
    \pddlline{\quad :precondition (and (robot\_has\_hand ?r ?hand) (holding ?r ?hand ?o) (bin ?b) (robot\_at\_node ?r ?node) (object\_at\_node ?b ?node))}
    \pddlline{\quad :effect (and (in\_bin ?o ?b) (hand\_free ?r ?hand) (not (holding ?r ?hand ?o)) (object\_at\_node ?o ?node) (increase (total-cost) 1))}
    \pddlline{)}
    \addvspace{8pt}

    \pddlline{(:action wipe\_table}
    \pddlline{\quad :parameters (?r ?hand ?o ?t ?node)}
    \pddlline{\quad :precondition (and (robot\_has\_hand ?r ?hand) (holding ?r ?hand ?o) (can\_wipe\_table ?o) (table ?t) (not (wiped ?t)) (robot\_at\_node ?r ?node) (object\_at\_node ?t ?node))}
    \pddlline{\quad :effect (and (wiped ?t) (increase (total-cost) 1))}
    \pddlline{)}
    \addvspace{8pt}

    \pddlline{(:action turn\_on\_faucet}
    \pddlline{\quad :parameters (?r ?hand ?t ?node)}
    \pddlline{\quad :precondition (and (robot\_has\_hand ?r ?hand) (hand\_free ?r ?hand) (tap ?t) (not (is\_on ?t)) (robot\_at\_node ?r ?node) (object\_at\_node ?t ?node))}
    \pddlline{\quad :effect (and (is\_on ?t) (increase (total-cost) 1))}
    \pddlline{)}
    \addvspace{8pt}

    \pddlline{(:action wash\_under\_faucet}
    \pddlline{\quad :parameters (?r ?hand ?o ?t ?node)}
    \pddlline{\quad :precondition (and (robot\_has\_hand ?r ?hand) (holding ?r ?hand ?o) (tap ?t) (is\_on ?t) (robot\_at\_node ?r ?node) (object\_at\_node ?t ?node))}
    \pddlline{\quad :effect (and (washed ?o) (increase (total-cost) 1))}
    \pddlline{)}
    \addvspace{8pt}

    \pddlline{(:action turn\_off\_faucet}
    \pddlline{\quad :parameters (?r ?hand ?t ?node)}
    \pddlline{\quad :precondition (and (robot\_has\_hand ?r ?hand) (hand\_free ?r ?hand) (tap ?t) (is\_on ?t) (robot\_at\_node ?r ?node) (object\_at\_node ?t ?node))}
    \pddlline{\quad :effect (and (not (is\_on ?t)) (increase (total-cost) 1))}
    \pddlline{)}
    \addvspace{8pt}

    \pddlline{(:action place\_on\_coffee\_maker}
    \pddlline{\quad :parameters (?r ?hand ?c ?m ?node)}
    \pddlline{\quad :precondition (and (robot\_has\_hand ?r ?hand) (holding ?r ?hand ?c) (cup ?c) (coffee\_maker ?m) (robot\_at\_node ?r ?node) (object\_at\_node ?m ?node))}
    \pddlline{\quad :effect (and (on\_coffee\_maker ?c ?m) (hand\_free ?r ?hand) (not (holding ?r ?hand ?c)) (object\_at\_node ?c ?node) (increase (total-cost) 1))}
    \pddlline{)}
    \addvspace{8pt}

    \pddlline{(:action pick\_from\_coffee\_maker}
    \pddlline{\quad :parameters (?r ?hand ?c ?m ?node)}
    \pddlline{\quad :precondition (and (robot\_has\_hand ?r ?hand) (hand\_free ?r ?hand) (cup ?c) (coffee\_maker ?m) (on\_coffee\_maker ?c ?m) (robot\_at\_node ?r ?node) (object\_at\_node ?c ?node) (object\_at\_node ?m ?node))}
    \pddlline{\quad :effect (and (not (on\_coffee\_maker ?c ?m)) (not (hand\_free ?r ?hand)) (holding ?r ?hand ?c) (not (object\_at\_node ?c ?node)) (increase (total-cost) 1))}
    \pddlline{)}
    \addvspace{8pt}

    \pddlline{(:action fill\_coffee\_into\_cup}
    \pddlline{\quad :parameters (?r ?hand ?c ?m ?node)}
    \pddlline{\quad :precondition (and (robot\_has\_hand ?r ?hand) (hand\_free ?r ?hand) (cup ?c) (coffee\_maker ?m) (on\_coffee\_maker ?c ?m) (robot\_at\_node ?r ?node) (object\_at\_node ?c ?node) (object\_at\_node ?m ?node))}
    \pddlline{\quad :effect (and (filled\_coffee ?c) (increase (total-cost) 1))}
    \pddlline{)}
    \addvspace{8pt}

    \pddlline{(:action open\_laptop}
    \pddlline{\quad :parameters (?r ?hand ?l ?node)}
    \pddlline{\quad :precondition (and (robot\_has\_hand ?r ?hand) (hand\_free ?r ?hand) (laptop ?l) (not (is\_open ?l)) (not (covered ?l)) (robot\_at\_node ?r ?node) (object\_at\_node ?l ?node))}
    \pddlline{\quad :effect (and (is\_open ?l) (increase (total-cost) 1))}
    \pddlline{)}
    \addvspace{8pt}

    \pddlline{(:action close\_laptop}
    \pddlline{\quad :parameters (?r ?hand ?l ?node)}
    \pddlline{\quad :precondition (and (robot\_has\_hand ?r ?hand) (hand\_free ?r ?hand) (laptop ?l) (is\_open ?l) (robot\_at\_node ?r ?node) (object\_at\_node ?l ?node))}
    \pddlline{\quad :effect (and (not (is\_open ?l)) (increase (total-cost) 1))}
    \pddlline{)}
    \addvspace{8pt}

    \pddlline{(:action turn\_on\_laptop}
    \pddlline{\quad :parameters (?r ?hand ?l ?node)}
    \pddlline{\quad :precondition (and (robot\_has\_hand ?r ?hand) (hand\_free ?r ?hand) (laptop ?l) (is\_open ?l) (robot\_at\_node ?r ?node) (object\_at\_node ?l ?node))}
    \pddlline{\quad :effect (and (is\_on ?l) (increase (total-cost) 1))}
    \pddlline{)}
    \addvspace{8pt}

    \pddlline{(:action turn\_on\_lamp}
    \pddlline{\quad :parameters (?r ?hand ?l ?node)}
    \pddlline{\quad :precondition (and (robot\_has\_hand ?r ?hand) (hand\_free ?r ?hand) (lamp ?l) (robot\_at\_node ?r ?node) (object\_at\_node ?l ?node))}
    \pddlline{\quad :effect (and (is\_on ?l) (increase (total-cost) 1))}
    \pddlline{)}
    \addvspace{8pt}

    \pddlline{(:action close\_window}
    \pddlline{\quad :parameters (?r ?hand ?w ?c ?node)}
    \pddlline{\quad :precondition (and (robot\_has\_hand ?r ?hand) (hand\_free ?r ?hand) (window ?w) (curtain ?c) (link ?w ?c) (is\_open ?w) (is\_open ?c) (robot\_at\_node ?r ?node) (object\_at\_node ?w ?node) (object\_at\_node ?c ?node))}
    \pddlline{\quad :effect (and (not (is\_open ?w)) (increase (total-cost) 1))}
    \pddlline{)}
    \addvspace{8pt}

    \pddlline{(:action open\_window}
    \pddlline{\quad :parameters (?r ?hand ?w ?c ?node)}
    \pddlline{\quad :precondition (and (robot\_has\_hand ?r ?hand) (hand\_free ?r ?hand) (window ?w) (curtain ?c) (link ?w ?c) (not (is\_open ?w)) (is\_open ?c) (robot\_at\_node ?r ?node) (object\_at\_node ?w ?node) (object\_at\_node ?c ?node))}
    \pddlline{\quad :effect (and (is\_open ?w) (increase (total-cost) 1))}
    \pddlline{)}
    \addvspace{8pt}

    \pddlline{(:action open\_curtain}
    \pddlline{\quad :parameters (?r ?hand ?c ?node)}
    \pddlline{\quad :precondition (and (robot\_has\_hand ?r ?hand) (hand\_free ?r ?hand) (curtain ?c) (not (is\_open ?c)) (robot\_at\_node ?r ?node) (object\_at\_node ?c ?node))}
    \pddlline{\quad :effect (and (is\_open ?c) (increase (total-cost) 1))}
    \pddlline{)}
    \addvspace{8pt}

    \pddlline{(:action close\_curtain}
    \pddlline{\quad :parameters (?r ?hand ?c ?node)}
    \pddlline{\quad :precondition (and (robot\_has\_hand ?r ?hand) (hand\_free ?r ?hand) (curtain ?c) (is\_open ?c) (robot\_at\_node ?r ?node) (object\_at\_node ?c ?node))}
    \pddlline{\quad :effect (and (not (is\_open ?c)) (increase (total-cost) 1))}
    \pddlline{)}
    \addvspace{8pt}

    \pddlline{(:action wipe\_blackboard}
    \pddlline{\quad :parameters (?r ?hand ?o ?b ?node)}
    \pddlline{\quad :precondition (and (robot\_has\_hand ?r ?hand) (holding ?r ?hand ?o) (can\_wipe\_blackboard ?o) (blackboard ?b) (robot\_at\_node ?r ?node) (object\_at\_node ?b ?node))}
    \pddlline{\quad :effect (and (wiped ?b) (increase (total-cost) 1))}
    \pddlline{)}
    \addvspace{8pt}

    \pddlline{(:action open\_remote}
    \pddlline{\quad :parameters (?r ?hand ?rm ?node)}
    \pddlline{\quad :precondition (and (robot\_has\_hand ?r ?hand) (hand\_free ?r ?hand) (remote ?rm) (not (is\_open ?rm)) (robot\_at\_node ?r ?node) (object\_at\_node ?rm ?node))}
    \pddlline{\quad :effect (and (is\_open ?rm) (increase (total-cost) 1))}
    \pddlline{)}
    \addvspace{8pt}

    \pddlline{(:action close\_remote}
    \pddlline{\quad :parameters (?r ?hand ?rm ?node)}
    \pddlline{\quad :precondition (and (robot\_has\_hand ?r ?hand) (hand\_free ?r ?hand) (remote ?rm) (is\_open ?rm) (robot\_at\_node ?r ?node) (object\_at\_node ?rm ?node))}
    \pddlline{\quad :effect (and (not (is\_open ?rm)) (increase (total-cost) 1))}
    \pddlline{)}
    \addvspace{8pt}

    \pddlline{(:action place\_in\_remote}
    \pddlline{\quad :parameters (?r ?hand ?b ?rm ?node)}
    \pddlline{\quad :precondition (and (robot\_has\_hand ?r ?hand) (holding ?r ?hand ?b) (remote ?rm) (battery ?b) (is\_open ?rm) (not (has\_battery ?rm)) (robot\_at\_node ?r ?node) (object\_at\_node ?rm ?node))}
    \pddlline{\quad :effect (and (in\_remote ?b ?rm) (hand\_free ?r ?hand) (not (holding ?r ?hand ?b)) (object\_at\_node ?b ?node) (has\_battery ?rm) (increase (total-cost) 1))}
    \pddlline{)}
    \addvspace{8pt}

    \pddlline{(:action pick\_from\_remote}
    \pddlline{\quad :parameters (?r ?hand ?b ?rm ?node)}
    \pddlline{\quad :precondition (and (robot\_has\_hand ?r ?hand) (hand\_free ?r ?hand) (remote ?rm) (battery ?b) (in\_remote ?b ?rm) (has\_battery ?rm) (is\_open ?rm) (robot\_at\_node ?r ?node) (object\_at\_node ?b ?node) (object\_at\_node ?rm ?node))}
    \pddlline{\quad :effect (and (not (in\_remote ?b ?rm)) (not (hand\_free ?r ?hand)) (holding ?r ?hand ?b) (not (object\_at\_node ?b ?node)) (not (has\_battery ?rm)) (increase (total-cost) 1))}
    \pddlline{)}
    \addvspace{8pt}

    \pddlline{(:action move\_robot}
    \pddlline{\quad :parameters (?r ?from ?to)}
    \pddlline{\quad :precondition (and (robot\_at\_node ?r ?from) (connected ?from ?to))}
    \pddlline{\quad :effect (and (robot\_at\_node ?r ?to) (not (robot\_at\_node ?r ?from)) (increase (total-cost) (travel\_cost ?from ?to)))}
    \pddlline{)}
    \addvspace{8pt}

    \pddlline{(:action open\_door}
    \pddlline{\quad :parameters (?r ?hand ?from ?to)}
    \pddlline{\quad :precondition (and (robot\_has\_hand ?r ?hand) (robot\_at\_node ?r ?from) (has\_door ?from ?to) (hand\_free ?r ?hand) (not (connected ?from ?to)))}
    \pddlline{\quad :effect (and (connected ?from ?to) (connected ?to ?from) (increase (total-cost) 1))}
    \pddlline{)}
\end{pddlbox}

\newpage
\subsection{Baselines}
As SayPlan and DELTA require a text-only scene graph as the environmental representation for task planning, we prompt a VLM to convert the images anchored at asset nodes into localized textual descriptions in JSON format. Specifically, we prompt the VLM to generate: (1) a unique object name, category, and affordances (e.g., turn on, turn off, support); (2) object attributes (e.g., color, shape, size); and (3) relations with other objects on the same asset node (e.g., on, left\_of).
An example is shown below:
\begin{jsonbox}
\jsonline{\{}
\jsonline{\quad "office\_612\_table": \{}
\jsonline{\quad\quad "desk\_1": \{"category": "furniture", "affordance": ["support\_objects", "place\_on"], "state": "static", "attributes": ["white", "flat\_surface"], "relation": \{"against": "wall\_1"\}\},}
\jsonline{\quad\quad "wall\_1": \{"category": "structure", "affordance": ["support\_against"], "state": "static", "attributes": ["textured", "vertical"]\},}
\jsonline{\quad\quad "pen\_cup\_1": \{}
\jsonline{\quad\quad\quad "category": "container", "affordance": ["contain", "hold", "organize"], "state": "upright",}
\jsonline{\quad\quad\quad "attributes": ["black", "mesh", "cylindrical"], "relation": \{"on": "desk\_1", "left\_of": "book\_holder\_1"\}}
\jsonline{\quad\quad \},}
\jsonline{\quad\quad "marker\_1": \{"category": "stationery", "affordance": ["write", "mark"], "state": "capped", "attributes": ["white\_body", "red\_cap", "blue\_markings"], "relation": \{"in": "pen\_cup\_1", "on": "desk\_1"\}\},}
\jsonline{\quad\quad "scissors\_1": \{"category": "tool", "affordance": ["cut"], "state": "closed", "attributes": ["teal\_handles"], "relation": \{"in": "pen\_cup\_1", "on": "desk\_1"\}\},}
\jsonline{\quad\quad "pen\_1": \{"category": "stationery", "affordance": ["write"], "state": "stored", "attributes": ["blue"], "relation": \{"in": "pen\_cup\_1", "on": "desk\_1"\}\},}
\jsonline{\quad\quad "green\_book\_or\_box\_1": \{}
\jsonline{\quad\quad\quad "category": "book", "affordance": ["read", "store"], "state": "closed", "attributes": ["dark\_green", "rectangular"],}
\jsonline{\quad\quad\quad "relation": \{"on": "desk\_1", "behind": "pen\_cup\_1", "left\_of": "book\_holder\_1"\}}
\jsonline{\quad\quad \},}
\jsonline{\quad\quad "book\_holder\_1": \{}
\jsonline{\quad\quad\quad "category": "organizer", "affordance": ["hold", "organize", "separate"], "state": "upright",}
\jsonline{\quad\quad\quad "attributes": ["mint\_green", "metal", "multi\_slot"], "relation": \{"on": "desk\_1", "right\_of": "pen\_cup\_1", "left\_of": "laptop\_1"\}}
\jsonline{\quad\quad \},}
\jsonline{\quad\quad "book\_1": \{"category": "book", "affordance": ["read", "open", "store"], "state": "upright\_closed", "attributes": ["white\_cover", "large\_print\_spine"], "relation": \{"in": "book\_holder\_1", "left\_of": "book\_2", "on": "desk\_1"\}\},}
\jsonline{\quad\quad "book\_2": \{"category": "book", "affordance": ["read", "open", "store"], "state": "upright\_closed", "attributes": ["yellow\_spine"], "relation": \{"in": "book\_holder\_1", "right\_of": "book\_1", "left\_of": "book\_3", "on": "desk\_1"\}\},}
\jsonline{\quad\quad "book\_3": \{"category": "book", "affordance": ["read", "open", "store"], "state": "upright\_closed", "attributes": ["light\_blue\_spine", "title\_visible"], "relation": \{"in": "book\_holder\_1", "right\_of": "book\_2", "left\_of": "book\_4", "on": "desk\_1"\}\},}
\jsonline{\quad\quad "book\_4": \{"category": "book", "affordance": ["read", "open", "store"], "state": "upright\_closed", "attributes": ["light\_pink\_or\_white\_cover"], "relation": \{"in": "book\_holder\_1", "right\_of": "book\_3", "on": "desk\_1"\}\},}
\jsonline{\quad\quad "laptop\_1": \{"category": "electronics", "affordance": ["open", "close", "compute", "charge"], "state": "closed", "attributes": ["silver", "gradient\_case\_blue\_pink"], "relation": \{"on": "desk\_1", "right\_of": "book\_holder\_1"\}\}}
\jsonline{\quad \},}
\jsonline{\quad "sink\_table": \{}
\jsonline{\quad\quad "countertop\_1": \{"category": "fixed\_asset", "affordance": ["support", "place\_on", "wipe"], "state": "dry", "attributes": ["blue\_green", "flat"], "relation": \{\}\},}
\jsonline{\quad\quad "sink\_basin\_1": \{"category": "fixed\_asset", "affordance": ["contain", "wash", "drain"], "state": "empty", "attributes": ["stainless\_steel", "rectangular"], "relation": \{"inset\_in": "countertop\_1", "below": "faucet\_1"\}\},}
\jsonline{\quad\quad "faucet\_1": \{"category": "fixture", "affordance": ["turn\_on", "turn\_off", "dispense\_water"], "state": "off", "attributes": ["metallic", "gooseneck"], "relation": \{"mounted\_on": "countertop\_1", "above": "sink\_basin\_1"\}\},}
\jsonline{\quad\quad "drain\_strainer\_1": \{"category": "plumbing\_component", "affordance": ["drain"], "state": "in\_place", "attributes": ["metal", "circular"], "relation": \{"in": "sink\_basin\_1", "at": "sink\_basin\_1"\}\},}
\jsonline{\quad\quad "soap\_dispenser\_1": \{"category": "container", "affordance": ["grasp", "pump", "dispense"], "state": "closed", "attributes": ["green", "plastic", "pump\_top"], "relation": \{"on": "countertop\_1", "behind": "sink\_basin\_1", "left\_of": "faucet\_1"\}\},}
\jsonline{\quad\quad "cleaner\_bottle\_1": \{"category": "container", "affordance": ["grasp", "pour", "store"], "state": "closed", "attributes": ["yellow", "tall", "orange\_cap", "labeled"], "relation": \{"on": "countertop\_1", "behind": "sink\_basin\_1", "right\_of": "faucet\_1"\}\},}
\jsonline{\quad\quad "cleaning\_cloths\_stack\_1": \{"category": "cleaning\_item", "affordance": ["grasp", "wipe", "fold"], "state": "folded", "attributes": ["green", "stacked", "cloth"], "relation": \{"on": "countertop\_1", "left\_of": "sink\_basin\_1", "against": "backsplash\_1"\}\},}
\jsonline{\quad\quad "rags\_pile\_1": \{"category": "cleaning\_item", "affordance": ["grasp", "wipe"], "state": "crumpled", "attributes": ["light\_colored", "cloth"], "relation": \{"on": "countertop\_1", "in\_front\_of": "cleaning\_cloths\_stack\_1", "left\_of": "sink\_basin\_1"\}\},}
\jsonline{\quad\quad "dish\_brush\_1": \{"category": "cleaning\_tool", "affordance": ["grasp", "scrub"], "state": "idle", "attributes": ["gray\_handle", "yellow\_head", "handheld"], "relation": \{"on": "countertop\_1", "left\_of": "sink\_basin\_1", "in\_front\_of": "rags\_pile\_1"\}\},}
\jsonline{\quad\quad "backsplash\_1": \{"category": "structure", "affordance": ["support\_mounting"], "state": "static", "attributes": ["blue\_green", "vertical\_surface"], "relation": \{"behind": "countertop\_1"\}\}}
\jsonline{\quad \}}
\jsonline{\}}
\end{jsonbox}

\subsubsection{\textbf{SayPlan Adaptation}}
After converting the multi-modal scene representation into a text-only scene graph (by replacing each image anchored at an asset node with a local textual description), we adapt SayPlan to our setting by introducing an contracted graph view and a customized expand/contract interface for efficient semantic search and planning.

\paragraph{\textbf{Contracted graph initialization}}
To obtain an initially contracted scene graph view, we apply the following simplifications:
(1) Pose nodes are omitted from the LLM-facing graph state to reduce graph size and improve semantic search efficiency; and
(2) all asset nodes are initialized as contracted (collapsed). In the initial view, assets are visible by name but do not expose their internal object descriptions. The environment state presented to the LLM consists of: (i) the \texttt{current\_location}, (ii) \texttt{visible\_assets} in the current area (collapsed by default), (iii) \texttt{connected\_rooms/areas} with door status, and (iv) a list of \texttt{expandable\_nodes}.
This yields a lightweight representation for Stage-1 semantic search while preserving the topology needed for navigation and task grounding. The initial view of the contracted scene graph is shown below:

\begin{jsonbox}
\jsonline{\{}
\jsonline{\quad "connected\_rooms": [}
\jsonline{\quad\quad \{"door": "closed", "name": "office\_604", "status": "collapsed", "type": "Room"\},}
\jsonline{\quad\quad \{"door": "closed", "name": "office\_605", "status": "collapsed", "type": "Room"\},}
\jsonline{\quad\quad \{"door": "closed", "name": "office\_606", "status": "collapsed", "type": "Room"\},}
\jsonline{\quad\quad \{"door": "closed", "name": "office\_607", "status": "collapsed", "type": "Room"\},}
\jsonline{\quad\quad \{"door": "closed", "name": "office\_612", "status": "collapsed", "type": "Room"\},}
\jsonline{\quad\quad \{"door": "closed", "name": "office\_609", "status": "collapsed", "type": "Room"\},}
\jsonline{\quad\quad \{"door": "closed", "name": "office\_610", "status": "collapsed", "type": "Room"\},}
\jsonline{\quad\quad \{"door": "closed", "name": "office\_611", "status": "collapsed", "type": "Room"\},}
\jsonline{\quad\quad \{"door": "closed", "name": "office\_608\_left", "status": "collapsed", "type": "Room"\},}
\jsonline{\quad\quad \{"door": "closed", "name": "room\_608\_right", "status": "collapsed", "type": "Room"\},}
\jsonline{\quad\quad \{"door": "closed", "name": "storage\_room\_616", "status": "collapsed", "type": "Room"\},}
\jsonline{\quad\quad \{"door": "closed", "name": "office\_614", "status": "collapsed", "type": "Room"\},}
\jsonline{\quad\quad \{"door": "closed", "name": "kitchen", "status": "collapsed", "type": "Room"\},}
\jsonline{\quad\quad \{"door": "closed", "name": "office\_601", "status": "collapsed", "type": "Room"\},}
\jsonline{\quad\quad \{"door": "closed", "name": "office\_602", "status": "collapsed", "type": "Room"\},}
\jsonline{\quad\quad \{"door": "closed", "name": "classroom\_601\_left", "status": "collapsed", "type": "Room"\},}
\jsonline{\quad\quad \{"door": "closed", "name": "classroom\_601\_right", "status": "collapsed", "type": "Room"\},}
\jsonline{\quad\quad \{"door": "closed", "name": "office\_613", "status": "collapsed", "type": "Room"\}}
\jsonline{\quad ],}
\jsonline{\quad "expandable\_nodes": ["office\_604", "office\_605", "office\_606", "office\_607", "office\_612", "office\_609", "office\_610", "office\_611", "office\_608\_left", "room\_608\_right", "storage\_room\_616", "office\_614", "kitchen", "office\_601", "office\_602", "classroom\_601\_left", "classroom\_601\_right", "office\_613"],}
\jsonline{\quad "visible\_assets": [}
\jsonline{\quad\quad \{"name": "coffee\_maker", "status": "collapsed", "type": "Asset"\}, \{"name": "fridge", "status": "collapsed", "type": "Asset"\},}
\jsonline{\quad\quad \{"name": "microwave\_table", "status": "collapsed", "type": "Asset"\}, \{"name": "drying\_rack", "status": "collapsed", "type": "Asset"\},}
\jsonline{\quad\quad \{"name": "kettle\_table", "status": "collapsed", "type": "Asset"\}, \{"name": "sink\_table", "status": "collapsed", "type": "Asset"\},}
\jsonline{\quad\quad \{"name": "trash\_bin", "status": "collapsed", "type": "Asset"\}, \{"name": "washing\_machine", "status": "collapsed", "type": "Asset"\},}
\jsonline{\quad\quad \{"name": "flower", "status": "collapsed", "type": "Asset"\}, \{"name": "meeting\_table", "status": "collapsed", "type": "Asset"\}}
\jsonline{\quad ]}
\jsonline{\}}
\end{jsonbox}

\paragraph{\textbf{Expand/contract behavior}}
We follow the same high-level interaction pattern as SayPlan (i.e., \texttt{expand}, \texttt{contract}, \texttt{terminate}), but modify the expansion semantics to match our multi-modal-to-text conversion and improve search efficiency:
\begin{itemize}[leftmargin=1.2em]
    \item Expanding an asset node reveals the localized \text scene graph anchored to that asset (from \texttt{text\_scene\_graph.json}), including object names/categories/affordances/attributes and intra-asset relations.
    \item Expanding a room/area node expands all asset nodes inside that room/area at once. This reduces the number of semantic-search steps, since a single room expansion exposes all asset-level object descriptions within it.
    \item Contracting a node hides its expanded details and restores the compact view (assets/rooms revert to \texttt{collapsed}).
\end{itemize}
In our JSON state, each expandable entity is explicitly marked with a \texttt{status} field in \{\texttt{collapsed}, \texttt{expanded}\}, enabling the LLM to reason over visibility and decide the next search action.

\paragraph{Path planner and move post-processing.}
For motion planning, the LLM is allowed to output high-level \texttt{move(...)} actions between graph nodes, subject to door constraints (i.e., traversal is invalid if a closed door lies on the route). We then apply a deterministic post-processing step:
\begin{itemize}[leftmargin=1.2em]
    \item A Dijkstra-based path planner expands each high-level \texttt{move} into a sequence of intermediate \texttt{move} actions along the shortest path.
    \item To keep door interactions well-formed, when the target is a room node the expanded path stops at the pre-door node, so that a subsequent \texttt{open\_door(...)} can be executed before entering the room; after \texttt{open\_door(...)}, the post-processor inserts the minimal moves needed to cross the doorway and proceed inside.
\end{itemize}
This separation keeps the LLM plan concise while guaranteeing executable navigation sequences under the environment's door dynamics.

\subsubsection{\textbf{DELTA Adaptation}}
We adapt DELTA to our multi-modal setting by modifying three components to better align with our pipeline implementation: \emph{domain construction}, \emph{scene graph pruning}, and \emph{autoregressive sub-task planning}.

\paragraph{\textbf{Domain selection from an expanded reference domain}}
Directly generating a household PDDL domain from natural-language descriptions is difficult and often unstable. Instead, we first construct an expanded, feature-complete reference domain by enabling the required capabilities (e.g., mobility, multi-arm manipulation, and door interactions). We then prompt the LLM to perform domain selection conditioned on the task instruction, i.e., to select a relevant subset of actions and predicates from the reference domain. This design reduces generation complexity while keeping the resulting domain compatible with downstream PDDL planning.

\paragraph{\textbf{Asset-aware pruning on the text-only scene graph}}
For pruning, we leverage the text-only scene graph derived from images. We first extract an asset-indexed inventory of object names, where each asset node (e.g., a table or cabinet) is mapped to the list of object instances anchored on it. The LLM is then prompted with this inventory and asked to retain task-relevant object nodes based on the instruction.
Given the selected objects, we prune the scene graph by (i) filtering each asset's object list to the retained subset, and (ii) pruning the entire asset node if it contains no retained objects. To preserve global navigability, we keep all structural nodes (e.g., rooms and poses) and only remove irrelevant asset nodes.
In the final pruned representation, the retained object descriptions are attached directly to their corresponding asset nodes via an \texttt{objects} field, rather than being stored as a separate top-level text graph.

\paragraph{\textbf{Autoregressive sub-task planning with \textsc{ValStep} state propagation}}
We implement autoregressive sub-task planning by decomposing the overall goal into a sequence of PDDL sub-goals and solving them step-by-step. After obtaining a plan for each sub-goal, we use \textsc{ValStep} (from the VAL tool) to execute the plan and extract the resulting state. Concretely, we parse the next problem file produced by \textsc{ValStep} and reuse its \texttt{:init} block as the initial state for the subsequent sub-goal. This yields consistent state transitions across sub-goals and produces a final plan by concatenating all per-subgoal plans.

\newpage
\subsection{Simulation Environment Details}
The environment uses the same topological map and experimental settings (single-/dual-arm, with/without doors) as those in the evaluation tasks. For all asset nodes, human experts implemented the object types, relationships, and object names according to the task specifications.

\paragraph{Unified Action Interface.}
For all methods, the environment accepts a unified action-sequence format with 17 action types:
\texttt{pick}, \texttt{place\_in}, \texttt{place\_on}, \texttt{place\_under}, \texttt{open}, \texttt{close}, \texttt{pour}, \texttt{cut}, \texttt{stir},
\texttt{scoop}, \texttt{fold}, \texttt{wipe}, \texttt{turn\_on}, \texttt{turn\_off}, \texttt{hang\_on}, \texttt{open\_door}, and \texttt{move}.
The interfaces are as follows:
\begin{itemize}
    \item \texttt{pick(robot, hand, obj)}: The robot \texttt{robot} picks up object \texttt{obj} using \texttt{hand}.
    \item \texttt{place\_in(robot, hand, obj)}: The robot \texttt{robot} places the object held in \texttt{hand} into container \texttt{obj} (e.g., a drawer).
    \item \texttt{place\_on(robot, hand, obj)}: The robot \texttt{robot} places the object held in \texttt{hand} onto surface \texttt{obj} (e.g., a table).
    \item \texttt{open(robot, hand, obj)}: The robot \texttt{robot} opens container \texttt{obj} using \texttt{hand}.
    \item \texttt{close(robot, hand, obj)}: The robot \texttt{robot} closes container \texttt{obj} using \texttt{hand}.
    \item \texttt{place\_under(robot, hand, obj)}: The robot \texttt{robot} places the object held in \texttt{hand} under \texttt{obj} (e.g., a faucet) while still holding it.
    \item \texttt{pour(robot, hand, obj)}: The robot \texttt{robot} pours from the object held in \texttt{hand} into object \texttt{obj}.
    \item \texttt{cut(robot, hand, obj)}: The robot \texttt{robot} cuts object \texttt{obj} using the object held in \texttt{hand} (e.g., a knife).
    \item \texttt{stir(robot, hand, obj)}: The robot \texttt{robot} stirs object \texttt{obj} using the object held in \texttt{hand} (e.g., a spoon).
    \item \texttt{scoop(robot, hand, obj)}: The robot \texttt{robot} scoops from object \texttt{obj} using the object held in \texttt{hand} (e.g., a rice paddle).
    \item \texttt{fold(robot, hand, obj)}: The robot \texttt{robot} folds object \texttt{obj} using \texttt{hand}.
    \item \texttt{wipe(robot, hand, obj)}: The robot \texttt{robot} wipes object \texttt{obj} using the object held in \texttt{hand} (e.g., a cloth).
    \item \texttt{turn\_on(robot, hand, obj)}: The robot \texttt{robot} turns on object \texttt{obj} (e.g., a faucet or microwave) using \texttt{hand}.
    \item \texttt{turn\_off(robot, hand, obj)}: The robot \texttt{robot} turns off object \texttt{obj} (e.g., a faucet or lamp) using \texttt{hand}.
    \item \texttt{hang\_on(robot, hand, obj)}: The robot \texttt{robot} hangs object \texttt{obj} (e.g., a towel) on the object held in \texttt{hand} (e.g., a drying rack).
    \item \texttt{open\_door(robot, hand, door\_name)}: The robot \texttt{robot} opens a closed door \texttt{door\_name} using \texttt{hand}. The \texttt{door\_name} follows the format \texttt{"door\_\{node\_name\_1\}\_\{node\_name\_2\}"}.
    \item \texttt{move(robot, to\_node)}: The robot \texttt{robot} moves to location \texttt{to\_node}.
\end{itemize}

\paragraph{Action Parsing.}
For \textit{LLM as Planner} and \textit{SayPlan}, we prompt the LLM to output the specified action format directly (e.g., \texttt{pick(robot, left\_hand, apple)}).
For \textit{DELTA} and \textit{UniPlan}, we implement a parameterized mapping table based on the domain operators. For example, a PDDL solver may output an operator
\texttt{(put\_in\_bin robot left\_hand black\_cap\_bottle\_1 black\_trashbin\_1 trash\_bin)},
where \texttt{robot}, \texttt{left\_hand}, \texttt{black\_cap\_empty\_bottle\_1}, \texttt{black\_trash\_bin\_1}, and \texttt{trash\_bin} denote the robot, the hand, the object held by the specified hand, the target container, and the asset-node name, respectively. We extract the required fields and parse the operator into the unified action format, e.g.,
\texttt{place\_in(robot, left\_hand, black\_trash\_bin\_1)}.
In practice, we only need to recognize the parameter indices required by each action type. The mapping table consists of \texttt{Operator name $\rightarrow$ Action type} and the corresponding parameter indices.

\paragraph{Object Grounding.}
After aligning action formats, we ground object references into the environment, because different methods may use different names for the same object (e.g., \texttt{apple\_1} vs.\ \texttt{red\_apple\_1}).
For each action sequence, we first track locations to determine where each object appears for the first time; we treat that node as the object's initial node. We only map objects within the same asset node, because an object in asset node A generated by a method cannot correspond to an object in asset node B in the environment.

Within each asset node, for each method-specific object name, we evaluate two grounding approaches:
\begin{enumerate}
    \item \textbf{Bipartite graph matching} using pairwise similarity:
    \begin{itemize}
        \item \textbf{Fuzzy matching}: based on string distance. This is the fastest but cannot distinguish semantically similar names.
        \item \textbf{Embedding-based matching}: using embedding models (e.g., BGE-M3, Qwen-3 Embedding) and reranking models (e.g., BGE-M3 Reranking, Qwen-3 Reranking). This is generally more accurate, but can still miss fine-grained semantics; for example, \texttt{white\_table} may be incorrectly matched to \texttt{white\_plate} instead of \texttt{table}.
    \end{itemize}
    \item \textbf{LLM-based matching}: the slowest but most accurate. We prompt the LLM with all objects within the same asset node and ask it to align the names. For accurate evaluation, we ultimately use this method.
\end{enumerate}

\newpage
\subsection{UniPlan Planning Process}

With a given expanded domain, the full map, and the task instruction:
``Prepare two cups of coffee and place them on the meeting table.'' (Single-Arm with Door)

\paragraph{\textbf{Task-Oriented Retrieval}}
In this stage, the LLM reasons over the task instruction and the textual index to retrieve the most relevant key nodes, thereby filtering out redundant environmental information.

\begin{sayplanbox}
"reasoning": "Task: make two cups of coffee and deliver them to the meeting table. The coffee must be brewed at the coffee\_maker (source of coffee). Two cups are required; candidate nodes that visibly contain cups are relevant as potential containers (office\_602\_table has two cups/containers). The final destination is explicitly the meeting\_table, so it must be included as the target. Other nodes (fridge, kettle\_table, sink\_table, etc.) are not strictly required from the given candidates to execute the minimal plan (brew coffee + pour into two cups + place on meeting table)." \\
\addvspace{5pt}
"selected\_nodes": ["coffee\_maker", "office\_602\_table", "meeting\_table"]
\end{sayplanbox}

\begin{figure}[ht]
    \centering
    \includegraphics[width=0.9\linewidth]{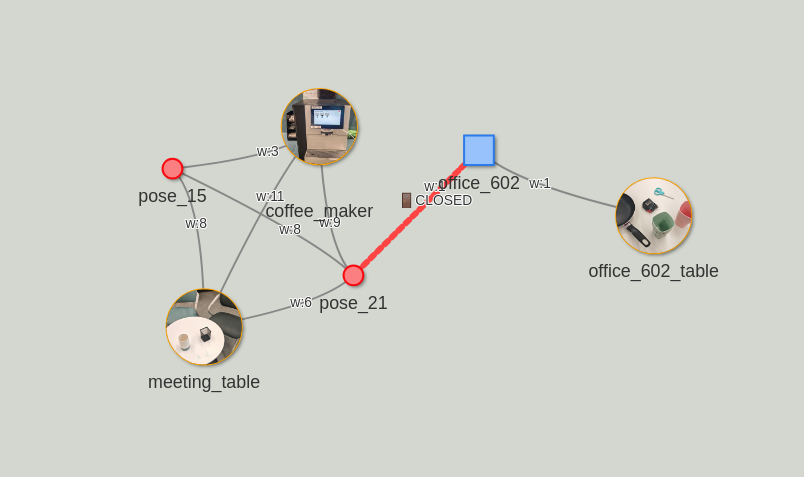}
    \caption{The compressed map generated through task-oriented retrieval. It focuses on the coffee maker, office 602 table (for cups), and the meeting table, while maintaining the minimal topological connectivity required for navigation.}
    \label{fig:compressed_map}
\end{figure}

By selecting these key nodes, we significantly reduce the complexity of the scene graph. The resulting compressed map, which retains only the task-relevant topology and necessary objects, is illustrated in Figure~\ref{fig:compressed_map}.

\paragraph{\textbf{VLM Grounding and State Extraction}}
After compressing the map, the VLM performs visual grounding on the images associated with the key nodes. It extracts object instances, their current states, and initial logical predicates required for the PDDL problem.

\begin{jsonbox}
\jsonline{\{}
\jsonline{\quad "reasoning": "Task: prepare two cups of coffee and place them on the meeting table. The coffee maker is available (image coffee\_maker). Two cups are available on the office table (image office\_602\_table): a green cup and a pink cup. The meeting table (image meeting\_table) already has one white cup and a black mesh holder on it; these are not required for the goal but indicate the destination surface...",}
\jsonline{\quad "objects": \{}
\jsonline{\quad\quad "coffee\_maker": ["coffee\_maker\_1"],}
\jsonline{\quad\quad "office\_602\_table": ["office\_table\_1", "green\_cup\_1", "pink\_cup\_1"],}
\jsonline{\quad\quad "meeting\_table": ["meeting\_table\_1", "white\_cup\_1", "black\_holder\_1"]}
\jsonline{\quad \},}
\jsonline{\quad "init": ["(coffee\_maker coffee\_maker\_1)", "(table office\_table\_1)", "(cup green\_cup\_1)", "(cup pink\_cup\_1)", "(on\_table green\_cup\_1 office\_table\_1)", "(on\_table pink\_cup\_1 office\_table\_1)", "(table meeting\_table\_1)", "(cup white\_cup\_1)", "(holder black\_holder\_1)", "(on\_table white\_cup\_1 meeting\_table\_1)", "(on\_table black\_holder\_1 meeting\_table\_1)"],}
\jsonline{\quad "goal": "(and (filled\_coffee green\_cup\_1) (filled\_coffee pink\_cup\_1) (on\_table green\_cup\_1 meeting\_table\_1) (on\_table pink\_cup\_1 meeting\_table\_1))"}
\jsonline{\}}
\end{jsonbox}

\paragraph{\textbf{Problem Construction}}
The system automatically populates the PDDL problem file by injecting the VLM-extracted states and the topological constraints (connections and costs) from the compressed map.

\begin{pddlbox}
\pddlline{(define (problem uniplan-problem)}
\pddlline{\quad (:domain uniplan-domain)}
\pddlline{\quad (:objects office\_602 office\_602\_table pose\_21 coffee\_maker pose\_15 meeting\_table robot coffee\_maker\_1 office\_table\_1 green\_cup\_1 pink\_cup\_1 meeting\_table\_1 white\_cup\_1 black\_holder\_1)}
\pddlline{\quad (:init}
\pddlline{\quad\quad (coffee\_maker coffee\_maker\_1) (table office\_table\_1) (cup green\_cup\_1) (cup pink\_cup\_1)}
\pddlline{\quad\quad (on\_table green\_cup\_1 office\_table\_1) (on\_table pink\_cup\_1 office\_table\_1)}
\pddlline{\quad\quad (table meeting\_table\_1) (cup white\_cup\_1) (holder black\_holder\_1)}
\pddlline{\quad\quad (on\_table white\_cup\_1 meeting\_table\_1) (on\_table black\_holder\_1 meeting\_table\_1)}
\pddlline{\quad\quad (connected office\_602 office\_602\_table) (= (travel\_cost office\_602 office\_602\_table) 1)}
\pddlline{\quad\quad (has\_door office\_602 pose\_21) (= (travel\_cost office\_602 pose\_21) 1)}
\pddlline{\quad\quad (connected pose\_21 coffee\_maker) (= (travel\_cost pose\_21 coffee\_maker) 9)}
\pddlline{\quad\quad (robot\_at\_node robot pose\_15) (hand\_free robot)}
\pddlline{\quad\quad (object\_at\_node green\_cup\_1 office\_602\_table) (object\_at\_node coffee\_maker\_1 coffee\_maker)}
\pddlline{\quad\quad (= (total-cost) 0)}
\pddlline{\quad )}
\pddlline{\quad (:goal (and (filled\_coffee green\_cup\_1) (filled\_coffee pink\_cup\_1) (on\_table green\_cup\_1 meeting\_table\_1) (on\_table pink\_cup\_1 meeting\_table\_1)))}
\pddlline{\quad (:metric minimize (total-cost))}
\pddlline{)}
\end{pddlbox}

\paragraph{\textbf{Plan Generation}}
The PDDL planner finds the optimal sequence of actions based on the travel costs and task constraints. The generated plan is shown below:

\begin{pddlbox}
\pddlline{(move\_robot robot pose\_15 pose\_21)}
\pddlline{(open\_door robot pose\_21 office\_602)}
\pddlline{(move\_robot robot pose\_21 office\_602)}
\pddlline{(move\_robot robot office\_602 office\_602\_table)}
\pddlline{(pick\_from\_table robot green\_cup\_1 office\_table\_1 office\_602\_table)}
\pddlline{(move\_robot robot office\_602\_table office\_602)}
\pddlline{(move\_robot robot office\_602 pose\_21)}
\pddlline{(move\_robot robot pose\_21 coffee\_maker)}
\pddlline{(place\_on\_coffee\_maker robot green\_cup\_1 coffee\_maker\_1 coffee\_maker)}
\pddlline{(fill\_coffee\_into\_cup robot green\_cup\_1 coffee\_maker\_1 coffee\_maker)}
\pddlline{(pick\_from\_coffee\_maker robot green\_cup\_1 coffee\_maker\_1 coffee\_maker)}
\pddlline{(move\_robot robot coffee\_maker meeting\_table)}
\pddlline{(place\_on\_table robot green\_cup\_1 meeting\_table\_1 meeting\_table)}
\pddlline{(move\_robot robot meeting\_table pose\_21)}
\pddlline{(move\_robot robot pose\_21 office\_602)}
\pddlline{(move\_robot robot office\_602 office\_602\_table)}
\pddlline{(pick\_from\_table robot pink\_cup\_1 office\_table\_1 office\_602\_table)}
\pddlline{(move\_robot robot office\_602\_table office\_602)}
\pddlline{(move\_robot robot office\_602 pose\_21)}
\pddlline{(move\_robot robot pose\_21 coffee\_maker)}
\pddlline{(place\_on\_coffee\_maker robot pink\_cup\_1 coffee\_maker\_1 coffee\_maker)}
\pddlline{(fill\_coffee\_into\_cup robot pink\_cup\_1 coffee\_maker\_1 coffee\_maker)}
\pddlline{(pick\_from\_coffee\_maker robot pink\_cup\_1 coffee\_maker\_1 coffee\_maker)}
\pddlline{(move\_robot robot coffee\_maker meeting\_table)}
\pddlline{(place\_on\_table robot pink\_cup\_1 meeting\_table\_1 meeting\_table)}
\addvspace{5pt}
\pddlline{; cost = 73 (general cost)}
\end{pddlbox}

\paragraph{\textbf{Final Plan Expansion}}
To execute the task in the actual environment, the compressed movement actions are expanded into fine-grained, point-to-point navigation sequences. The complete and executable final plan is detailed below:
\begin{pddlbox}
\pddlline{(move\_robot robot pose\_15 pose\_16)}
\pddlline{(move\_robot robot pose\_16 pose\_17)}
\pddlline{(move\_robot robot pose\_17 pose\_27)}
\pddlline{(move\_robot robot pose\_27 pose\_28)}
\pddlline{(move\_robot robot pose\_28 pose\_24)}
\pddlline{(move\_robot robot pose\_24 pose\_23)}
\pddlline{(move\_robot robot pose\_23 pose\_20)}
\pddlline{(move\_robot robot pose\_20 pose\_21)}
\pddlline{(open\_door robot pose\_21 office\_602)}
\pddlline{(move\_robot robot pose\_21 office\_602)}
\pddlline{(move\_robot robot office\_602 office\_602\_table)}
\pddlline{(pick\_from\_table robot green\_cup\_1 office\_table\_1 office\_602\_table)}
\pddlline{(move\_robot robot office\_602\_table office\_602)}
\pddlline{(move\_robot robot office\_602 pose\_21)}
\pddlline{(move\_robot robot pose\_21 pose\_20)}
\pddlline{(move\_robot robot pose\_20 pose\_19)}
\pddlline{(move\_robot robot pose\_19 pose\_18)}
\pddlline{(move\_robot robot pose\_18 pose\_13)}
\pddlline{(move\_robot robot pose\_13 pose\_6)}
\pddlline{(move\_robot robot pose\_6 pose\_7)}
\pddlline{(move\_robot robot pose\_7 pose\_1)}
\pddlline{(move\_robot robot pose\_1 pose\_3)}
\pddlline{(move\_robot robot pose\_3 coffee\_maker)}
\pddlline{(place\_on\_coffee\_maker robot green\_cup\_1 coffee\_maker\_1 coffee\_maker)}
\pddlline{(fill\_coffee\_into\_cup robot green\_cup\_1 coffee\_maker\_1 coffee\_maker)}
\pddlline{(pick\_from\_coffee\_maker robot green\_cup\_1 coffee\_maker\_1 coffee\_maker)}
\pddlline{(move\_robot robot coffee\_maker pose\_3)}
\pddlline{(move\_robot robot pose\_3 pose\_14)}
\pddlline{(move\_robot robot pose\_14 pose\_15)}
\pddlline{(move\_robot robot pose\_15 pose\_16)}
\pddlline{(move\_robot robot pose\_16 pose\_17)}
\pddlline{(move\_robot robot pose\_17 pose\_27)}
\pddlline{(move\_robot robot pose\_27 pose\_28)}
\pddlline{(move\_robot robot pose\_28 pose\_24)}
\pddlline{(move\_robot robot pose\_24 pose\_25)}
\pddlline{(move\_robot robot pose\_25 pose\_26)}
\pddlline{(move\_robot robot pose\_26 meeting\_table)}
\pddlline{(place\_on\_table robot green\_cup\_1 meeting\_table\_1 meeting\_table)}
\pddlline{(move\_robot robot meeting\_table pose\_26)}
\pddlline{(move\_robot robot pose\_26 pose\_25)}
\pddlline{(move\_robot robot pose\_25 pose\_24)}
\pddlline{(move\_robot robot pose\_24 pose\_23)}
\pddlline{(move\_robot robot pose\_23 pose\_20)}
\pddlline{(move\_robot robot pose\_20 pose\_21)}
\pddlline{(move\_robot robot pose\_21 office\_602)}
\pddlline{(move\_robot robot office\_602 office\_602\_table)}
\pddlline{(pick\_from\_table robot pink\_cup\_1 office\_table\_1 office\_602\_table)}
\pddlline{(move\_robot robot office\_602\_table office\_602)}
\pddlline{(move\_robot robot office\_602 pose\_21)}
\pddlline{(move\_robot robot pose\_21 pose\_20)}
\pddlline{(move\_robot robot pose\_20 pose\_19)}
\pddlline{(move\_robot robot pose\_19 pose\_18)}
\pddlline{(move\_robot robot pose\_18 pose\_13)}
\pddlline{(move\_robot robot pose\_13 pose\_6)}
\pddlline{(move\_robot robot pose\_6 pose\_7)}
\pddlline{(move\_robot robot pose\_7 pose\_1)}
\pddlline{(move\_robot robot pose\_1 pose\_3)}
\pddlline{(move\_robot robot pose\_3 coffee\_maker)}
\pddlline{(place\_on\_coffee\_maker robot pink\_cup\_1 coffee\_maker\_1 coffee\_maker)}
\pddlline{(fill\_coffee\_into\_cup robot pink\_cup\_1 coffee\_maker\_1 coffee\_maker)}
\pddlline{(pick\_from\_coffee\_maker robot pink\_cup\_1 coffee\_maker\_1 coffee\_maker)}
\pddlline{(move\_robot robot coffee\_maker pose\_3)}
\pddlline{(move\_robot robot pose\_3 pose\_14)}
\pddlline{(move\_robot robot pose\_14 pose\_15)}
\pddlline{(move\_robot robot pose\_15 pose\_16)}
\pddlline{(move\_robot robot pose\_16 pose\_17)}
\pddlline{(move\_robot robot pose\_17 pose\_27)}
\pddlline{(move\_robot robot pose\_27 pose\_28)}
\pddlline{(move\_robot robot pose\_28 pose\_24)}
\pddlline{(move\_robot robot pose\_24 pose\_25)}
\pddlline{(move\_robot robot pose\_25 pose\_26)}
\pddlline{(move\_robot robot pose\_26 meeting\_table)}
\pddlline{(place\_on\_table robot pink\_cup\_1 meeting\_table\_1 meeting\_table)}
\pddlline{; cost = 73 (general cost)}
\end{pddlbox}

\newpage
\subsection{Planning Examples}
To improve readability and conciseness, the plan sequences shown below are presented in a compressed format. Consecutive navigation actions (e.g., moving between adjacent waypoints) are aggregated into a single action, denoted as $\texttt{Move}(\textit{destination})$. However, the reported Cost reflects the actual total distance traversed in the complete, uncompressed plan.
\begin{examplebox}{4: Move apple from fridge to office 604 table}
\textit{Instruction: There is an apple in the fridge. Place it on the table in office 604.}

\begin{multicols}{2}
    \centering \textbf{Single-Arm with Door}
    \vfill\null \columnbreak
    \centering \textbf{Dual-Arm with Door}
\end{multicols}
\hrule

\begin{multicols}{2}
    \methodlabel{UniPlan (Cost: 27)}
    \begin{itemize}[leftmargin=10pt, noitemsep, topsep=0pt]
        \item \scriptsize Move(pose\_4)
        \item \scriptsize OpenDoor(hand, door\_604)
        \item \scriptsize Move(fridge)
        \item \scriptsize Open(hand, fridge)
        \item \scriptsize Pick(hand, apple)
        \item \scriptsize Move(office\_604\_table)
        \item \scriptsize PlaceOn(hand, table)
    \end{itemize}

    \columnbreak

    \methodlabel{UniPlan (Cost: 17)}
    \begin{itemize}[leftmargin=10pt, noitemsep, topsep=0pt]
        \item \scriptsize Move(fridge)
        \item \scriptsize Open(left\_hand, fridge)
        \item \scriptsize Pick(left\_hand, apple)
        \item \scriptsize Move(pose\_4)
        \item \scriptsize OpenDoor(right\_hand, door\_604)
        \item \scriptsize Move(office\_604\_table)
        \item \scriptsize PlaceOn(left\_hand, table)
    \end{itemize}
\end{multicols}
\hrule

\begin{multicols}{2}
    \methodlabel{llm-as-planner (Cost: 17)}
    \begin{itemize}[leftmargin=10pt, noitemsep, topsep=0pt]
        \item \scriptsize Move(fridge)
        \item \scriptsize Open(hand, fridge)
        \item \scriptsize Pick(hand, apple)
        \item \scriptsize Move(pose\_4)
        \item \scriptsize OpenDoor(hand, door\_604) \redcross \quad \textbf{\textcolor{red}{(Hand occupied)}}
        \item \textcolor{gray}{Move(office\_604\_table)}
        \item \textcolor{gray}{PlaceOn(hand, table)}
    \end{itemize}

    \columnbreak

    \methodlabel{llm-as-planner (Cost: 17)}
    \begin{itemize}[leftmargin=10pt, noitemsep, topsep=0pt]
        \item \scriptsize Move(fridge)
        \item \scriptsize Open(right\_hand, fridge)
        \item \scriptsize Pick(left\_hand, apple)
        \item \scriptsize Move(pose\_4)
        \item \scriptsize OpenDoor(right\_hand, door\_604)
        \item \scriptsize Move(office\_604\_table)
        \item \scriptsize PlaceOn(left\_hand, table)
    \end{itemize}
\end{multicols}
\hrule

\begin{multicols}{2}
    \methodlabel{SayPlan (Cost: 31)}
    \begin{itemize}[leftmargin=10pt, noitemsep, topsep=0pt]
        \item \scriptsize Move(fridge)
        \item \scriptsize Open(hand, fridge)
        \item \scriptsize Pick(hand, apple)
        \item \scriptsize Move(microwave\_table)
        \item \scriptsize PlaceOn(hand, table)
        \item \scriptsize Move(pose\_4)
        \item \scriptsize OpenDoor(hand, door\_604)
        \item \scriptsize Move(microwave\_table)
        \item \scriptsize Pick(hand, apple)
        \item \scriptsize Move(office\_604\_table)
        \item \scriptsize PlaceOn(hand, table)
    \end{itemize}

    \columnbreak

    \methodlabel{SayPlan (Cost: 18)}
    \begin{itemize}[leftmargin=10pt, noitemsep, topsep=0pt]
        \item \scriptsize Move(fridge)
        \item \scriptsize Open(right\_hand, fridge)
        \item \scriptsize Pick(left\_hand, apple)
        \item \scriptsize Close(right\_hand, fridge)
        \item \scriptsize Move(pose\_4)
        \item \scriptsize OpenDoor(right\_hand, door\_604)
        \item \scriptsize Move(office\_604\_table)
        \item \scriptsize PlaceOn(left\_hand, table)
    \end{itemize}
\end{multicols}
\hrule

\begin{multicols}{2}
    \methodlabel{DELTA (Cost: 0)}
    \begin{itemize}[leftmargin=10pt, noitemsep, topsep=0pt]
        \item \scriptsize \textit{No solution found.}
    \end{itemize}

    \columnbreak

    \methodlabel{DELTA (Cost: 27)}
    \begin{itemize}[leftmargin=10pt, noitemsep, topsep=0pt]
        \item \scriptsize Move(fridge)
        \item \scriptsize Open(left\_hand, fridge)
        \item \scriptsize Pick(left\_hand, apple)
        \item \scriptsize Move(pose\_4)
        \item \scriptsize OpenDoor(right\_hand, door\_604)
        \item \scriptsize Move(office\_604\_table)
        \item \scriptsize PlaceOn(left\_hand, table)
    \end{itemize}
\end{multicols}
\end{examplebox}

\newpage
\begin{examplebox}{22: Wash cloth and hang on drying rack}
\textit{Instruction: Wash the cloth on the couch and hang it on the drying rack.}

\begin{multicols}{2}
    \centering \textbf{Single-Arm with Door}
    \vfill\null \columnbreak
    \centering \textbf{Dual-Arm with Door}
\end{multicols}
\hrule

\begin{multicols}{2}
    \methodlabel{UniPlan (Cost: 32)}
    \begin{itemize}[leftmargin=8pt, noitemsep, topsep=0pt]
        \item \scriptsize Move(washing\_machine)
        \item \scriptsize Open(hand, washing\_machine)
        \item \scriptsize Move(pose\_26)
        \item \scriptsize OpenDoor(hand, door\_classroom\_601\_right\_pose\_26)
        \item \scriptsize Move(room\_601\_couch)
        \item \scriptsize Pick(hand, cloth)
        \item \scriptsize Move(washing\_machine)
        \item \scriptsize PlaceIn(hand, washing\_machine)
        \item \scriptsize Close(hand, washing\_machine)
        \item \scriptsize TurnOn(hand, washing\_machine)
        \item \scriptsize Open(hand, washing\_machine)
        \item \scriptsize Pick(hand, cloth)
        \item \scriptsize Move(drying\_rack)
        \item \scriptsize HangOn(hand, drying\_rack)
    \end{itemize}

    \columnbreak

    \methodlabel{UniPlan (Cost: 30)}
    \begin{itemize}[leftmargin=8pt, noitemsep, topsep=0pt]
        \item \scriptsize Move(pose\_26)
        \item \scriptsize OpenDoor(left\_hand, door\_classroom\_601\_right\_pose\_26)
        \item \scriptsize Move(room\_601\_couch)
        \item \scriptsize Pick(left\_hand, cloth)
        \item \scriptsize Move(washing\_machine)
        \item \scriptsize Open(right\_hand, washing\_machine)
        \item \scriptsize PlaceIn(left\_hand, washing\_machine)
        \item \scriptsize Close(left\_hand, washing\_machine)
        \item \scriptsize TurnOn(left\_hand, washing\_machine)
        \item \scriptsize Open(left\_hand, washing\_machine)
        \item \scriptsize Pick(left\_hand, cloth)
        \item \scriptsize Move(drying\_rack)
        \item \scriptsize HangOn(left\_hand, drying\_rack)
    \end{itemize}
\end{multicols}
\hrule

\begin{multicols}{2}
    \methodlabel{llm-as-planner (Cost: 36)}
    \begin{itemize}[leftmargin=8pt, noitemsep, topsep=0pt]
        \item \scriptsize Move(pose\_24)
        \item \scriptsize OpenDoor(hand, door\_classroom\_601\_left\_pose\_24)
        \item \scriptsize Move(room\_601\_couch)
        \item \scriptsize Pick(hand, cloth)
        \item \scriptsize Move(classroom\_601\_right)
        \item \scriptsize OpenDoor(hand, door\_classroom\_601\_right\_pose\_26) \redcross \quad \textbf{\textcolor{red}{(Hand occupied)}}
        \item \scriptsize \textcolor{gray}{Move(washing\_machine)}
        \item \scriptsize \textcolor{gray}{Open(hand, washing\_machine)}
        \item \scriptsize \textcolor{gray}{PlaceIn(hand, washing\_machine)}
        \item \scriptsize \textcolor{gray}{Close(hand, washing\_machine)}
        \item \scriptsize \textcolor{gray}{TurnOn(hand, washing\_machine)}
        \item \scriptsize \textcolor{gray}{TurnOff(hand, washing\_machine)}
        \item \scriptsize \textcolor{gray}{Open(hand, washing\_machine)}
        \item \scriptsize \textcolor{gray}{Pick(hand, dark\_blue\_cloth)}
        \item \scriptsize \textcolor{gray}{Move(drying\_rack)}
        \item \scriptsize \textcolor{gray}{HangOn(hand, drying\_rack)}
    \end{itemize}

    \columnbreak

    \methodlabel{llm-as-planner (Cost: 18)}
    \begin{itemize}[leftmargin=8pt, noitemsep, topsep=0pt]
        \item \scriptsize Move(room\_601\_couch) \redcross \quad \textbf{\textcolor{red}{(Blocked by closed door)}}
        \item \scriptsize \textcolor{gray}{Pick(left\_hand, dark\_blue\_cloth)}
        \item \scriptsize \textcolor{gray}{Move(washing\_machine)}
        \item \scriptsize \textcolor{gray}{Open(right\_hand, washing\_machine)}
        \item \scriptsize \textcolor{gray}{PlaceIn(left\_hand, washing\_machine)}
        \item \scriptsize \textcolor{gray}{Close(right\_hand, washing\_machine)}
        \item \scriptsize \textcolor{gray}{TurnOn(right\_hand, washing\_machine)}
        \item \scriptsize \textcolor{gray}{TurnOff(right\_hand, washing\_machine)}
        \item \scriptsize \textcolor{gray}{Open(right\_hand, washing\_machine)}
        \item \scriptsize \textcolor{gray}{Pick(left\_hand, dark\_blue\_cloth)}
        \item \scriptsize \textcolor{gray}{Move(drying\_rack)}
        \item \scriptsize \textcolor{gray}{Pick(right\_hand, drying\_rack)}
        \item \scriptsize \textcolor{gray}{HangOn(right\_hand, dark\_blue\_cloth)}
    \end{itemize}
\end{multicols}
\hrule

\begin{multicols}{2}
    \methodlabel{SayPlan (Cost: 39)}
    \begin{itemize}[leftmargin=8pt, noitemsep, topsep=0pt]
        \item \scriptsize Move(pose\_26)
        \item \scriptsize OpenDoor(hand, door\_classroom\_601\_right)
        \item \scriptsize Move(room\_601\_couch)
        \item \scriptsize Pick(hand, clothing\_1)
        \item \scriptsize Move(sink\_table)
        \item \scriptsize TurnOn(hand, faucet\_1) \redcross \quad \textbf{\textcolor{red}{(Hand occupied)}}
        \item \scriptsize \textcolor{gray}{PlaceUnder(hand, faucet\_1)}
        \item \scriptsize \textcolor{gray}{TurnOff(hand, faucet\_1)}
        \item \scriptsize \textcolor{gray}{Move(drying\_rack)}
        \item \scriptsize \textcolor{gray}{HangOn(hand, wooden\_rod\_1)}
    \end{itemize}

    \columnbreak

    \methodlabel{SayPlan (Cost: 32)}
    \begin{itemize}[leftmargin=8pt, noitemsep, topsep=0pt]
        \item \scriptsize OpenDoor(right\_hand, door\_classroom\_601\_left\_pose\_24)
        \item \scriptsize Move(room\_601\_couch)
        \item \scriptsize Pick(left\_hand, cloth)
        \item \scriptsize Move(washing\_machine)
        \item \scriptsize Open(right\_hand, washing\_machine)
        \item \scriptsize PlaceIn(left\_hand, washing\_machine)
        \item \scriptsize Close(right\_hand, washing\_machine)
        \item \scriptsize TurnOn(right\_hand, washing\_machine)
        \item \scriptsize Open(right\_hand, washing\_machine)
        \item \scriptsize Pick(left\_hand, cloth)
        \item \scriptsize Move(drying\_rack)
        \item \scriptsize HangOn(left\_hand, drying\_rack)
    \end{itemize}
\end{multicols}
\hrule

\begin{multicols}{2}
    \methodlabel{DELTA (Cost: 0)}
    \begin{itemize}[leftmargin=8pt, noitemsep, topsep=0pt]
        \item \scriptsize \textit{No solution found.}
    \end{itemize}

    \columnbreak

    \methodlabel{DELTA (Cost: 31)}
    \begin{itemize}[leftmargin=8pt, noitemsep, topsep=0pt]
        \item \scriptsize Move(pose\_24)
        \item \scriptsize OpenDoor(left\_hand, door\_classroom\_601\_left\_pose\_24)
        \item \scriptsize Move(room\_601\_couch)
        \item \scriptsize Pick(left\_hand, cloth)
        \item \scriptsize Move(classroom\_601\_right)
        \item \scriptsize OpenDoor(right\_hand, door\_classroom\_601\_right\_pose\_26)
        \item \scriptsize Move(washing\_machine)
        \item \scriptsize Open(right\_hand, washing\_machine)
        \item \scriptsize PlaceIn(left\_hand, washing\_machine)
        \item \scriptsize Close(left\_hand, washing\_machine)
        \item \scriptsize TurnOn(left\_hand, washing\_machine)
        \item \scriptsize Open(right\_hand, washing\_machine)
        \item \scriptsize Pick(left\_hand, cloth)
        \item \scriptsize Move(drying\_rack)
        \item \scriptsize HangOn(left\_hand, drying\_rack)
    \end{itemize}
\end{multicols}
\end{examplebox}

\begin{examplebox}{41: Prepare two cups of coffee for meeting}
\textit{Instruction: Prepare two cups of coffee and place them on the meeting table.}

\begin{multicols}{2}
    \centering \textbf{Single-Arm with Door}
    \vfill\null \columnbreak
    \centering \textbf{Dual-Arm with Door}
\end{multicols}
\hrule

\begin{multicols}{2}
    \methodlabel{UniPlan (Cost: 73)}
    \begin{itemize}[leftmargin=8pt, noitemsep, topsep=0pt]
        \item \scriptsize Move(pose\_21)
        \item \scriptsize OpenDoor(hand, door\_office\_602\_pose\_21)
        \item \scriptsize Move(office\_602\_table)
        \item \scriptsize Pick(hand, green\_cup)
        \item \scriptsize Move(coffee\_maker)
        \item \scriptsize PlaceOn(hand, coffee\_maker)
        \item \scriptsize TurnOn(hand, coffee\_maker)
        \item \scriptsize Pick(hand, green\_cup)
        \item \scriptsize Move(meeting\_table)
        \item \scriptsize PlaceOn(hand, table)
        \item \scriptsize Move(office\_602\_table)
        \item \scriptsize Pick(hand, pink\_cup)
        \item \scriptsize Move(coffee\_maker)
        \item \scriptsize PlaceOn(hand, coffee\_maker)
        \item \scriptsize TurnOn(hand, coffee\_maker)
        \item \scriptsize Pick(hand, pink\_cup)
        \item \scriptsize Move(meeting\_table)
        \item \scriptsize PlaceOn(hand, table)
    \end{itemize}

    \columnbreak

    \methodlabel{UniPlan (Cost: 43)}
    \begin{itemize}[leftmargin=8pt, noitemsep, topsep=0pt]
        \item \scriptsize Move(pose\_21)
        \item \scriptsize OpenDoor(left\_hand, door\_office\_602\_pose\_21)
        \item \scriptsize Move(office\_602\_table)
        \item \scriptsize Pick(right\_hand, green\_cup)
        \item \scriptsize Pick(left\_hand, pink\_cup)
        \item \scriptsize Move(coffee\_maker)
        \item \scriptsize PlaceOn(left\_hand, coffee\_maker)
        \item \scriptsize TurnOn(left\_hand, coffee\_maker)
        \item \scriptsize Pick(left\_hand, pink\_cup)
        \item \scriptsize PlaceOn(right\_hand, coffee\_maker)
        \item \scriptsize TurnOn(right\_hand, coffee\_maker)
        \item \scriptsize Pick(right\_hand, green\_cup)
        \item \scriptsize Move(meeting\_table)
        \item \scriptsize PlaceOn(left\_hand, table)
        \item \scriptsize PlaceOn(right\_hand, table)
    \end{itemize}
\end{multicols}
\hrule

\begin{multicols}{2}
    \methodlabel{llm-as-planner (Cost: 91)}
    \begin{itemize}[leftmargin=8pt, noitemsep, topsep=0pt]
        \item \scriptsize Move(pose\_21)
        \item \scriptsize OpenDoor(hand, door\_office\_602\_pose\_21)
        \item \scriptsize Move(office\_602\_table)
        \item \scriptsize Pick(hand, green\_cup)
        \item \scriptsize PlaceOn(hand, table)
        \item \scriptsize Pick(hand, pink\_cup)
        \item \scriptsize PlaceOn(hand, table)
        \item \scriptsize Pick(hand, green\_cup)
        \item \scriptsize Move(coffee\_maker)
        \item \scriptsize PlaceOn(hand, coffee\_maker)
        \item \scriptsize TurnOn(hand, coffee\_maker)
        \item \scriptsize Pick(hand, green\_cup)
        \item \scriptsize Move(meeting\_table)
        \item \scriptsize PlaceOn(hand, table)
        \item \scriptsize Move(office\_602\_table)
        \item \scriptsize Pick(hand, pink\_cup)
        \item \scriptsize Move(coffee\_maker)
        \item \scriptsize PlaceOn(hand, coffee\_maker)
        \item \scriptsize TurnOn(hand, coffee\_maker)
        \item \scriptsize Pick(hand, pink\_cup)
        \item \scriptsize Move(meeting\_table)
        \item \scriptsize PlaceOn(hand, table)
    \end{itemize}

    \columnbreak

    \methodlabel{llm-as-planner (Cost: 43)}
    \begin{itemize}[leftmargin=8pt, noitemsep, topsep=0pt]
        \item \scriptsize Move(pose\_21)
        \item \scriptsize OpenDoor(left\_hand, door\_office\_602\_pose\_21)
        \item \scriptsize Move(office\_602\_table)
        \item \scriptsize Pick(left\_hand, green\_cup)
        \item \scriptsize Pick(right\_hand, pink\_cup)
        \item \scriptsize Move(coffee\_maker)
        \item \scriptsize PlaceOn(left\_hand, coffee\_maker)
        \item \scriptsize TurnOn(left\_hand, coffee\_maker)
        \item \scriptsize Pick(left\_hand, green\_cup)
        \item \scriptsize PlaceOn(right\_hand, coffee\_maker)
        \item \scriptsize TurnOn(right\_hand, coffee\_maker)
        \item \scriptsize Pick(right\_hand, pink\_cup)
        \item \scriptsize Move(meeting\_table)
        \item \scriptsize PlaceOn(left\_hand, table)
        \item \scriptsize PlaceOn(right\_hand, table)
    \end{itemize}
\end{multicols}
\hrule

\begin{multicols}{2}
    \methodlabel{SayPlan (Cost: 85)}
    \begin{itemize}[leftmargin=8pt, noitemsep, topsep=0pt]
        \item \scriptsize Move(pose\_38)
        \item \scriptsize OpenDoor(hand, door\_kitchen)
        \item \scriptsize Move(kitchen\_table\_2)
        \item \scriptsize Pick(hand, bowl\_1)
        \item \scriptsize Move(coffee\_maker)
        \item \scriptsize TurnOn(hand, coffee\_maker) \redcross \quad \textbf{\textcolor{red}{(Hand occupied)}}
        \item \scriptsize \textcolor{gray}{PlaceUnder(hand, coffee\_maker)}
        \item \scriptsize \textcolor{gray}{Move(meeting\_table)}
        \item \scriptsize \textcolor{gray}{PlaceOn(hand, meeting\_table)}
        \item \scriptsize \textcolor{gray}{Move(kitchen\_table\_2)}
        \item \scriptsize \textcolor{gray}{Pick(hand, bowl\_2)}
        \item \scriptsize \textcolor{gray}{Move(coffee\_maker)}
        \item \scriptsize \textcolor{gray}{TurnOn(hand, coffee\_maker)}
        \item \scriptsize \textcolor{gray}{PlaceUnder(hand, coffee\_maker)}
        \item \scriptsize \textcolor{gray}{Move(meeting\_table)}
        \item \scriptsize \textcolor{gray}{PlaceOn(hand, meeting\_table)}
    \end{itemize}

    \columnbreak

    \methodlabel{SayPlan (Cost: 40)}
    \begin{itemize}[leftmargin=8pt, noitemsep, topsep=0pt]
        \item \scriptsize Move(meeting\_table)
        \item \scriptsize Pick(left\_hand, cup\_1)
        \item \scriptsize Pick(right\_hand, pen\_holder\_1)
        \item \scriptsize Move(coffee\_maker)
        \item \scriptsize PlaceOn(left\_hand, coffee\_maker)
        \item \scriptsize TurnOn(left\_hand, coffee\_maker)
        \item \scriptsize Pick(left\_hand, cup\_1)
        \item \scriptsize PlaceOn(right\_hand, coffee\_maker)
        \item \scriptsize TurnOn(right\_hand, coffee\_maker) \redcross \quad \textbf{\textcolor{red}{(Invalid container type)}}
        \item \scriptsize \textcolor{gray}{Pick(right\_hand, pen\_holder\_1)}
        \item \scriptsize \textcolor{gray}{Move(meeting\_table)}
        \item \scriptsize \textcolor{gray}{PlaceOn(left\_hand, table\_1)}
        \item \scriptsize \textcolor{gray}{PlaceOn(right\_hand, table\_1)}
    \end{itemize}
\end{multicols}
\hrule

\begin{multicols}{2}
    \methodlabel{DELTA (Cost: 77)}
    \begin{itemize}[leftmargin=8pt, noitemsep, topsep=0pt]
        \item \scriptsize Move(pose\_21)
        \item \scriptsize OpenDoor(hand, door\_office\_602\_pose\_21)
        \item \scriptsize Move(office\_602\_table)
        \item \scriptsize Pick(hand, green\_cup)
        \item \scriptsize Move(coffee\_maker)
        \item \scriptsize PlaceOn(hand, coffee\_maker)
        \item \scriptsize TurnOn(hand, coffee\_maker)
        \item \scriptsize Pick(hand, green\_cup)
        \item \scriptsize Move(meeting\_table)
        \item \scriptsize PlaceOn(hand, table)
        \item \scriptsize Move(office\_602\_table)
        \item \scriptsize Pick(hand, pink\_cup)
        \item \scriptsize Move(coffee\_maker)
        \item \scriptsize PlaceOn(hand, coffee\_maker)
        \item \scriptsize TurnOn(hand, coffee\_maker)
        \item \scriptsize Pick(hand, pink\_cup)
        \item \scriptsize Move(meeting\_table)
        \item \scriptsize PlaceOn(hand, table)
    \end{itemize}

    \columnbreak

    \methodlabel{DELTA (Cost: 73)}
    \begin{itemize}[leftmargin=8pt, noitemsep, topsep=0pt]
        \item \scriptsize Move(pose\_21)
        \item \scriptsize OpenDoor(left\_hand, door\_office\_602\_pose\_21)
        \item \scriptsize Move(office\_602\_table)
        \item \scriptsize Pick(left\_hand, green\_cup)
        \item \scriptsize Move(coffee\_maker)
        \item \scriptsize PlaceOn(left\_hand, coffee\_maker)
        \item \scriptsize TurnOn(left\_hand, coffee\_maker)
        \item \scriptsize Pick(left\_hand, green\_cup)
        \item \scriptsize Move(meeting\_table)
        \item \scriptsize PlaceOn(left\_hand, table)
        \item \scriptsize Move(office\_602\_table)
        \item \scriptsize Pick(right\_hand, pink\_cup)
        \item \scriptsize Move(coffee\_maker)
        \item \scriptsize PlaceOn(right\_hand, coffee\_maker)
        \item \scriptsize TurnOn(left\_hand, coffee\_maker)
        \item \scriptsize Pick(right\_hand, pink\_cup)
        \item \scriptsize Move(meeting\_table)
        \item \scriptsize PlaceOn(right\_hand, table)
    \end{itemize}
\end{multicols}
\end{examplebox}

\newpage
\subsection{Prompts}
\paragraph{VLM Grounding Prompt}
The following prompt is used to guide the Vision-Language Model (VLM) in analyzing scene images, grounding objects, and generating the initial PDDL state and goal predicates.

\begin{sayplanbox}
You are an expert PDDL problem generator for a robotic agent. Your goal is to generate a PDDL problem file based on a given PDDL domain and images showing different parts of the whole scene.\\

\noindent Instructions: \{instructions\} \\
Available PDDL domain: \{domain\} \\
Image IDs: \{image\_ids\} \\

\noindent Output Requirements: \\
Your output should include four parts:
\begin{enumerate}[leftmargin=20pt, noitemsep, topsep=2pt]
    \item reasoning: Analyze the image and provide the reasoning.
    \item objects: Locate objects related to the task from the images respectively.
    \item init: Describe the PDDL init state from the images based on the given PDDL predicates respectively.
    \item goal: Generate the PDDL goal from human instructions.
\end{enumerate}

\noindent Notes:
\begin{enumerate}[leftmargin=20pt, noitemsep, topsep=2pt]
    \item When generating PDDL init, you should use type predicates, state predicates, spatial or position relationship predicates, and affordance predicates in order.
    \item You should generate objects and init based on each image respectively.
    \item When generating, you should analyze the goal, recognize the objects related to the goal, and only include relevant objects and init states.
    \item Some objects that seem unrelated to the goal but are necessary for solving the task (e.g., blocked or hindered objects) should be recognized and included as well. Just focus on the scene and the goal; IGNORE any information about the robot (e.g., hand\_free, holding, robot).
    \item The image\_ids correspond to the images in the provided order.
    \item When naming objects, prioritize color attributes as the primary grounding feature. This includes the color of the object's body, the color of its lid or cap, or the color of the contents inside.
\end{enumerate}

\noindent Output Format: \\
Return a JSON object with the following structure:
\begin{quote}
\begin{verbatim}
{
    "reasoning": "your analysis",
    "objects": {
        "image_id_1": ["red_apple_1", "green_apple_1", "yellow_bowl_1"],
        "image_id_2": ["red_apple_2", "orange_bowl_2"]
    },
    "init": ["(on_table yellow_bowl_1)", "(on_table orange_bowl_2)"],
    "goal": "(and (in red_apple_1 orange_bowl_2) (in red_apple_2 yellow_bowl_1))"
}
\end{verbatim}
\end{quote}
\end{sayplanbox}

\newpage
\paragraph{Textual Indexing Prompt}
The following prompt instructs the VLM to generate concise textual descriptions for specific nodes in the visual-topological map, serving as the semantic index for retrieval.

\begin{sayplanbox}
You are an intelligent visual perception assistant helping a robot to index scenes for semantic retrieval. The provided image shows a specific location or asset in the environment.\\

\noindent Task: \\
Generate a concise text description of the image.\\

\noindent Location or asset name: \{node\_name\} \\

\noindent Output Requirements: \\
Your output should include two parts:
\begin{enumerate}[leftmargin=20pt, noitemsep, topsep=2pt]
    \item reasoning: Analyze the image in detail. Identify the main objects, their quantities, and categories. You can be descriptive here to ensure accuracy of detection.
    \item description: A concise summary string of the visual content. Focus mainly on the existence and quantity of objects (e.g., "two stacked blocks", "a green drawer", "a fridge"). Ignore complex spatial relationships, precise coordinates, or minor decorative details. Keep it short and keyword-rich.
\end{enumerate}

\noindent Notes:
\begin{enumerate}[leftmargin=20pt, noitemsep, topsep=2pt]
    \item Remember to include the location name in the description, in a natural language style (expanding underscores if any).
    \item While keeping it short, all names of different objects MUST be included. Object names should be concise, unambiguous, and in a natural style.
\end{enumerate}

\noindent Output Format: \\
Return a JSON object with the following structure:
\begin{quote}
\begin{verbatim}
{
    "reasoning": "The image captures a wooden tabletop. On the left,
    there are two toy blocks stacked vertically. On the right, there
    is a small storage unit with three drawers.",
    "description": "two stacked blocks and three drawers on the
    room 601 table 1."
}
\end{verbatim}
\end{quote}
\end{sayplanbox}

\newpage
\paragraph{Task-Oriented Retrieval Prompt}
This prompt guides the LLM to filter the scene graph and select only the nodes that are functionally necessary for the given instruction, based on semantic reasoning.

\begin{sayplanbox}
You are the Scene Analyzer for a robotic agent. Your goal is to identify the Task-Relevant Nodes from a list of candidates based on a user instruction.\\

\noindent Input Format:
\begin{enumerate}[leftmargin=20pt, noitemsep, topsep=2pt]
    \item User Instruction: A natural language command (e.g., "Put the block in the drawer" or "Make a cup of milk").
    \item Candidate Nodes: A list of nodes with their IDs and visual descriptions.
\end{enumerate}

\noindent Selection Criteria (The "Must-Have" Rule): \\
Select a node IF AND ONLY IF it plays a specific role in the task execution plan:
\begin{enumerate}[leftmargin=20pt, noitemsep, topsep=2pt]
    \item Source: Where the object/ingredient is currently located.
    \item Target: The final destination specified in the instruction.
    \item Intermediate: Contains necessary tools (e.g., a cup for milk) or secondary ingredients.
\end{enumerate}

\noindent Critical Rules:
\begin{enumerate}[leftmargin=20pt, noitemsep, topsep=2pt]
    \item Exact Match: If the instruction explicitly names a location (e.g., "table\_3"), you MUST ignore similar objects on other tables (e.g., "table\_6").
    \item Implicit Reasoning: You must analyze possible tools to complete the task and include related locations, even if it is not explicitly stated in the instruction.
    \item Minimalism: Do not select nodes just because they are "nearby" or "look nice". Only select what is functionally required.
\end{enumerate}

\noindent Output Format: \\
Return a JSON object with the following structure:
\begin{quote}
\begin{verbatim}
{
    "reasoning": "Briefly explain the plan and why each node
    was selected or rejected.",
    "selected_nodes": ["node_id_1", "node_id_2"]
}
\end{verbatim}
\end{quote}

\noindent Input User Instruction: \{instruction\} \\
Candidate Nodes: \{candidate\_nodes\}
\end{sayplanbox}

\newpage
\paragraph{Textual Map Construction Prompt}
This prompt directs the VLM to analyze images of fixed assets and generate a detailed, structured Scene Graph (SG) in JSON format, capturing object attributes, affordances, and spatial relations.

\begin{sayplanbox}
You are an expert robot perception system designed to generate a semantic Scene Graph (SG) for robotic planning in a large-scale environment.\\

\noindent Task: \\
Analyze the provided image, which represents a specific fixed asset (e.g., a desk, shelf, or cabinet) in the environment. Your goal is to detect all objects positioned on or within this asset and generate a structured JSON output representing the local scene graph.\\

\noindent Output Structure Requirements: \\
The output must be a valid JSON object. The structure should combine the information requirements of standard robotic planners (like SayPlan and DELTA) while capturing detailed spatial context.

Use the following JSON schema. Your output should include two parts:
\begin{enumerate}[leftmargin=20pt, noitemsep, topsep=2pt]
    \item reasoning: Analyze the image in detail. Identify the main objects, their quantities, and categories. You can be descriptive here to ensure accuracy of detection.
    \item objects: For each detected object, provide its unique identifier, category, affordance, state, attributes, relations to other objects, and other information. The objects should be described in a structured format as shown in the example below.
\end{enumerate}

\noindent JSON Schema Example:
\begin{quote}
\begin{verbatim}
{
  "coffee_machine": {
    "category": "electrical",
    "affordance": ["turn on", "turn off", "release"],
    "state": "off",
    "attributes": ["black", "automatic"],
    "relation": {
      "on": "table"
    }
  },
  "unique_object_id_2": { ... },
  ...
}
\end{verbatim}
\end{quote}

\noindent Output Format: \\
Return a JSON object with the following structure:
\begin{quote}
\begin{verbatim}
{
  "reasoning": "The image captures a wooden tabletop. On the left,
  there are two toy blocks stacked vertically. On the right, there
  is a small storage unit with three drawers.",
  "objects": { "JSON Schema..." }
}
\end{verbatim}
\end{quote}
\end{sayplanbox}

\newpage
\paragraph{LLM-as-Planner Prompt}
This prompt is used for the LLM-based baseline, providing the model with a comprehensive list of primitive actions, physical constraints, and the environment state to generate executable plans.

\begin{sayplanbox}
You are an expert robotic task planner. Your goal is to generate an executable action sequence for a single-arm or dual-arm robot to complete a specific instruction. When completing the task, you need to consider constraints among actions. Multiple solutions would be feasible; you should complete the task in an optimal way as possible.\\

\noindent Available Actions \& Parameters:
\begin{description}[leftmargin=5pt, style=unboxed, font=\small\bfseries\ttfamily]
    \item[pick(robot, hand, obj):] Robot `robot` picks up object `obj` using `hand`.
    \begin{itemize}[leftmargin=15pt, label=-, noitemsep, topsep=0pt]
        \item You cannot pick an object if a hand is holding something.
        \item You cannot pick an object if it is under other objects.
    \end{itemize}
    
    \item[place\_in(robot, hand, obj):] Robot `robot` places the object held in `hand` into container `obj` (e.g., a drawer).
    \begin{itemize}[leftmargin=15pt, label=-, noitemsep, topsep=0pt]
        \item You cannot place an object into a container if it is closed.
    \end{itemize}

    \item[place\_on(robot, hand, obj):] Robot `robot` places the object held in `hand` onto surface `obj` (e.g., a table).
    \begin{itemize}[leftmargin=15pt, label=-, noitemsep, topsep=0pt]
        \item You cannot place an object onto another object if the latter is under other objects.
    \end{itemize}

    \item[open(robot, hand, obj):] Robot `robot` opens container `obj` using `hand`.
    \begin{itemize}[leftmargin=15pt, label=-, noitemsep, topsep=0pt]
        \item You cannot open an object if a hand is holding something.
    \end{itemize}

    \item[close(robot, hand, obj):] Robot `robot` closes container `obj` using `hand`.
    \begin{itemize}[leftmargin=15pt, label=-, noitemsep, topsep=0pt]
        \item You cannot close an object if a hand is holding something.
    \end{itemize}

    \item[place\_under(robot, hand, obj):] Robot `robot` places the object held in `hand` under `obj` (e.g., a faucet), while still holding it.
    \begin{itemize}[leftmargin=15pt, label=-, noitemsep, topsep=0pt]
        \item If you place an opened kettle or a cup under a faucet that is turned on, it will be filled with water.
        \item If you place something under a faucet that is turned on, the object will be washed.
    \end{itemize}

    \item[pour(robot, hand, obj):] Robot `robot` pours from the object held in `hand` into object `obj`.
    \begin{itemize}[leftmargin=15pt, label=-, noitemsep, topsep=0pt]
        \item You cannot pour from an object if it is empty or closed.
        \item You cannot pour into an object if it is closed.
    \end{itemize}

    \item[cut(robot, hand, obj):] Robot `robot` cuts object `obj` using the object held in `hand` (e.g., a knife).
    \item[stir(robot, hand, obj):] Robot `robot` stirs object `obj` using the object held in `hand` (e.g., a spoon).
    \item[scoop(robot, hand, obj):] Robot `robot` scoops from object `obj` using the object held in `hand` (e.g., a rice paddle).
    \begin{itemize}[leftmargin=15pt, label=-, noitemsep, topsep=0pt]
        \item You cannot scoop from an object if it is closed.
    \end{itemize}
    
    \item[fold(robot, hand, obj):] Robot `robot` folds object `obj` using `hand`.
    \item[wipe(robot, hand, obj):] Robot `robot` wipes object `obj` using the object held in `hand` (e.g., a cloth).
    
    \item[turn\_on(robot, hand, obj):] Robot `robot` turns on object `obj` (e.g., a faucet, microwave) using `hand`.
    \begin{itemize}[leftmargin=15pt, label=-, noitemsep, topsep=0pt]
        \item If you turn on a closed microwave, the object inside will be heated.
        \item If you turn on a faucet, water will flow out.
        \item If you turn on a coffee machine, coffee will flow out.
        \item If you turn on a closed kettle on its base, the water inside will be heated.
    \end{itemize}

    \item[turn\_off(robot, hand, obj):] Robot `robot` turns off object `obj` (e.g., a faucet, lamp) using `hand`.
    \item[hang\_on(robot, hand, obj):] Robot `robot` hangs object `obj` (e.g., a towel) on the object held in `hand` (e.g., a drying rack).
    
    \item[open\_door(robot, hand, door\_name):] Robot `robot` opens a closed door `door\_name` using `hand`. The door\_name is like "door\_\{node\_name\_1\}\_\{node\_name\_2\}".
    \begin{itemize}[leftmargin=15pt, label=-, noitemsep, topsep=0pt]
        \item You cannot open a door if a hand is holding something.
    \end{itemize}

    \item[move(robot, to\_node):] Robot `robot` moves to location `to\_node`.
    \begin{itemize}[leftmargin=15pt, label=-, noitemsep, topsep=0pt]
        \item You cannot move to a node if the door between the current and target nodes is closed.
        \item You cannot move to a node if the current and target nodes are not connected.
    \end{itemize}
\end{description}

\noindent Environment Configuration:
\begin{enumerate}[leftmargin=20pt, noitemsep, topsep=2pt]
    \item Available Robot(s): \{robot\_name\}
    \item Available Hand(s): \{hand\_names\}
    \item Robot Initial Location: \{robot\_init\_node\}
    \item Scene Graph Topology (Connectivity): \{scene\_graph\}
    \item Task-Relevant Key Nodes: \{key\_nodes\}
\end{enumerate}

\noindent Task Instruction: \\
\{instruction\} \\

\noindent Notes:
\begin{enumerate}[leftmargin=20pt, noitemsep, topsep=2pt]
    \item The key nodes correspond to the images in the order provided.
    \item When naming objects, prioritize color attributes as the primary grounding feature. This includes the color of the object's body, its lid or cap, or the contents inside.
    \item Maintain strict naming consistency for the same object throughout the entire plan.
\end{enumerate}

\noindent Output Requirement: \\
Please output the response in JSON format containing two parts:
\begin{enumerate}[leftmargin=20pt, noitemsep, topsep=2pt]
    \item "reasoning": Step-by-step thinking to decompose the task. Analyze the logical dependencies (e.g., "I need to go to node A to pick up item X before going to node B") and path planning constraints based on the topology.
    \item "plan": A list of function calls representing the complete action sequence.
\end{enumerate}

\noindent Example Output Format:
\begin{quote}
\begin{verbatim}
{
    "reasoning": "First, I need to move the robot to the
    Kitchen_Table to pick up the apple. Since the fridge is
    closed, I must move to the Fridge_Node, open the fridge,
    and then place the apple inside.",
    "plan": [
        "move(robot_a, Kitchen_Table)",
        "pick(robot_a, left_hand, apple)",
        "move(robot_a, Fridge_Node)",
        "open(robot_a, right_hand, fridge)",
        "place_in(robot_a, left_hand, fridge)"
    ]
}
\end{verbatim}
\end{quote}
\end{sayplanbox}

\newpage
\paragraph{SayPlan Semantic Search Prompt}
This prompt directs the SayPlan baseline to navigate the hierarchical scene graph, expanding or contracting nodes to locate task-relevant assets before planning.

\begin{sayplanbox}
You are the Semantic Search Module for a robotic agent. Your ONLY goal is to explore the Scene Graph to locate all relevant objects required for the user's instruction. You do NOT generate the plan to manipulate objects yet; you only manipulate the Graph View.\\

\noindent Input Data:
\begin{enumerate}[leftmargin=20pt, noitemsep, topsep=2pt]
    \item Instruction: "\{instruction\}"
    \item Current Graph State: A hierarchical view of the environment.
    \begin{enumerate}[leftmargin=15pt, label=(\alph*), noitemsep, topsep=0pt]
        \item Nodes can be "collapsed" (summary view) or "expanded" (detailed view).
        \item "visible\_assets": Objects currently visible to you.
        \item "connected\_rooms": Nearby rooms you can access.
    \end{enumerate}
    \item Memory: A list of nodes you have already explored.
\end{enumerate}

\noindent Available Graph Actions:
\begin{enumerate}[leftmargin=20pt, noitemsep, topsep=2pt]
    \item expand(node\_name):
    \begin{enumerate}[leftmargin=15pt, label=(\alph*), noitemsep, topsep=0pt]
        \item Use this to look inside a "collapsed" Room or Asset.
        \item Strategy: Use semantic reasoning. If looking for an "apple", expand the "kitchen" or "fridge", not the "bathroom".
        \item CONSTRAINT: You can ONLY expand nodes listed in the "expandable\_nodes" field of the current graph. Do NOT try to expand objects found inside assets (e.g., if you see 'fridge\_1' inside 'fridge', do not expand 'fridge\_1').
    \end{enumerate}
    \item contract(node\_name):
    \begin{enumerate}[leftmargin=15pt, label=(\alph*), noitemsep, topsep=0pt]
        \item Use this to hide details of an "expanded" node.
        \item Strategy: If you expanded a node and found NOTHING relevant to the instruction, contract it immediately to keep the graph clean.
    \end{enumerate}
    \item terminate:
    \begin{enumerate}[leftmargin=15pt, label=(\alph*), noitemsep, topsep=0pt]
        \item Use this ONLY when you have successfully located ALL objects mentioned or implied by the instruction in the visible\_assets or internal\_assets of the current graph.
    \end{enumerate}
\end{enumerate}

\noindent Current Graph State: \\
\{graph\_state\_json\} \\

\noindent Memory (Visited): \\
\{memory\_list\} \\

\noindent Output Requirement: \\
Please output a JSON object with the following fields:
\begin{enumerate}[leftmargin=20pt, noitemsep, topsep=2pt]
    \item "reasoning": Step-by-step reasoning. Analyze the instruction, identify missing objects, and decide which node is most likely to contain them based on common sense.
    \item "command": One of ["expand", "contract", "terminate"].
    \item "target": The exact name of the node to operate on (only if the command is expand/contract). If the command is terminate, output "terminate" as well.
\end{enumerate}

\noindent Example Output Format:
\begin{quote}
\begin{verbatim}
{
   "reasoning": "The user wants to 'wash the apple'. I see
   'microwave_table' and 'kitchen'. Apples are usually in
   the kitchen. I need to check inside the kitchen.",
   "command": "expand",
   "target": "kitchen"
}
\end{verbatim}
\end{quote}
\end{sayplanbox}

\newpage
\paragraph{SayPlan Iterative Planner Prompt}
This prompt is designed for the iterative planning phase. It generates an initial plan or corrects a failed plan based on execution feedback, handling constraints like closed doors and logical dependencies.

\begin{sayplanbox}
You are an expert robotic task planner.

\noindent Goal:
\begin{enumerate}[leftmargin=20pt, noitemsep, topsep=2pt]
    \item If this is the first attempt, generate an executable action sequence to complete the instruction.
    \item If there is Execution History provided, you must FIX the previous plan based on the reported error. Analyze exactly which step failed and why (e.g., failed to open a door, moved to a blocked node, hand occupied), and generate a corrected plan.
\end{enumerate}

You are provided with a Hierarchical Scene Graph representing the current environment state. Multiple solutions may be feasible; you should complete the task in an optimal way.

\noindent Environment Configuration:
\begin{enumerate}[leftmargin=20pt, noitemsep, topsep=2pt]
    \item Available Robot(s): \{robot\_name\}
    \item Available Hand(s): \{hand\_names\}
    \item Robot State: Refer to the ``current\_location'' field in the Scene Graph State below to know where the robot starts.
    \item Scene Graph State (Explored): \{graph\_state\_json\}
\end{enumerate}

\noindent Available Actions \& Parameters:
\begin{description}[leftmargin=5pt, style=unboxed, font=\small\bfseries\ttfamily]
    \item[pick(robot, hand, obj):] Robot `robot` picks up object `obj` using `hand`.
    \begin{itemize}[leftmargin=15pt, label=-, noitemsep, topsep=0pt]
        \item You cannot pick an object if a hand is holding something.
        \item You cannot pick an object if it is under other objects.
    \end{itemize}

    \item[place\_in(robot, hand, obj):] Robot `robot` places the object held in `hand` into container `obj`.
    \begin{itemize}[leftmargin=15pt, label=-, noitemsep, topsep=0pt]
        \item You cannot place an object into a container if it is closed.
    \end{itemize}

    \item[place\_on(robot, hand, obj):] Robot `robot` places the object held in `hand` onto surface `obj`.
    \begin{itemize}[leftmargin=15pt, label=-, noitemsep, topsep=0pt]
        \item You cannot place an object onto another object if the latter is under other objects.
    \end{itemize}

    \item[open(robot, hand, obj):] Robot `robot` opens container `obj` using `hand`.
    \begin{itemize}[leftmargin=15pt, label=-, noitemsep, topsep=0pt]
        \item You cannot open an object if a hand is holding something.
    \end{itemize}

    \item[close(robot, hand, obj):] Robot `robot` closes container `obj` using `hand`.
    \begin{itemize}[leftmargin=15pt, label=-, noitemsep, topsep=0pt]
        \item You cannot close an object if a hand is holding something.
    \end{itemize}

    \item[place\_under(robot, hand, obj):] Robot `robot` places the object held in `hand` under `obj`, while still holding it.
    \begin{itemize}[leftmargin=15pt, label=-, noitemsep, topsep=0pt]
        \item If you place an opened kettle or a cup under a turned-on faucet, it will be filled with water.
        \item If you place something under a turned-on faucet, it will be washed.
    \end{itemize}

    \item[pour(robot, hand, obj):] Robot `robot` pours from the object held in `hand` into object `obj`.
    \begin{itemize}[leftmargin=15pt, label=-, noitemsep, topsep=0pt]
        \item You cannot pour from an object if it is empty or closed.
        \item You cannot pour into an object if it is closed.
    \end{itemize}

    \item[cut(robot, hand, obj):] Robot `robot` cuts object `obj` using the object held in `hand`.
    \item[stir(robot, hand, obj):] Robot `robot` stirs object `obj` using the object held in `hand`.
    \item[scoop(robot, hand, obj):] Robot `robot` scoops from object `obj` using the object held in `hand`.
    \begin{itemize}[leftmargin=15pt, label=-, noitemsep, topsep=0pt]
        \item You cannot scoop from a closed object.
    \end{itemize}

    \item[fold(robot, hand, obj):] Robot `robot` folds object `obj` using `hand`.
    \item[wipe(robot, hand, obj):] Robot `robot` wipes object `obj` using the object held in `hand`.

    \item[turn\_on(robot, hand, obj):] Robot `robot` turns on object `obj` using `hand`.
    \begin{itemize}[leftmargin=15pt, label=-, noitemsep, topsep=0pt]
        \item If you turn on a closed microwave, the object inside will be heated.
        \item If you turn on a faucet, water will flow out.
        \item If you turn on a coffee machine, coffee will flow out.
    \end{itemize}

    \item[turn\_off(robot, hand, obj):] Robot `robot` turns off object `obj` using `hand`.
    \item[hang\_on(robot, hand, obj):] Robot `robot` hangs object `obj` on the object held in `hand`.

    \item[open\_door(robot, hand, door\_name):] Robot `robot` opens a closed door `door\_name` using `hand`.
    \begin{itemize}[leftmargin=15pt, label=-, noitemsep, topsep=0pt]
        \item The `door\_name` format is like ``door\_\{room\_name\}''.
        \item You cannot open a door if a hand is holding something.
        \item Constraint: You must move to the door location (room name) before opening it.
    \end{itemize}

    \item[move(robot, to\_node):] Robot `robot` moves to location `to\_node`.
    \begin{itemize}[leftmargin=15pt, label=-, noitemsep, topsep=0pt]
        \item Intra-Area Move: If `to\_node` is in the same area, the path is usually clear.
        \item Inter-Area Move: If `to\_node` is in a different room, check the `connected\_rooms` list.
        \item Constraint: You cannot move to a node if the door between the current and target nodes is closed.
        \item Note: `to\_node` should be a room name or an asset name. Ignore other nodes like ``Common\_Area\_zone\_9''.
    \end{itemize}
\end{description}

\noindent Task Instruction: \\
\{instruction\} \\

\noindent Critical Planning Constraints:
\begin{enumerate}[leftmargin=20pt, noitemsep, topsep=2pt]
    \item The ``Closed Door'' Rule:
    \begin{enumerate}[leftmargin=15pt, label=(\alph*), noitemsep, topsep=0pt]
        \item Look at connected\_rooms in the provided Scene Graph.
        \item If a room has ``door'': ``closed'', you CANNOT directly move to any assets inside it.
        \item You MUST generate open\_door first.
    \end{enumerate}
    \item Initial Navigation:
    \begin{enumerate}[leftmargin=15pt, label=(\alph*), noitemsep, topsep=0pt]
        \item You are currently at ``current\_location''. The first action is almost always a move to the target asset or door.
        \item Do not assume you are already standing in front of the target object.
    \end{enumerate}
    \item Object Visibility:
    \begin{enumerate}[leftmargin=15pt, label=(\alph*), noitemsep, topsep=0pt]
        \item You can only manipulate objects listed in visible\_assets (current area) or internal\_assets (inside an expanded room).
    \end{enumerate}
\end{enumerate}

\noindent Notes:
\begin{enumerate}[leftmargin=20pt, noitemsep, topsep=2pt]
    \item It is not necessary to use the same name as in the provided scene graph. When naming objects, prioritize color attributes as the primary grounding feature.
    \item Maintain strict naming consistency for the same object throughout the entire plan.
\end{enumerate}

\noindent Execution History (Accumulated Failures): \\
The following plans were attempted in previous steps and FAILED. Analyze the errors carefully. Do NOT generate the same plans again. Fix the logic issues based on the feedback. \\
\{execution\_history\}

\noindent Output Requirement: \\
Please output the response in JSON format containing two parts:
\begin{enumerate}[leftmargin=20pt, noitemsep, topsep=2pt]
    \item "reasoning": Step-by-step thinking to decompose the task. Analyze logical dependencies and path planning constraints based on the topology.
    \item "plan": A list of function calls representing the complete action sequence.
\end{enumerate}

\noindent Example Output Format:
\begin{quote}
\begin{verbatim}
{
    "reasoning": "First, I need to get the apple. It is in the
    kitchen. I am in the Common_Area. The kitchen door is
    closed, so I must open it first. Then I will move to the
    kitchen table, pick the apple, move to the fridge, open
    it, and place the apple inside.",
    "plan": [
        "move(robot, kitchen)",
        "open_door(robot, right_hand, door_kitchen)",
        "move(robot, kitchen_table_1)",
        "pick(robot, left_hand, red_apple)",
        "move(robot, fridge)",
        "open(robot, right_hand, fridge)",
        "place_in(robot, left_hand, fridge)"
    ]
}
\end{verbatim}
\end{quote}
\end{sayplanbox}

\newpage
\paragraph{DELTA Domain Selection Prompt}
This prompt guides the DELTA baseline to prune and select a minimal sufficient PDDL domain from a provided reference, adhering to specific configuration flags.

\begin{sayplanbox}
You are an expert PDDL domain editor for a mobile manipulation robot in indoor environments. Given a natural-language task instruction, two configuration flags, and a provided ground-truth reference PDDL domain, your goal is to output a complete PDDL domain file that is minimal but sufficient to solve the task, by only selecting/pruning from the reference domain and doing minimal mechanical adaptation if strictly necessary.

\noindent IMPORTANT:
\begin{enumerate}[leftmargin=20pt, noitemsep, topsep=2pt]
    \item The reference domain is authoritative: you should NOT invent new manipulation predicates or new manipulation actions.
    \item Your main job is selection/pruning (keep only the needed predicates/actions and remove the rest).
    \item You may do minimal edits only when strictly necessary to satisfy configuration flags (doors / multi-arms) or to keep the domain syntactically valid and planner-friendly.
\end{enumerate}

\noindent Inputs:
\begin{enumerate}[leftmargin=20pt, noitemsep, topsep=2pt]
    \item Task Instruction: \{instruction\}
    \item Configuration Flags:
    \begin{enumerate}[leftmargin=15pt, label=(\alph*), noitemsep, topsep=0pt]
        \item Require Doors: \{require\_doors\} \\
        - If Require Doors = True: the final domain MUST support doors and movement respecting door state. \\
        - If Require Doors = False: the final domain MUST NOT contain door predicates/actions; movement only uses connectivity.
        \item Dual Arms: \{dual\_arms\} \\
        - If Dual Arms = True: the domain should be able to represent independent hand occupancy if the reference domain already supports it. \\
        - If the reference domain is single-hand, you may keep the single-hand model.
    \end{enumerate}
    \item Reference Domain File (Ground-Truth): \{reference\_domain\_pddl\}
\end{enumerate}

\noindent Hard constraints (must follow):
\begin{enumerate}[leftmargin=20pt, noitemsep, topsep=2pt]
    \item NO PDDL Types (do NOT use typing) \\
    Your final domain MUST follow the reference domain's ``unary type predicate'' style and remain untyped:
    \begin{enumerate}[leftmargin=15pt, label=(\alph*), noitemsep, topsep=0pt]
        \item Do NOT include \texttt{(:types ...)}.
        \item Do NOT use typed variables like \texttt{?x - obj} / \texttt{?n - node}.
        \item Do NOT rely on \texttt{:typing} requirement. Remove \texttt{:typing} unless the reference domain absolutely requires it.
    \end{enumerate}

    \item Selection/pruning only
    \begin{enumerate}[leftmargin=15pt, label=(\alph*), noitemsep, topsep=0pt]
        \item Keep the reference domain's predicate names and action schemas whenever possible.
        \item Remove predicates and actions that are not needed for the given instruction.
        \item Do NOT create new task-specific predicates/actions if the reference already has equivalent ones.
        \item Do NOT add ``mapping comments'' or ``grounding signatures''. None are needed.
        \item Ensure definition consistency: predicates in operators MUST be defined. Especially note \texttt{pick\_from\_table} and \texttt{place\_on\_table}.
    \end{enumerate}

    \item Minimal navigation support \\
    The task is mobile manipulation, so the final domain must include navigation if and only if the task requires moving between locations.
    \begin{enumerate}[leftmargin=15pt, label=(\alph*), noitemsep, topsep=0pt]
        \item If the reference domain already contains navigation predicates/actions, keep the minimal subset (e.g., \texttt{at}, \texttt{connected}, \texttt{move}).
        \item If the reference domain does NOT contain navigation but the task clearly requires it, you may add the smallest possible navigation modeling:
        \begin{itemize}[leftmargin=10pt, label=-, noitemsep, topsep=0pt]
            \item predicates: \texttt{(at ?r ?n)}, \texttt{(connected ?from ?to)}
            \item action: \texttt{move} with standard STRIPS preconditions/effects
        \end{itemize}
        \item Keep it minimal; do not add extra navigation machinery.
    \end{enumerate}

    \item Doors (ONLY if Require Doors = True)
    \begin{enumerate}[leftmargin=15pt, label=(\alph*), noitemsep, topsep=0pt]
        \item If Require Doors = True and the reference domain already has door modeling, keep the minimal door subset.
        \item If Require Doors = True and the reference domain has no door modeling, you may add the smallest possible door support:
        \begin{itemize}[leftmargin=10pt, label=-, noitemsep, topsep=0pt]
            \item a door state predicate between node pairs (e.g., \texttt{(door\_open ?n1 ?n2)} or \texttt{(door\_closed ?n1 ?n2)} -- pick ONE style)
            \item an \texttt{open\_door} action
            \item \texttt{move} must respect door openness
        \end{itemize}
        \item If Require Doors = False: do NOT include any door predicates/actions.
    \end{enumerate}

    \item Planner-friendliness
    \begin{enumerate}[leftmargin=15pt, label=(\alph*), noitemsep, topsep=0pt]
        \item Avoid advanced features unless already present and truly required by the kept subset.
        \item Prefer no conditional effects.
        \item Prefer no quantifiers.
        \item Keep \texttt{:requirements} minimal and consistent with the kept operators.
    \end{enumerate}
\end{enumerate}

\noindent Output Requirement: \\
Output JSON with two fields:
\begin{enumerate}[leftmargin=20pt, noitemsep, topsep=2pt]
    \item "reasoning": brief, step-by-step reasoning describing which actions/predicates you kept and why, which parts you removed, whether you had to minimally add navigation/door support, and confirmation that you did not use types.
    \item "domain": a single complete PDDL domain file as a multiline string.
\end{enumerate}

\noindent Example Output Format:
\begin{quote}
\begin{verbatim}
{
  "reasoning": "...",
  "domain": "(define (domain ...)\n ...\n)"
}
\end{verbatim}
\end{quote}
\end{sayplanbox}

\newpage
\paragraph{DELTA Scene Graph Pruning Prompt}
This prompt guides the DELTA baseline to filter the scene graph, selecting only the objects and assets strictly necessary for the task while maintaining logical dependencies.

\begin{sayplanbox}
You are an expert robotic task planner. Given a list of items located in various assets and a task instruction, your goal is to filter out irrelevant items and keep only the ones necessary to complete the task. When completing the selection, you need to consider logical dependencies implied by the instruction (e.g., needing a container to access an item, or needing a tool to perform an action). Multiple selections could be feasible; you should keep the selection minimal but sufficient to complete the task.

\noindent Inputs:
\begin{enumerate}[leftmargin=20pt, noitemsep, topsep=2pt]
    \item Task Instruction: \{instruction\}
    \item Available Items in Environment (grouped by asset): \{items\}
\end{enumerate}

\noindent Input Format Note (important):
\begin{enumerate}[leftmargin=15pt, label=(\alph*), noitemsep, topsep=0pt]
    \item The \{items\} field is a JSON dictionary.
    \item Each key is an asset\_node\_name (i.e., a location/container/surface node).
    \item Each value is a list of object\_name identifiers currently associated with that asset.
    \item Objects must remain under their original asset key; do not move objects across assets.
\end{enumerate}

\noindent Selection Criteria:
\begin{enumerate}[leftmargin=20pt, noitemsep, topsep=2pt]
    \item Select items explicitly mentioned in the instruction.
    \item Select containers or surfaces that are necessary to interact with to complete the task (e.g., if you need an item inside a fridge, select the fridge asset's relevant contents).
    \item Select implicit tools required for the task (e.g., ``cut'' implies a knife; ``clean'' may imply a sponge/cloth).
    \item Handle name collisions safely: items may share similar names across assets. Only keep an item under the asset where it is listed.
    \item Keep the output minimal: do not include extra objects unless they are required as tools, target containers/surfaces, or explicitly mentioned.
\end{enumerate}

\noindent Output Requirement: \\
Please output the response in JSON format containing two parts:
\begin{enumerate}[leftmargin=20pt, noitemsep, topsep=2pt]
    \item "reasoning": Step-by-step thinking to justify which objects are needed and why, including any implicit tools/containers required by the instruction.
    \item "relevant\_items": A dictionary mapping asset\_node\_name to a list of kept object identifiers under that asset.
\end{enumerate}

\noindent Rules for "relevant\_items":
\begin{enumerate}[leftmargin=15pt, label=(\alph*), noitemsep, topsep=0pt]
    \item Only include assets that have at least one kept object.
    \item Only include object names that appear in the provided list under that asset.
    \item Do NOT output any objects not present in the environment list.
    \item Do NOT create new asset keys that are not in \{items\}.
    \item If an asset itself is a necessary target container/surface AND it appears as an object identifier in that asset's object list, include it in that asset's kept list.
    \item If the same object name appears under multiple assets, keep it only under the asset(s) where it is truly needed; never relocate it.
\end{enumerate}

\noindent Example Output Format:
\begin{quote}
\begin{verbatim}
{
  "reasoning": "The instruction requires cutting the apple, so I
  must keep the apple and a knife. If the apple is stored in a
  fridge/drawer, I must also keep the relevant container asset
  contents needed to access it. I will keep only the minimal set
  of items required.",
  "relevant_items": {
    "office_table_1": ["apple_1", "knife_2"],
    "fridge_1": ["milk_1"]
  }
}
\end{verbatim}
\end{quote}
\end{sayplanbox}

\newpage
\paragraph{DELTA Problem Generation Prompt}
This prompt directs the DELTA baseline to instantiate a valid PDDL problem file by grounding the provided domain and task-relevant scene graph into concrete objects, initial states, and goal conditions.

\begin{sayplanbox}
You are an excellent PDDL problem file generator for a mobile manipulation robot. Given (1) a pruned, task-relevant Scene Graph (topology + node-local objects), (2) a PDDL domain file, and (3) a natural-language task instruction, you will generate a valid PDDL problem file.

You must strictly follow the modeling choices of the provided domain:
\begin{enumerate}[leftmargin=20pt, noitemsep, topsep=2pt]
    \item Use the same predicate names and argument structure as the domain.
    \item If the domain uses \texttt{(:types)}, then declare objects with types in \texttt{(:objects)}.
    \item If the domain uses unary ``type predicates'' (e.g., \texttt{(table ?t)}), then list objects without typing in \texttt{(:objects)} and assert type predicates in \texttt{(:init)}.
    \item Do NOT invent new predicates not present in the domain.
\end{enumerate}

\noindent Inputs:
\begin{enumerate}[leftmargin=20pt, noitemsep, topsep=2pt]
    \item Task Instruction / Goal (natural language): \{instruction\}
    \item Robot Configuration:
    \begin{enumerate}[leftmargin=15pt, label=(\alph*), noitemsep, topsep=0pt]
        \item Robot Name: \{robot\_name\}
        \item Available Hands: \{hand\_list\}
        \item Robot Initial Node: \{robot\_init\_node\}
    \end{enumerate}
    \item PDDL Domain File (reference): \{domain\_pddl\}
    \item Pruned Scene Graph (task-relevant, complete): \{pruned\_scene\_graph\_json\}
\end{enumerate}

\noindent Input Format Note:
\begin{enumerate}[leftmargin=20pt, noitemsep, topsep=2pt]
    \item The scene graph JSON contains a list of nodes. Each node has a name, a type in \{room, pose, asset\}, neighbors (adjacency), and optionally node-local objects.
    \item Nodes of type room, pose, or asset are all valid navigation targets for the robot. Objects (items) are NOT navigable nodes and must never be used as move targets.
\end{enumerate}

\noindent Object Naming Note (important): \\
When naming or selecting object identifiers for goals, prioritize color attributes as the primary grounding feature. This includes the color of the object's body, its lid/cap, or the contents inside.

\noindent Scene Graph Interpretation Rules:
\begin{enumerate}[leftmargin=20pt, noitemsep, topsep=2pt]
    \item Navigation nodes:
    \begin{enumerate}[leftmargin=15pt, label=(\alph*), noitemsep, topsep=0pt]
        \item Treat every node (room / pose / asset) as a location object the robot can move to.
        \item The robot can only manipulate objects/devices that are at the same node as the robot (locality constraint), if such a predicate exists in the domain (e.g., \texttt{(at robot node)}).
    \end{enumerate}
    \item Manipulation locality constraint (important):
    \begin{enumerate}[leftmargin=15pt, label=(\alph*), noitemsep, topsep=0pt]
        \item All nodes (room / pose / asset) are navigable and can appear as move targets.
        \item However, for any action other than \texttt{move} and \texttt{open\_door} (e.g., \texttt{pick}, \texttt{place}, \texttt{open}, \texttt{close}), the robot must be located at an asset node.
        \item Even if an asset node looks like a physical object (e.g., fridge, microwave), in this setting it is treated as a location node for manipulation preconditions.
    \end{enumerate}
    \item Connectivity:
    \begin{enumerate}[leftmargin=15pt, label=(\alph*), noitemsep, topsep=0pt]
        \item For each undirected edge A-B in the scene graph, add both directions in the init section using the domain's predicate for connectivity (e.g., \texttt{connected}, \texttt{neighbor}).
        \item Example pattern: \texttt{(connected A B)} and \texttt{(connected B A)}.
        \item If the domain uses a different predicate name, follow the domain exactly.
    \end{enumerate}
    \item Doors (ONLY if the domain includes door modeling):
    \begin{enumerate}[leftmargin=15pt, label=(\alph*), noitemsep, topsep=0pt]
        \item Doors are represented between node pairs.
        \item If the scene graph edge between node\_1 and node\_2 has a door attribute, initialize the corresponding door state predicates consistent with the domain.
        \item The door-opening skill signature is: \texttt{open\_door(robot, hand, node\_1, node\_2)}. Your problem initialization should support door state for the node pair, not a standalone door object, unless the domain explicitly models door objects.
        \item If the domain requires a predicate like \texttt{(door\_between node\_1 node\_2)}, include it ONLY if it exists in the domain.
    \end{enumerate}
\end{enumerate}

\noindent Extracting Objects \& Initial Facts (must align with the domain):
\begin{enumerate}[leftmargin=20pt, noitemsep, topsep=2pt]
    \item What counts as an "object" for (:objects):
    \begin{enumerate}[leftmargin=15pt, label=(\alph*), noitemsep, topsep=0pt]
        \item Include:
        \begin{itemize}[leftmargin=10pt, label=-, noitemsep, topsep=0pt]
            \item robot(s) and hands (as required by the domain)
            \item all necessary navigable nodes (room/pose/asset) as location objects
            \item all necessary item/object identifiers attached to nodes from the pruned scene graph
        \end{itemize}
        \item Do NOT include any identifiers that do not appear in the scene graph input.
    \end{enumerate}
    \item Object categories / types: \\
    Determine whether the domain uses:
    \begin{enumerate}[leftmargin=15pt, label=(\alph*), noitemsep, topsep=0pt]
        \item \texttt{:types} in \texttt{(:objects)}, OR
        \item unary type predicates asserted in \texttt{(:init)}.
    \end{enumerate}
    Follow the domain's style exactly. Do not mix styles.
    \item Initial positions / containment relations:
    \begin{enumerate}[leftmargin=15pt, label=(\alph*), noitemsep, topsep=0pt]
        \item For each object listed under a node/asset, assert the correct spatial relation predicates required by the domain (e.g., \texttt{on\_}, \texttt{in\_}).
        \item If an asset node is a container/device (e.g., fridge), treat it as both a navigable node AND a manipulable object/device, using only predicates the domain supports.
    \end{enumerate}
    \item Hand state / holding: \\
    Initialize hands as free unless specified otherwise, using the domain's exact predicates for holding/hand-free.
    \item Open/closed / on/off states: \\
    Initialize state facts (e.g., \texttt{is\_open}, \texttt{is\_on}) ONLY if such predicates exist in the domain and the scene provides this information. If unknown, prefer a conservative default (closed/off).
    \item Accessibility / occlusion (optional): \\
    If the domain contains accessibility predicates (e.g., \texttt{covered}), initialize them only if the scene provides enough information. Otherwise, omit them.
\end{enumerate}

\noindent Goal Construction Rules:
\begin{enumerate}[leftmargin=20pt, noitemsep, topsep=2pt]
    \item Translate \{instruction\} into a formal goal using ONLY predicates defined in \{domain\_pddl\}.
    \item The goal must be a conjunction \texttt{(and ...)} of atomic predicates.
    \item Do NOT use quantifiers (\texttt{forall/exists}) or any keyword not present in the domain.
    \item Keep the goal minimal: only include predicates necessary to represent task completion.
\end{enumerate}

\noindent Output Requirement: \\
Output JSON with two fields:
\begin{enumerate}[leftmargin=20pt, noitemsep, topsep=2pt]
    \item "reasoning": A brief, step-by-step reasoning describing:
    \begin{enumerate}[leftmargin=15pt, label=(\alph*), noitemsep, topsep=0pt]
        \item how you interpreted nodes vs. objects,
        \item how you extracted \texttt{(:objects)} consistent with the domain typing style,
        \item how you initialized connectivity and doors,
        \item how you mapped scene object placements/states into init predicates,
        \item how you formed goal predicates from the instruction.
    \end{enumerate}
    \item "problem": A complete PDDL problem file as a multiline string, including helpful comments like ``; Objects'', ``; Connections'', ``; Doors'', ``; Positions / Containment'', ``; States'', and ``; Begin goal'' / ``; End goal''.
\end{enumerate}

\noindent Example Output Format:
\begin{quote}
\begin{verbatim}
{
  "reasoning": "...",
  "problem": "(define (problem ...)\n (:domain ...)\n
  (:objects ...)\n (:init ...)\n (:goal ...)\n)"
}
\end{verbatim}
\end{quote}
\end{sayplanbox}

\newpage
\paragraph{Simulated Environment Object Mapping Prompt}
This prompt directs the Semantic Object Grounding Assistant to map high-level plan objects to concrete simulation assets based on semantic relevance.
\begin{sayplanbox}
You are an expert Semantic Object Grounding Assistant for a robotic system. Your task is to map ``Query Objects'' (generated by a high-level planner) to ``Candidate Objects'' (existing in the simulation environment) based on semantic meaning.

\noindent Input Format: \\
You will receive a list of matching tasks. Each task represents a specific location (Node) and contains:
\begin{enumerate}[leftmargin=20pt, noitemsep, topsep=2pt]
    \item node\_id: The name of the location.
    \item plan\_objects: A list of object names the robot WANTS to manipulate (Queries).
    \item env\_objects: A list of object names ACTUALLY available in the environment (Candidates).
\end{enumerate}

\noindent Matching Rules:
\begin{enumerate}[leftmargin=20pt, noitemsep, topsep=2pt]
    \item Category Priority:
    \begin{enumerate}[leftmargin=15pt, label=(\alph*), noitemsep, topsep=0pt]
        \item You must prioritize the object's noun or category over its adjectives.
        \item Example: ``white\_plate'' matches ``red\_plate'' (same category) significantly better than ``white\_table'' (same adjective but different category). Never match objects with different core categories just because they share a color or attribute string.
    \end{enumerate}
    \item Attribute Tie-Breaking:
    \begin{enumerate}[leftmargin=15pt, label=(\alph*), noitemsep, topsep=0pt]
        \item If multiple candidates have the same category, use attributes (color, size, state) to distinguish them.
        \item Example: If the Query is ``red\_apple'' and Candidates are [``red\_apple\_v1'', ``green\_apple\_v2''], pick ``red\_apple\_v1''.
    \end{enumerate}
    \item Forced Alignment:
    \begin{enumerate}[leftmargin=15pt, label=(\alph*), noitemsep, topsep=0pt]
        \item If the Query describes an attribute that does not exist in the Candidates (e.g., Query ``bigger\_apple'' but Candidates are only ``green\_apple'' and ``red\_apple''), you must still pick one valid candidate of the correct category. Do not leave it unmatched. In such cases, choose the most logical option or arbitrarily pick one to satisfy the mapping requirement, provided it is the correct object type.
    \end{enumerate}
    \item One-to-One Mapping:
    \begin{enumerate}[leftmargin=15pt, label=(\alph*), noitemsep, topsep=0pt]
        \item Each plan\_object must match exactly one env\_object. Each env\_object can be used at most once per node. You must find the global optimal assignment for that node.
    \end{enumerate}
\end{enumerate}

\noindent Example:
\begin{quote}
Input:
\begin{verbatim}
[
  {
    "node_id": "Kitchen_Counter",
    "plan_objects": ["white_plate", "bigger_apple"],
    "env_objects": ["red_plate_v1", "white_table_main",
                    "green_apple_01", "red_apple_02"]
  },
  {
  "node_id": "LivingRoom_Table",
  "plan_objects": ["cap", "remote"],
  "env_objects": ["hat_v1", "wooden_table",
                  "tv_remote_control"]
  }
]
\end{verbatim}

Output:
\begin{verbatim}
{
  "reasoning": "Node Kitchen_Counter: 1. 'white_plate': Candidate
  'white_table_main' shares color but is wrong category. Candidate
  'red_plate_v1' is correct category. Match -> 'red_plate_v1'.
  2. 'bigger_apple': No 'bigger' attribute found in candidates,
  but 'green_apple_01' and 'red_apple_02' are apples. Arbitrarily
  match -> 'green_apple_01'. Node LivingRoom_Table: 1. 'cap' is
  headwear, matches synonym 'hat_v1'. 2. 'remote' matches
  'tv_remote_control'.",
  "mapping": {
    "Kitchen_Counter": {
      "white_plate": "red_plate_v1",
      "bigger_apple": "green_apple_01"
    },
    "LivingRoom_Table": {
      "cap": "hat_v1",
      "remote": "tv_remote_control"
    }
  }
}
\end{verbatim}
\end{quote}

\noindent Task Data: \{task\_data\_json\} \\

\noindent Output Requirement: \\
Return a valid JSON object only. The keys in the inner mapping must be the exact strings from plan\_objects, and values must be the exact strings from env\_objects.
\end{sayplanbox}

\newpage
\paragraph{Simulated Environment Object Mapping Prompt}
This prompt directs the Semantic Object Grounding Assistant to map high-level plan objects to concrete simulation assets based on semantic relevance.

\begin{sayplanbox}
You are an expert Semantic Object Grounding Assistant for a robotic system. Your task is to map ``Query Objects'' (generated by a high-level planner) to ``Candidate Objects'' (existing in the simulation environment) based on semantic meaning.

\noindent Input Format: \\
You will receive a list of matching tasks. Each task represents a specific location (Node) and contains:
\begin{enumerate}[leftmargin=20pt, noitemsep, topsep=2pt]
    \item node\_id: The name of the location.
    \item plan\_objects: A list of object names the robot WANTS to manipulate (Queries).
    \item env\_objects: A list of object names ACTUALLY available in the environment (Candidates).
\end{enumerate}

\noindent Matching Rules:
\begin{enumerate}[leftmargin=20pt, noitemsep, topsep=2pt]
    \item Category Priority:
    \begin{enumerate}[leftmargin=15pt, label=(\alph*), noitemsep, topsep=0pt]
        \item You must prioritize the object's noun or category over its adjectives.
        \item Example: ``white\_plate'' matches ``red\_plate'' (same category) significantly better than ``white\_table'' (same adjective but different category). Never match objects with different core categories just because they share a color or attribute string.
    \end{enumerate}
    \item Attribute Tie-Breaking:
    \begin{enumerate}[leftmargin=15pt, label=(\alph*), noitemsep, topsep=0pt]
        \item If multiple candidates have the same category, use attributes (color, size, state) to distinguish them.
        \item Example: If the Query is ``red\_apple'' and Candidates are [``red\_apple\_v1'', ``green\_apple\_v2''], pick ``red\_apple\_v1''.
    \end{enumerate}
    \item Forced Alignment:
    \begin{enumerate}[leftmargin=15pt, label=(\alph*), noitemsep, topsep=0pt]
        \item If the Query describes an attribute that does not exist in the Candidates (e.g., Query ``bigger\_apple'' but Candidates are only ``green\_apple'' and ``red\_apple''), you must still pick one valid candidate of the correct category. Do not leave it unmatched. In such cases, choose the most logical option or arbitrarily pick one to satisfy the mapping requirement, provided it is the correct object type.
    \end{enumerate}
    \item One-to-One Mapping:
    \begin{enumerate}[leftmargin=15pt, label=(\alph*), noitemsep, topsep=0pt]
        \item Each plan\_object must match exactly one env\_object. Each env\_object can be used at most once per node. You must find the global optimal assignment for that node.
    \end{enumerate}
\end{enumerate}

\noindent Example:
\begin{quote}
Input:
\begin{verbatim}
[
  {
    "node_id": "Kitchen_Counter",
    "plan_objects": ["white_plate", "bigger_apple"],
    "env_objects": ["red_plate_v1", "white_table_main",
                    "green_apple_01", "red_apple_02"]
  },
  {
  "node_id": "LivingRoom_Table",
  "plan_objects": ["cap", "remote"],
  "env_objects": ["hat_v1", "wooden_table",
                  "tv_remote_control"]
  }
]
\end{verbatim}

Output:
\begin{verbatim}
{
  "reasoning": "Node Kitchen_Counter: 1. 'white_plate': Candidate
  'white_table_main' shares color but is wrong category. Candidate
  'red_plate_v1' is correct category. Match -> 'red_plate_v1'.
  2. 'bigger_apple': No 'bigger' attribute found in candidates,
  but 'green_apple_01' and 'red_apple_02' are apples. Arbitrarily
  match -> 'green_apple_01'. Node LivingRoom_Table: 1. 'cap' is
  headwear, matches synonym 'hat_v1'. 2. 'remote' matches
  'tv_remote_control'.",
  "mapping": {
    "Kitchen_Counter": {
      "white_plate": "red_plate_v1",
      "bigger_apple": "green_apple_01"
    },
    "LivingRoom_Table": {
      "cap": "hat_v1",
      "remote": "tv_remote_control"
    }
  }
}
\end{verbatim}
\end{quote}

\noindent Task Data: \{task\_data\_json\} \\

\noindent Output Requirement: \\
Return a valid JSON object only. The keys in the inner mapping must be the exact strings from plan\_objects, and values must be the exact strings from env\_objects.
\end{sayplanbox}

\newpage
\paragraph{Ablation Study: Without Vision Prompt}
This prompt is used for the ablation study where the vision-based verification is disabled, forcing the model to generate the PDDL problem file relying solely on the textual scene graph.

\begin{sayplanbox}
You are an expert PDDL problem generator for a robotic agent. Your goal is to generate a PDDL problem file based on a given PDDL domain and a JSON scene graph, where keys represent asset names and values list the objects located in or on them.

\noindent Inputs:
\begin{enumerate}[leftmargin=20pt, noitemsep, topsep=2pt]
    \item Instructions: \{instructions\}
    \item Available PDDL domain: \{domain\}
    \item Scene Graph: \{scene\_graph\}
\end{enumerate}

\noindent Output Requirements: \\
Your output should be a JSON object containing four parts:
\begin{enumerate}[leftmargin=20pt, noitemsep, topsep=2pt]
    \item "reasoning": Analyze the scene graph and output your reasoning.
    \item "objects": Locate objects related to the task from the assets respectively.
    \item "init": Describe the PDDL init state based on the assets and the given PDDL predicates.
    \item "goal": Generate the PDDL goal from the human instructions.
\end{enumerate}

\noindent Notes:
\begin{enumerate}[leftmargin=20pt, noitemsep, topsep=2pt]
    \item When generating the PDDL init, you should list predicates in the following order: type predicates, state predicates, spatial/position relationship predicates, and affordance predicates.
    \item You should generate the objects and init state based on each asset respectively.
    \item When generating the output, you should analyze the goal, recognize the objects related to it, and only include relevant objects and initial facts.
    \item Some objects that seem unrelated to the goal but are necessary for solving the task (e.g., blocked or hindering objects) should also be recognized and included. Focus only on the scene and the goal; IGNORE any information about the robot (e.g., \texttt{hand\_free}, \texttt{holding}, \texttt{robot}).
    \item It is not necessary to use the exact names from the provided scene graph. When naming objects, prioritize color attributes as the primary grounding feature. This includes the color of the object's body, its lid or cap, or the contents inside.
\end{enumerate}

\noindent Example Output Format:
\begin{quote}
\begin{verbatim}
{
    "reasoning": "your analysis",
    "objects": {
        "asset_name_1": ["red_apple_1", "green_apple_1",
                         "yellow_bowl_1"],
        "asset_name_2": ["red_apple_2", "orange_bowl_2"]
    },
    "init": [
        "(on_table yellow_bowl_1)",
        "(on_table orange_bowl_2)"
    ],
    "goal": "(and (in red_apple_1 orange_bowl_2)
             (in red_apple_2 yellow_bowl_1))"
}
\end{verbatim}
\end{quote}
\end{sayplanbox}

\newpage
\paragraph{Ablation Study: Without Expansion Prompt}
This prompt is used for the ablation study where the hierarchical scene graph expansion is bypassed, asking the model to generate the PDDL problem directly from specific images and the domain.

\begin{sayplanbox}
You are an expert PDDL problem generator for a robotic system. Your goal is to generate a PDDL problem file based on a given PDDL domain and images depicting various parts of the environment.

\noindent Inputs:
\begin{enumerate}[leftmargin=20pt, noitemsep, topsep=2pt]
    \item Instructions: \{instructions\}
    \item Available PDDL domain: \{domain\}
    \item Image\_IDs: \{image\_ids\}
\end{enumerate}

\noindent Output Requirements: \\
Your output should include four parts:
\begin{enumerate}[leftmargin=20pt, noitemsep, topsep=2pt]
    \item "reasoning": Analyze the image and provide the reasoning.
    \item "objects": Locate objects related to the task from the images respectively.
    \item "init": Describe the PDDL init state from the images based on given PDDL predicates respectively.
    \item "goal": Generate the PDDL goal from human instructions.
\end{enumerate}

\noindent Notes:
\begin{enumerate}[leftmargin=20pt, noitemsep, topsep=2pt]
    \item When generating the PDDL init state, you should list predicates in the following order: type predicates, state predicates, spatial or position relationship predicates, and affordance predicates.
    \item You should generate objects and the initial state based on each image respectively.
    \item When generating, you should analyze the goal, recognize the objects related to the goal, and only include relevant objects and initial facts.
    \item Some objects that are seemingly not related to the goal, but are necessary for solving the task (e.g., blocked or hindered objects), should be recognized and included as well. Focus only on the scene and the goal; IGNORE any information about the robot (e.g., \texttt{hand\_free}, \texttt{holding}, \texttt{robot}, etc.).
    \item The image\_ids correspond to the images in the order provided.
    \item When naming objects, prioritize color attributes as the primary grounding feature. This includes the color of the object's body, the color of its lid or cap, or the color of the contents inside.
\end{enumerate}

\noindent Example Output Format:
\begin{quote}
\begin{verbatim}
{
    "reasoning": "your analysis",
    "objects": {
        "image_id_1": ["red_apple_1", "green_apple_1",
                       "yellow_bowl_1"],
        "image_id_2": ["red_apple_2", "orange_bowl_2"]
    },
    "init": [
        "(on_table yellow_bowl_1)",
        "(on_table orange_bowl_2)"
    ],
    "goal": "(and (in red_apple_1 orange_bowl_2)
             (in red_apple_2 yellow_bowl_1))"
}
\end{verbatim}
\end{quote}
\end{sayplanbox}

\newpage
\paragraph{Ablation Study: Without Injection Prompt}
This prompt is for the ablation study where the object injection step is removed, requiring the model to generate the complete PDDL problem file directly from the domain and scene images without intermediate pruning.

\begin{sayplanbox}
You are an expert PDDL problem generator for a mobile manipulation robot. Your goal is to generate a PDDL problem file based on a given PDDL domain and images showing the different parts of the whole scene.

\noindent Environment Configuration:
\begin{enumerate}[leftmargin=20pt, noitemsep, topsep=2pt]
    \item Available Robot(s): \{robot\_name\}
    \item Available Hand(s): \{hand\_names\}
    \item Robot Initial Location: \{robot\_init\_node\}
    \item Scene Graph Topology (Connectivity): \{scene\_graph\}
    \item Task-Relevant Key Nodes (corresponding to the provided images in order): \{key\_nodes\}
\end{enumerate}

\noindent Available PDDL Domain: \\
\{domain\} \\

\noindent Task Instruction: \\
\{instruction\} \\

\noindent Notes:
\begin{enumerate}[leftmargin=20pt, noitemsep, topsep=2pt]
    \item When generating PDDL init, you should use type predicates, state predicates, spatial or position relationship predicates, and affordance predicates in order.
    \item When generating, you should analyze the goal, recognize the objects related to the goal, and only include relevant objects and initial facts.
    \item Some objects that are seemingly not related to the goal, but are necessary for solving the task (e.g., blocked or hindered objects), should be recognized and included as well.
    \item DO NOT use any type declarations in the generated problem file. Keep consistency with the domain file (e.g., using unary type predicates in the init section if required).
    \item When naming objects, prioritize color attributes as the primary grounding feature. This includes the color of the object's body, the color of its lid or cap, or the color of the contents inside.
\end{enumerate}

\noindent Output Requirement: \\
Return a JSON object with the following structure:
\begin{quote}
\begin{verbatim}
{
    "reasoning": "your analysis",
    "problem_file": "the complete PDDL problem file"
}
\end{verbatim}
\end{quote}
\end{sayplanbox}

\end{document}